\documentclass[preprint,12pt]{elsarticle}

\usepackage{amssymb}

\usepackage{fontawesome5} 
\usepackage{xcolor}
\usepackage{booktabs}
\usepackage{colortbl}
\usepackage{graphicx}
\usepackage{caption}
\usepackage{multirow}
\usepackage[table]{xcolor}
\usepackage{booktabs}
\usepackage{pifont}
\usepackage{graphicx}
\graphicspath{{Fig/}} 
\usepackage{amsmath}
\usepackage{booktabs}   
\usepackage{amsmath}    
\usepackage{bm}         
\usepackage{multirow}   
\PassOptionsToPackage{hyphens}{url}
\PassOptionsToPackage{hidelinks}{hyperref}
\usepackage{hyperref}
\usepackage{xcolor}
\usepackage[hidelinks]{hyperref}
\usepackage[table]{xcolor}
\usepackage{array}
\usepackage{url}
\usepackage{colortbl}
\usepackage{arydshln} 
\usepackage{hyperref}
\usepackage{threeparttable}
\hypersetup{
    colorlinks=true,
    linkcolor=blue,
    urlcolor=blue,
    citecolor=blue
}
\usepackage{xcolor}
\definecolor{best}{HTML}{FFCCC9}   
\definecolor{second}{HTML}{FFE4CF} 
\definecolor{third}{HTML}{FFFFD4}  
\definecolor{customPurple}{RGB}{065,014,115} 
\usepackage{txfonts}
\usepackage{wasysym} 
\usepackage{amssymb} 
\definecolor{customHeader}{HTML}{D8D6C0} 
\newcolumntype{L}{>{\raggedright\arraybackslash\large}l}            
\newcolumntype{C}{>{\centering\arraybackslash\large}c}              
\newcolumntype{S}{>{\raggedright\arraybackslash\footnotesize}p{0.62\textwidth}}

\definecolor{groupA}{RGB}{236,248,236} 
\definecolor{groupB}{RGB}{255,243,224} 
\definecolor{groupC}{RGB}{232,246,252} 
\definecolor{groupD}{RGB}{244,238,252} 

\DeclareRobustCommand{\PHtwo}{\ifmmode \mathrm{PH}^{2}\else PH\textsuperscript{2}\fi}

\usepackage{caption}

\journal{Nuclear Physics B}

\begin{document}

\begin{frontmatter}

\title{A Large Scale Benchmark for Test Time Adaptation Methods in Medical Image Segmentation} 


\author[hdu]{Wenjing Yu\corref{equal1}}
\author[hdu]{Shuo Jiang\corref{equal1}}
\author[thu]{Yifei Chen\corref{equal1}}
\author[hdu]{Shuo Chang}
\author[hdu]{Yuanhan Wang}
\author[hdu]{Beining Wu}
\author[hdu]{Jie Dong}
\author[thu]{Mingxuan Liu}
\author[hdu]{Shenghao Zhu}
\author[hdu]{Feiwei Qin\corref{cor1}}
\author[wcm]{Changmiao Wang}
\author[thu]{Qiyuan Tian\corref{cor1}}

\affiliation[hdu]{organization={Hangzhou Dianzi University},
    city={Hangzhou},
    postcode={310018},
    country={China}}

\affiliation[thu]{organization={Tsinghua University},
            city={Beijing},
            postcode={100084},
            country={China}}

\affiliation[wcm]{organization={Shenzhen Research Institute of Big Data},
            city={Shenzhen},
            postcode={518172},
            country={China}}

\cortext[equal1]{These authors contributed equally to this work.}
\cortext[cor1]{Corresponding authors: qinfeiwei@hdu.edu.cn, qiyuantian@tsinghua.edu.cn.}

\begin{abstract}
Test time Adaptation is a promising approach for mitigating domain shift in medical image segmentation; however, current evaluations remain limited in terms of modality coverage, task diversity, and methodological consistency. We present MedSeg-TTA, a comprehensive benchmark that examines twenty representative adaptation methods across seven imaging modalities, including MRI, CT, ultrasound, pathology, dermoscopy, OCT, and chest X-ray, under fully unified data preprocessing, backbone configuration, and test time protocols. The benchmark encompasses four significant adaptation paradigms: Input-level Transformation, Feature-level Alignment, Output-level Regularization, and Prior Estimation, enabling the first systematic cross-modality comparison of their reliability and applicability. The results show that no single paradigm performs best in all conditions. Input-level methods are more stable under mild appearance shifts. Feature-level and Output-level methods offer greater advantages in boundary-related metrics, whereas prior-based methods exhibit strong modality dependence. Several methods degrade significantly under large inter-center and inter-device shifts, which highlights the importance of principled method selection for clinical deployment. MedSeg-TTA provides standardized datasets, validated implementations, and a public leaderboard, establishing a rigorous foundation for future research on robust, clinically reliable test-time adaptation. All source codes and open-source datasets are available at
\href{https://github.com/wenjing-gg/MedSeg-TTA}{https://github.com/wenjing-gg/MedSeg-TTA}.
\end{abstract}


\end{frontmatter}



\section{Introduction}
\label{sec1}

Medical image segmentation plays a pivotal role in clinical diagnosis, therapeutic evaluation, and preoperative planning~\cite{you2024learning, ZhuShe_Bridging_MICCAI2025}. Its precision directly determines the reliability of lesion quantification, organ morphometric analysis, and the development of individualized treatment strategies~\cite{meng2024multi, Zhu2025NoML,10981275}. However, in practical applications, distribution shifts across institutions, scanners, and populations severely hinder the generalization capacity of deep learning models, leading to substantial performance degradation when tested on the target domain~\cite{zheng2024dual, zenk2025comparative, liu2025spectrum}. To address this challenge, Test-Time Adaptation (TTA) has recently attracted considerable attention~\cite{niu2022efficient, zhu2025improving}. Unlike conventional domain adaptation methods, TTA neither requires access to source domain data nor an additional training stage in the target domain~\cite{li2025multi, Wang2025SmaRTSR}. Instead, it dynamically adjusts model parameters during deployment and instantaneously corrects predictions in a batch-wise or streaming manner, thereby offering greater clinical applicability~\cite{wang2021tent}. Liang \textit{et al.}~\cite{liang2025comprehensive} systematically categorized existing TTA approaches into four groups based on their adaptation mechanisms: Input-level Transformation, Feature-level Alignment, Output-level Regularization, and Prior Estimation. This taxonomy provides a clear framework for subsequent investigations.

Input-level Transformation methods adjust the appearance and statistical characteristics of test samples in the pixel or frequency domain, making their distributions more similar to those of the source domain, thereby enabling inference under familiar input conditions. A representative example is the reliable source approximation method proposed by Zeng \textit{et al.}~\cite{zeng2024reliable}, which uses diffusion models to map target-domain images into the source-domain style before prediction. Feature-level Alignment focuses on reducing distributional discrepancies between the source and target domains in the intermediate representation space. A typical approach is adversarial training, in which Hu \textit{et al.}~\cite{ganin2016domain} encouraged discriminative features to become domain-invariant through the interplay of a domain discriminator and a feature extractor. In addition, lightweight adaptation and statistical recalibration can be integrated to achieve latent space alignment without modifying the backbone architecture~\cite{guo2017calibration}. Output-level Regularization methods treat the model’s predictions as supervisory signals and impose self-supervised constraints directly on the output distribution to enhance cross-domain robustness. Common strategies include entropy minimization, Maximum Squares, Batch Nuclear-norm Maximization, and mutual information maximization. For instance, Weihsbach \textit{et al.}~\cite{weihsbach2023dg} improved the stability of out-of-distribution predictions by introducing a dual-branch consistency mechanism. Prior Estimation methods explicitly incorporate relatively stable categorical or anatomical priors during inference to constrain the reasoning process via energy functions, regularization terms, or conditional prompts. For example, Hu \textit{et al.}~\cite{hu2025source} utilized prompt learning to encode both task semantics and domain-level priors, thereby guiding segmentation models toward convergent alignment when source data are unavailable.

Despite substantial advances in related research, the application of TTA to medical image segmentation still encounters several bottlenecks. The main limitations are as follows: most existing studies demonstrate effectiveness only on single-modality, single-task, or single-center datasets, with little systematic evaluation across diverse imaging modalities and tasks~\cite{yi2024question}. Evaluation protocols are also highly heterogeneous, and there is no agreement on critical design choices, such as whether cross-sample accumulation of statistics or external updates to batch normalization parameters are allowed. Some experimental setups further pose risks of source data or label leakage. In addition, evaluation metrics are often oversimplified, with many works reporting only the Dice coefficient while neglecting structurally sensitive measures such as Sen and PPV, thereby weakening comparability and clinical interpretability. Collectively, these issues hinder fair benchmarking and reproducibility, making it difficult to faithfully characterize the strengths, limitations, and failure modes of different TTA paradigms under heterogeneous-domain and multi-task conditions~\cite{wang2025search}.

Inspired by the comprehensive review of Liang \textit{et al.}~\cite{liang2025comprehensive}, this study introduces a unified evaluation framework that, for the first time in the field of medical imaging, provides a systematic benchmark across various modalities, centers, and tasks. We assemble datasets from seven primary imaging modalities: MRI, CT, US, PATH, Dermoscopy (DER), OCT, and CXR. Under a unified backbone, data partitioning, and testing protocol, we systematically compare the performance of the four paradigms: Input-level Transformation, Feature-level Alignment, Output-level Regularization, and Prior Estimation.

The main contributions are summarized as follows: 
\begin{itemize}
    \item \textbf{Multi-modal and multi-center open-source dataset:} We construct a dataset covering tumor, organ, and lesion segmentation across seven imaging modalities: MRI, CT, US, PATH, DER, OCT, and CXR. The dataset uses standardized preprocessing and partitioning, reflecting distribution shifts across institutions, scanners, and populations, thereby providing a basis for unified TTA in medical image segmentation benchmarks.

    \item \textbf{Strong and reproducible TTA baselines:} We design a unified TTA protocol that fixes the backbone and strictly forbids source-domain access as well as any implicit information leakage. In this setting, we systematically reproduce and validate 20 representative state-of-the-art TTA methods for medical image segmentation across 4 paradigms.

    \item \textbf{Unified benchmark and public leaderboard:} Building on these implementations, we establish a public benchmark and leaderboard that support comparisons across imaging modalities, organs, and tasks using both region-overlap and structurally sensitive metrics such as Dice and HD95.

    \item \textbf{Paradigm taxonomy and applicability lineage:} We categorize TTA methods into four paradigms according to their locus of operation and, based on evaluations across modalities, organs, and tasks, construct \emph{lineage maps} that highlight effective and ineffective regimes. This delineates applicability boundaries and provides practical guidance for future method selection.
\end{itemize}

\section{Related Work}
\label{sec2}

From a functional perspective, TTA methods for medical image segmentation can be broadly categorized into four paradigms~\cite{liang2025comprehensive}. Input-level Transformation achieves cross-domain consistency by adjusting appearance and statistical styles while preserving structural semantics as much as possible. Feature-level Alignment reconstructs robust latent representations and establishes distributional consistency, thereby enhancing feature resilience against domain shifts. Output-level Regularization imposes constraints on the prediction distribution and decision boundaries, enabling models to maintain stability and coherence under uncertain or out-of-distribution conditions. Prior Estimation introduces relatively stable forms of knowledge, such as anatomy, shape, and class frequency, to narrow the feasible solution space and improve interpretability and consistency of predictions. Around these four themes, TTA has gradually developed into a complete technical taxonomy. The following sections review representative methods and analyze their advantages, applicability boundaries, and common challenges in multi-modal and multi-center scenarios.

\begin{figure}[h!]
    \centering
    \makebox[\textwidth]{\includegraphics[width=\textwidth]{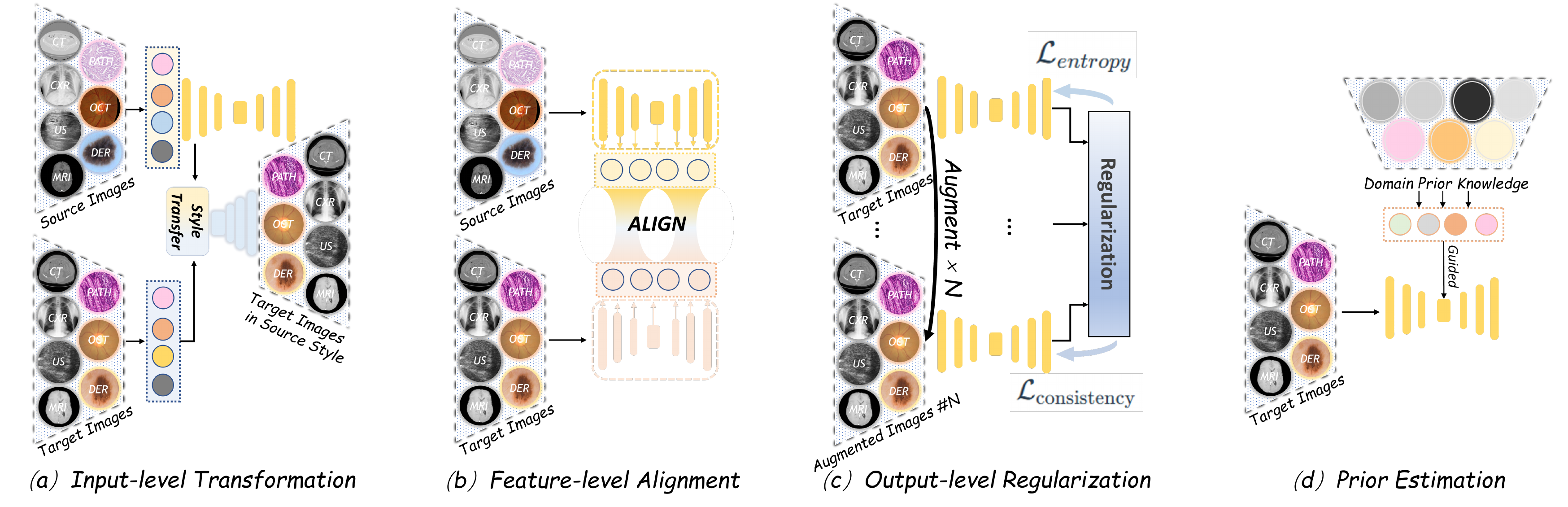}}
    \caption{\textbf{Schematic diagram of the four paradigms in TTA.} (a) Input-level Transformation, adapting the model to target domain images by altering image appearances; (b) Feature-level Alignment, aligning features from the source and target domains; (c) Output-level Regularization, regularizing the model outputs for target domain adaptation; (d) Prior Estimation, forcing the model to update towards the target domain using prior information of the target domain.}
    \label{fig:paradigm}
\end{figure}

\subsection{Input-level Transformation TTA}

Input-level methods generally follow the principle of aligning appearance first, then inferring with the pretrained model. That is, they attenuate cross-center imaging discrepancies by aligning styles or reconstructing images while retaining anatomical semantics, thereby achieving recategorized results without modifying model parameters. Depending on the implementation, they can be divided into style-replacement and generative model approaches.

\subsubsection{Style Replacement}

Style replacement aims to mitigate the discrepancies between the source and target domains by modifying the visual appearance of cross-domain images~\cite{liu2020remove}. A representative work is that of Yang \textit{et al.}~\cite{yang2022source}, who proposed replacing low-frequency amplitudes in the frequency domain to achieve cross-domain style transfer while effectively preserving boundaries and delicate textures. This method assumes that low-frequency differences are the primary source of appearance shifts and demonstrates strong adaptability in 3D medical imaging. Nevertheless, frequency-domain style replacement still faces challenges, including ensuring anatomical consistency and avoiding structural distortions across varying slice thicknesses, window widths, and scanning protocols.

\subsubsection{Generative Models}

Generative model approaches reconstruct target-domain images at the explicit pixel level into a source-style representation, which a pre-trained segmentation model then processes~\cite{li2024learning}. Their powerful modeling capacity enables stronger and more robust appearance remapping across centers and modalities. Existing works can be broadly categorized into two main types: generative adversarial networks and diffusion models~\cite{sandfort2019data, zeng2024reliable}. The former generates images consistent with the target domain through the antagonistic interplay between a generator and a discriminator and has been widely applied in medical imaging. The latter produces high-quality images via a progressive denoising process, thereby substantially enhancing cross-domain segmentation performance. However, generative models still face challenges with computational overhead and potential artifacts, particularly in 3D imaging tasks.

\subsection{Feature-level Alignment TTA}

Feature-level Alignment aims to alleviate distributional shifts in the latent space, rendering intermediate representations more robust and discriminative in the target domain. Omidi \textit{et al.}~\cite{omidi2024unsupervised} analyzed skull stripping from adults to neonates, demonstrating that relying solely on Input-level Transformation is insufficient to handle substantial morphological differences, as style normalization may still result in inter-class entanglement and boundary ambiguity. To address this issue, Zhang \textit{et al.} proposed TestFit~\cite{zhang2024testfit}, which applies one-time, plug-and-play calibration on intermediate representations within the network, achieving lightweight adaptation without access to source data. However, this method remains vulnerable to noise and gradient drift under sequential domain shifts and small-batch inference. Chen \textit{et al.}~\cite{chen2025gradient} introduced a gradient alignment strategy, which constrains parameter update directions to suppress noise amplification and catastrophic forgetting, although stability remains limited. Subsequently, Hu \textit{et al.}~\cite{hu2024unsupervised} proposed mutual information maximization alignment, which enhances feature clustering and inter-cluster separation under unlabeled conditions, thereby improving discriminability when significant appearance discrepancies exist. Nevertheless, this approach is sensitive to the quality of pretrained representations, and optimization may fail to converge when initial features are unsuitable. To further enhance robustness, Wen \textit{et al.}~\cite{wen2024denoising} combined denoising pretraining with test-time alignment, achieving more stable improvements in multi-center segmentation tasks. Feature-level Alignment directly regularizes representations without modifying inputs and incurs low computational overhead. However, it remains sensitive to small-batch statistics and anatomical variability, making it challenging to consistently enforce clear boundaries when used alone.

\subsection{Output-level Regularization TTA}

Output-level Regularization methods treat the model’s prediction distribution in the target domain as a supervisory signal, establishing lightweight optimization objectives during testing that focus on entropy minimization, self-training with pseudo labels, and consistency regularization. This line of work offers low deployment costs and zero reliance on source data, making it a mainstream strategy for TTA in medical image segmentation. Currently, methods can be primarily divided into TENT-type methods, which focus on entropy minimization, and self-supervised methods, which center on self-supervision or pseudo-labeling.

\subsubsection{TENT-Type Methods}

Wang \textit{et al.}~\cite{wang2021tent} proposed TENT, which minimizes prediction entropy and updates only batch normalization statistics and affine parameters. This approach achieves cross-domain improvements without source data and establishes the baseline for Output-level Regularization. However, in small-batch and strong domain-shift scenarios, entropy minimization alone tends to cause overconfidence and cumulative drift. To address this issue, Niu \textit{et al.}~\cite{niu2022eata} introduced EATA, which alleviates noise amplification through uncertainty filtering and conservative regularization, although it is mainly applicable to one-shot domain shifts and struggles with continuous distributional changes. Wang \textit{et al.}~\cite{wang2022cotta} subsequently proposed CoTTA, which incorporates teacher-student exponential moving averages and random replay mechanisms to balance short-term adaptation with long-term memory. However, this approach introduces additional storage and computational overhead in 3D inference. Later, Weihsbach \textit{et al.}~\cite{weihsbach2023dg} presented DG-TTA, which combines domain-generalization pretraining with test-time entropy regularization to improve cold-start stability, although its performance remains limited in adapting to small lesions and scale variations. Additionally, Li \textit{et al.} proposed SaTTCA~\cite{li2023scale}, which incorporates scale-aware geometric priors during testing, thereby improving boundary delineation and recall of small targets; however, it relies heavily on interactions and prior knowledge.

\subsubsection{Self-Supervised Methods}

Another line of research introduces self-supervision or pseudo-labeling mechanisms to mitigate noisy supervision by leveraging consistency regularization and teacher-student frameworks. Sun \textit{et al.}~\cite{sun2020ttt} proposed TTT, which augments representations during inference by paralleling self-supervised tasks, although its semantic coupling with the main task is weak. Liu \textit{et al.}~\cite{liu2021tttpp} proposed TTT$\text{+}$$\text{+}$, which enhances task-relevant self-supervised signals via robust optimization to improve adaptation stability, but it still lacks direct constraints in the output space. Weihsbach \textit{et al.}~\cite{weihsbach2023dg} further proposed dual-branch consistency regularization, which establishes lightweight calibration directly on unlabeled predictions; however, it is susceptible to noise amplification in class-imbalanced settings. Wu \textit{et al.}~\cite{wu2023upl} introduced UPL-SFDA, which reduces noise accumulation by filtering pseudo-labels with uncertainty-aware filtering, thereby significantly improving the quality of pseudo-labels under source-free conditions. However, in continuous and non-stationary scenarios, single regularization remains prone to suboptimality and requires integration with Feature-level Alignment or anatomical priors.

\subsection{Prior Estimation TTA}

Prior-based methods emphasize explicitly incorporating anatomical and shape priors during testing to constrain the rationality and interpretability of predictions. Bateson \textit{et al.}~\cite{bateson2022test} proposed Shape-Moment TTA, which employs geometric moments, such as area and perimeter, as global constraints to ensure predictions adhere to plausible shapes. However, static global priors struggle to capture individual variability. Zhang \textit{et al.} introduced PASS~\cite{zhang2024pass}, which encodes style and semantic shapes into test-time prompts to achieve organ-level controllable alignment, thereby alleviating the under-constrained nature of global statistics. However, prompt acquisition and cross-center generalization remain limited. Chen \textit{et al.}~\cite{chen2024each} further proposed case-specific customized prompts, dynamically injecting location and shape information during inference to enhance individual robustness, though balancing prompt-representation coupling remains challenging. Hu \textit{et al.}~\cite{hu2025source} employed prompt learning to explicitly encode task semantics and domain-level priors, enabling alignment under source-free conditions while ensuring privacy and compliance. Omolegan \textit{et al.}~\cite{omolegan2025exploring} explored combining structural priors with TTA, providing stable anchors in fetal brain 3D ultrasound, though real-time budgets and 3D accuracy remain challenging. As a lightweight approach, DenseCRF can rapidly refine thin-wall structures during testing by leveraging regional similarity and boundary smoothing~\cite{krahenbuhl2011efficient}. This method improves local consistency but lacks global control and is therefore often integrated with shape or prompt priors to achieve overall stability.

\section{Benchmarks and Implementation}
\label{subsec3}

This section outlines the benchmarking protocol, including the datasets and implementation, to facilitate a comprehensive and equitable comparison of TTA approaches for medical image segmentation across multi-modality, multi-task, and multi-center conditions. Section~\ref{subsec3.1} systematically examines seven representative dataset pairs from the same modality and task, namely MRI, CT, US, PATH, DER, OCT, and CXR, which exhibit domain shifts, and then introduces a unified evaluation framework. Section~\ref{subsec3.2} presents standardized implementation, training, and testing protocols that specify the default configurations and constraints for the four TTA paradigms.

\begin{figure}[h!]
  \centering
  \includegraphics[width=\linewidth]{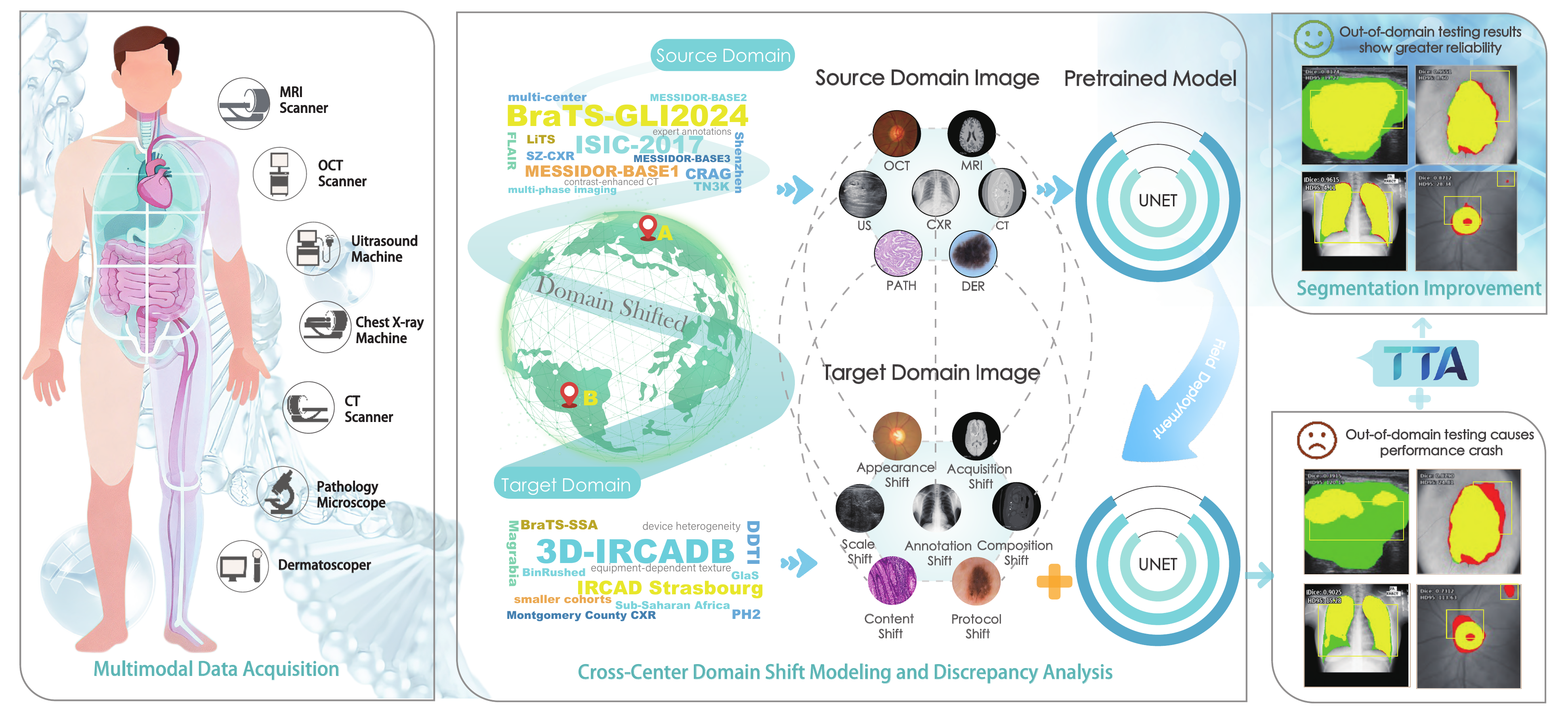}
  \caption{\textbf{The overall workflow of TTA methods for medical image segmentation.} It begins with the collection of multi-center data across seven modalities, which include MRI, CT, US, PATH, DER, OCT, and CXR. Next, a pretrained network is trained on source domain data, and benchmark testing is conducted on the target domain. Finally, out-of-domain segmentation performance improves with TTA methods.}
  \label{fig: framework}
\end{figure}

\subsection{Datasets and Evaluation Metrics} 
\label{subsec3.1}

\begin{figure}[h!]
  \centering
  \includegraphics[width=\linewidth]{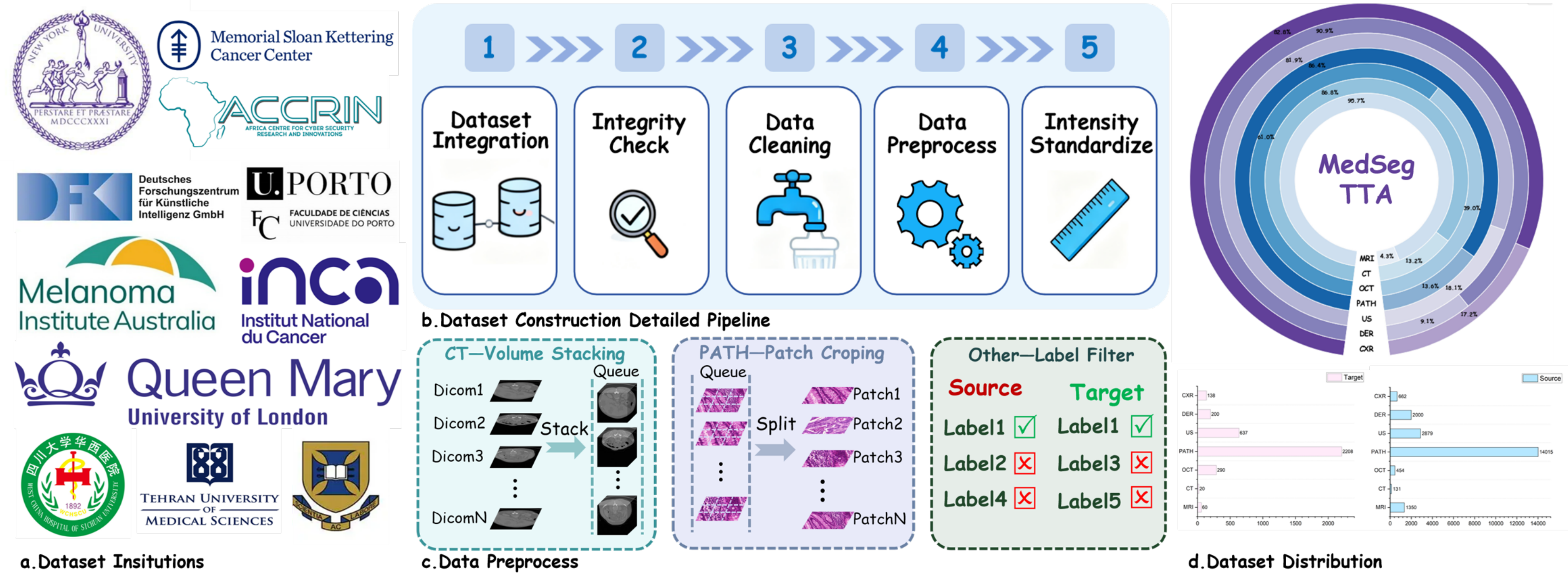}
  \caption{\textbf{Overview of the multi-institutional benchmark and dataset construction pipeline.} (a) participating institutions contributing data for the benchmark;(b) unified construction pipeline comprising dataset integration, integrity checking, data cleaning, data processing, and intensity standardization; and (c) data preprocessing procedures, including CT volume stacking from DICOM slices to 3D volumes, PATH patch cropping from whole-slide images, and source–target label filtering to harmonize label sets. (d) distribution of datasets across imaging modalities and their assignment to source or target domains;}

  \label{fig: datasetv1}
\end{figure}

\subsubsection{MRI-based Brain Tumor Segmentation Datasets}
\begin{figure}[h!]
  \centering
  \includegraphics[width=\linewidth]{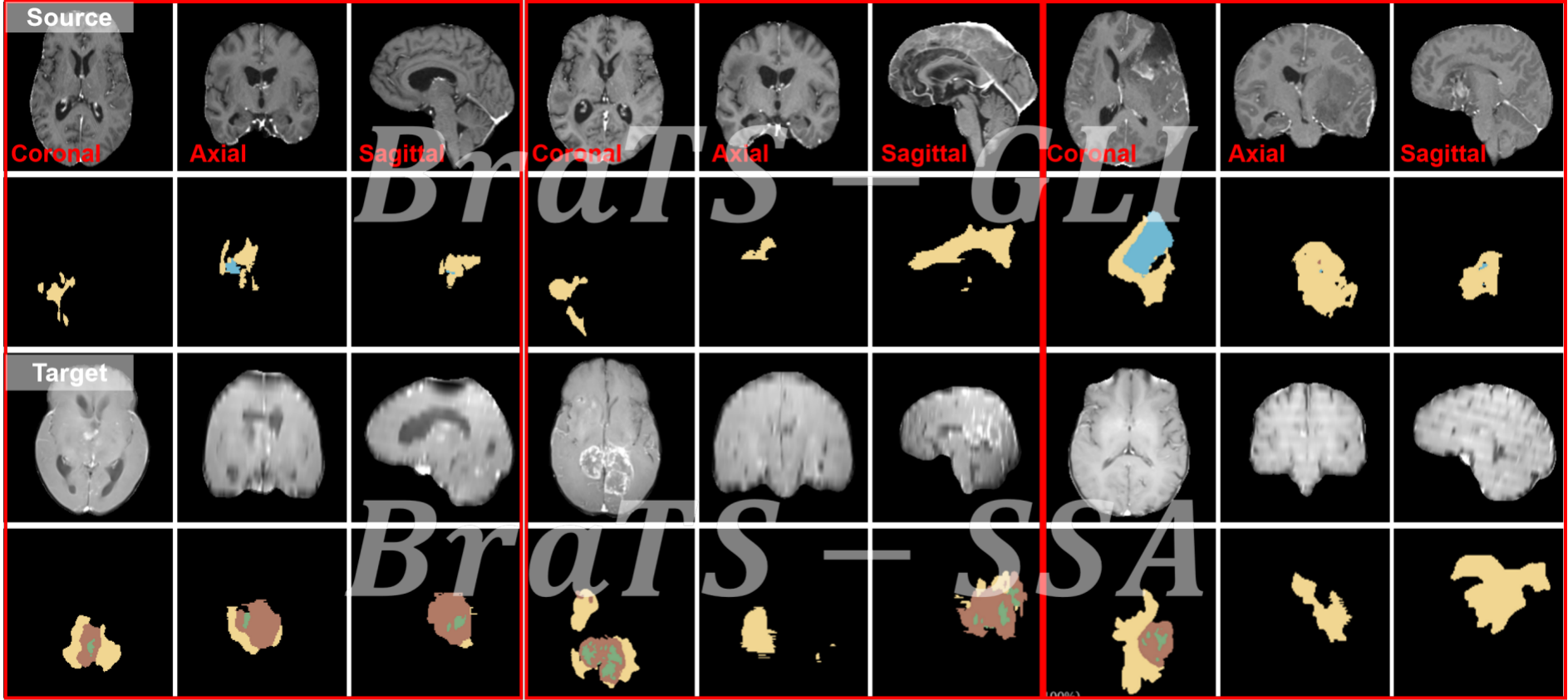}
  \caption{\textbf{Multi-sequence MRI for brain tumor segmentation.} Cases from the source dataset BraTS-GLI2024 and the target dataset BraTS-SSA. For the source dataset area, the first row displays T1, T1-weighted contrast-enhanced, T2 and FLAIR sequence images and  enhancing tumor, tumor core and whole tumor annotations from anatomical views including axial, coronal and sagittal planes; the second row presents the corresponding annotation masks. The bottom section shows the equivalent multi-sequence MRI images and annotations for the target dataset.}
  \label{fig: MRIdataset}
\end{figure}

Source domain: BraTS-GLI2024~\cite{de2024brats}, Target domain: BraTS-SSA~\cite{adewole2023brain}, as shown in Fig.~\ref{fig: MRIdataset}. 
The BraTS series provides large-scale, multi-center, multi-sequence MRI data, including T1, T1c, T2, and FLAIR, with standardized annotations, serving as the de facto benchmark for adult glioma segmentation. We use 1,350 cases from BraTS-GLI2024 as the source-domain training set, with labels for enhancing tumor (ET), tumor core (CT), and whole tumor (WT). BraTS-SSA contains 60 multi-sequence MRI cases from the SSA region. Substantial differences in scanner conditions, acquisition protocols, and population composition naturally induce cross-region and cross-device domain shifts, making it an appropriate target domain for TTA evaluation under resource-limited scenarios.
\subsubsection{CT-based Liver Segmentation Datasets}
\begin{figure}[h!]
  \centering
  \includegraphics[width=\linewidth]{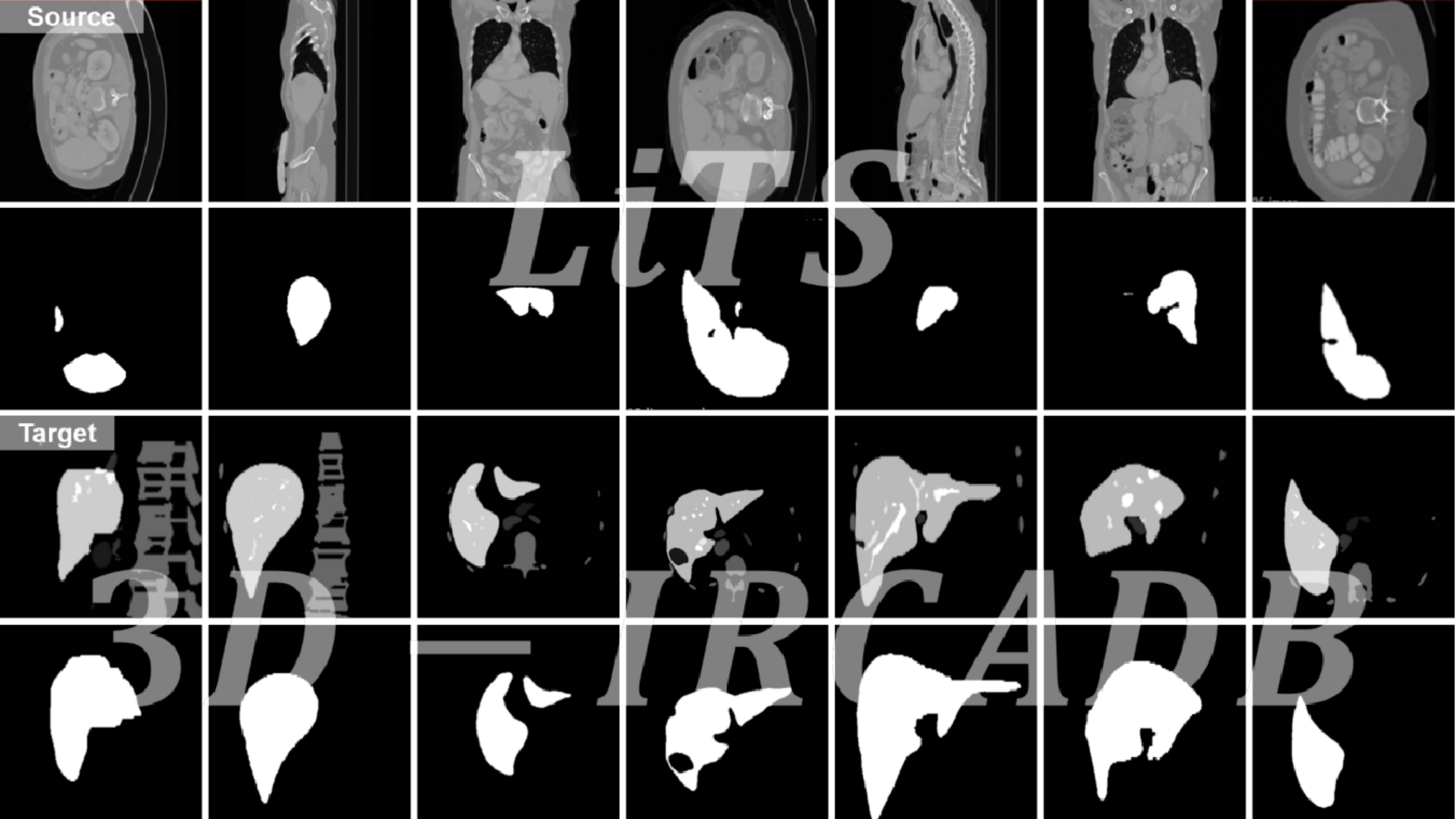}
  \caption{\textbf{CT volumes for liver segmentation.} Cases from the source dataset LiTS and the target dataset 3D-IRCADB. For the source dataset area, the first row displays CT images and liver-tumor annotations from anatomical views, including axial and coronal planes; the second row presents the corresponding annotation masks. The bottom section shows the equivalent CT images and annotations for the target dataset.}
  \label{fig: CTdataset}
\end{figure}

Source domain: LiTS~\cite{bilic2023liver}, Target domain: 3D-IRCADB~\cite{soler20103d}, as shown in Fig.~\ref{fig: CTdataset}. The LiTS dataset comprises multi-institutional, multi-protocol CT volumes of the liver and tumors, from which we select 131 publicly available training cases as the source domain. 3D-IRCADB, collected at IRCAD in France, comprises 20 enhanced CT cases with distinct acquisition devices, contrast phases, and a smaller patient cohort, resulting in a substantial domain shift. While systematic differences exist in contrast phases, convolution kernels, lesion burdens, and institutional sources, the anatomical structure and segmentation task remain consistent, rendering this dataset pair highly suitable for rigorous TTA evaluation.
\subsubsection{OCT-based fundus Segmentation Datasets}
\begin{figure}[h!]
  \centering
  \includegraphics[width=\linewidth]{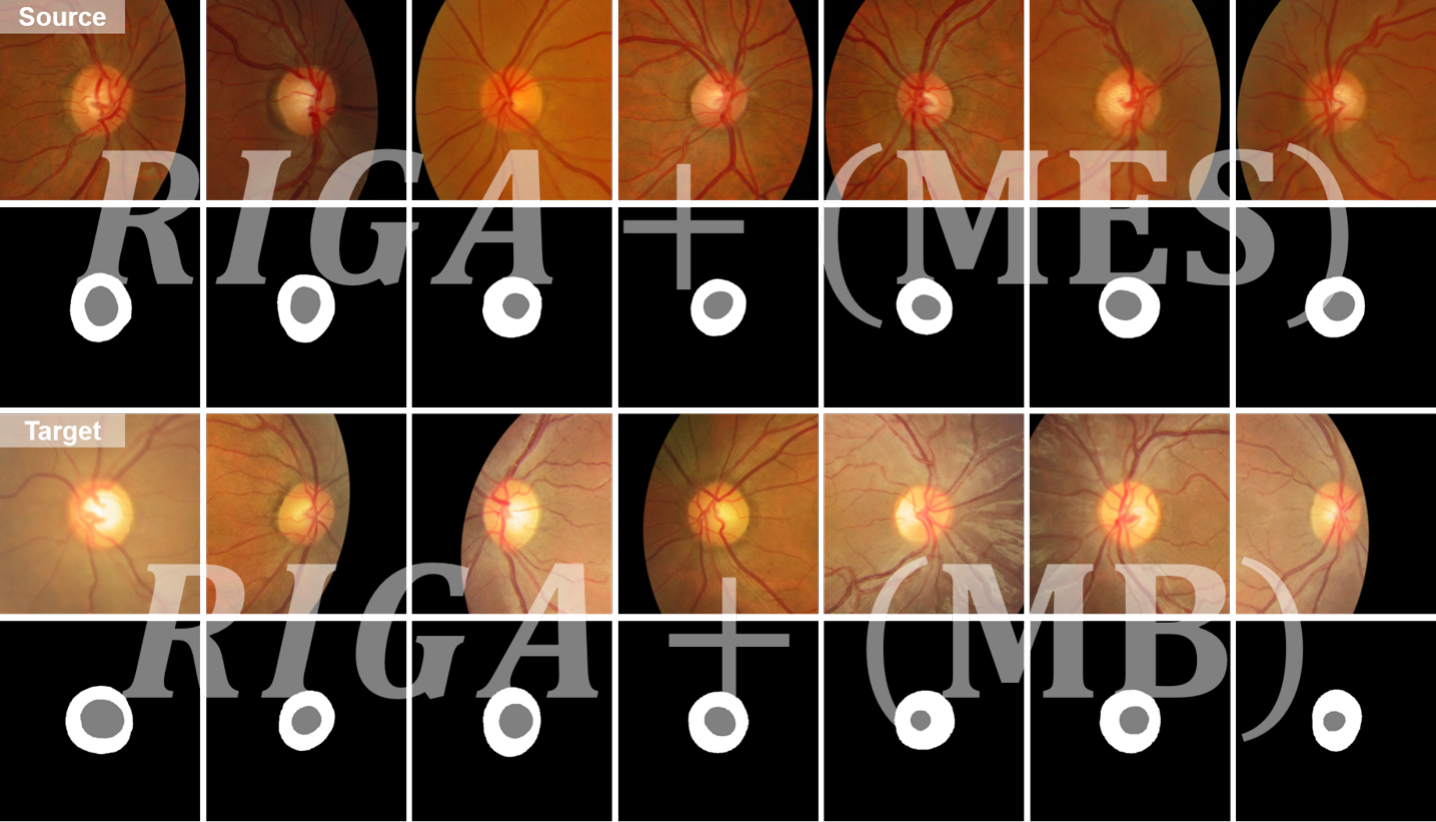}
  \caption{\textbf{Fundus images and annotations for fundus segmentation.} Cases from the source dataset RIGA+ (MES) and the target dataset RIGA+ (MB). In the top source dataset section, the first row displays original glaucomatous fundus images and pixel-level segmentation annotations combined from MESSIDOR-BASE1, BASE2, and BASE3; the second row presents the corresponding annotation masks. The bottom target dataset section shows the combined fundus images and annotations from Magrabia and BinRushed. }
  \label{fig: OCTdataset}
\end{figure}

Source domain combined: MESSIDOR-BASE1, BASE2, and BASE3~\cite{hu2022domain}, Target domain combined: Magrabia and BinRushed~\cite{hu2022domain}, as shown in Fig.~\ref{fig: OCTdataset}.
The RIGA$\text{$\text{+}$}$ collection aggregates fundus images from diverse sources. The MESSIDOR series, acquired with various cameras and workflows, offers a large set of consistently annotated images. Magrabia and BinRushed, collected in Middle Eastern institutions, exhibit marked differences in equipment, illumination, and demographics. We merge glaucomatous fundus images and their pixel-level segmentation annotations from MESSIDOR-BASE1, BASE2, and BASE3 and refer to this dataset as RIGA$\text{$\text{+}$}$ (MES), comprising 454 cases, to form the source domain. We combine Magrabia and BinRushed as RIGA$\text{$\text{+}$}$ (MB), totaling 290 cases, as the target domain. This setup ensures a stable domain shift within the same modality and anatomical region, enabling robust evaluation of TTA methods for fundus segmentation.

\subsubsection{Ultrasound-based Thyroid Nodule Segmentation Datasets}
\begin{figure}[h!]
  \centering
  \includegraphics[width=\linewidth]{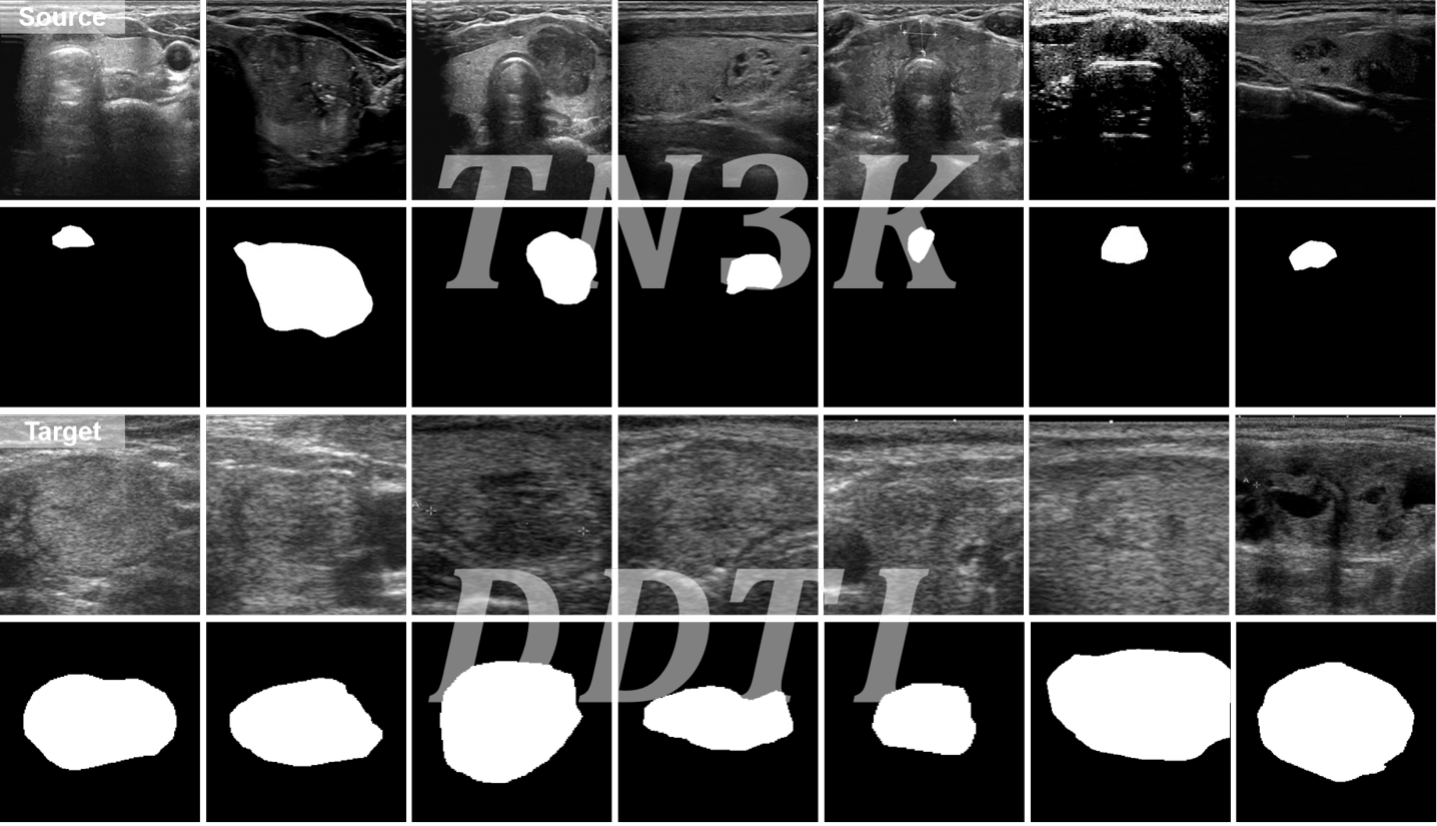}
  \caption{\textbf{Ultrasound images for thyroid nodule segmentation.} Cases from the source dataset TN3K and the target dataset DDTI. For the source dataset, the first row displays ultrasound images and thyroid nodule annotations; the second row presents the corresponding annotation masks. The bottom section shows the equivalent ultrasound images and annotations for the target dataset.}
  \label{fig: USdataset}
\end{figure}

Source domain: TN3K~\cite{gong2021multi}, Target domain: DDTI~\cite{pedraza2015open}, as shown in Fig.~\ref{fig: USdataset}.
TN3K is a large-scale, multi-center ultrasound dataset of thyroid nodules, comprising 3,493 images from 2,421 patients. It was collected across diverse probe brands and acquisition workflows, making it a suitable starting point for multi-source generalization. We adopt its training split of 2879 images as the source domain. DDTI, in contrast, originates from a single device with 637 images showing more concentrated sources and distinct distributions of image resolution and appearance. The two datasets differ markedly in probe models, gain, and dynamic range configurations, as well as in image texture statistics, yet they share the identical task of nodule segmentation, thereby constituting a compelling testbed for TTA in ultrasound scenarios.

\subsubsection{Pathology-based Colorectal Adenocarcinoma Tissue Segmentation Datasets}
\begin{figure}[h!]
  \centering
  \includegraphics[width=\linewidth]{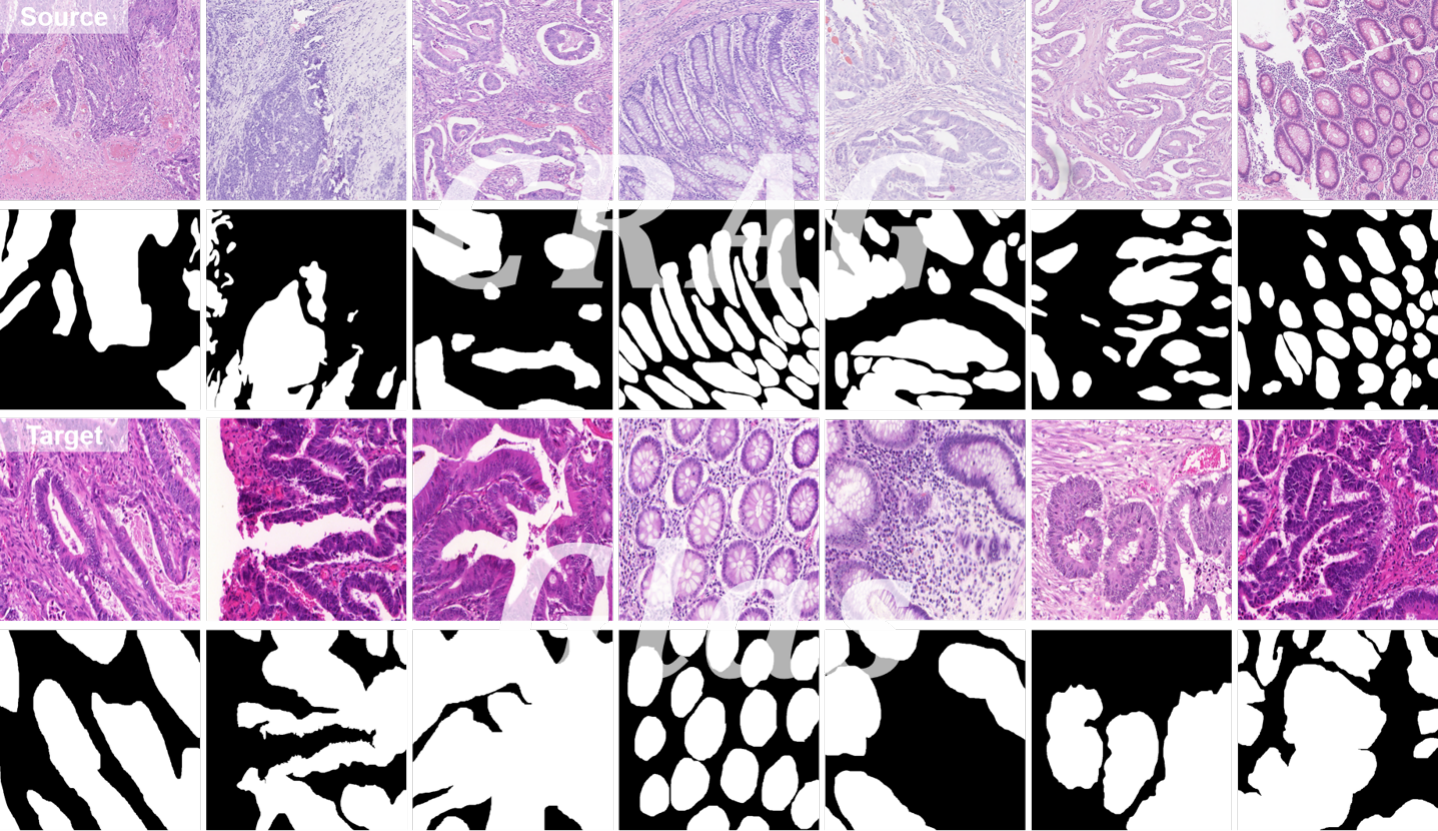}
  \caption{\textbf{Histopathological images for colorectal adenocarcinoma tissue segmentation.} Cases from the source dataset CRAG and the target dataset GlaS. For the source dataset area, the first row displays 256×256 cropped images after sliding-window processing and binary segmentation, with foreground as tumor tissue and background as normal tissue; the second row presents the corresponding annotation masks. The bottom section shows the equivalent cropped images and annotations for the target dataset.}
  \label{fig: PATHdataset}
\end{figure}

Source domain: CRAG~\cite{graham2019mild}, Target domain: GlaS~\cite{SIRINUKUNWATTANA2017489}, as shown in Fig.~\ref{fig: PATHdataset}.
CRAG contains 213 adenocarcinoma slides scanned at \(20\times\) magnification with detailed instance-level annotations, whereas GlaS comprises 165 images derived from 16 slides, reflecting pronounced variations in staining batches, scanners, magnifications, and tissue morphology. We harmonize the labels of the two datasets into binary masks with foreground and background classes, and we employ sliding-window cropping with a \(256\times 256\) size and \(50\%\) overlap to accommodate substantial resolution differences. After discarding all background patches with less than \(1\%\) foreground ratio, we obtain \(14{,}015\) image-mask pairs from CRAG and \(2{,}208\) pairs from GlaS. This dataset pair, while maintaining task consistency, exhibits substantial appearance and texture domain shifts, making it an ideal testing ground for pathology segmentation under real-world TTA settings.

\subsubsection{Dermoscopy-based Melanoma Segmentation Datasets}
\begin{figure}[h!]
  \centering
  \includegraphics[width=\linewidth]{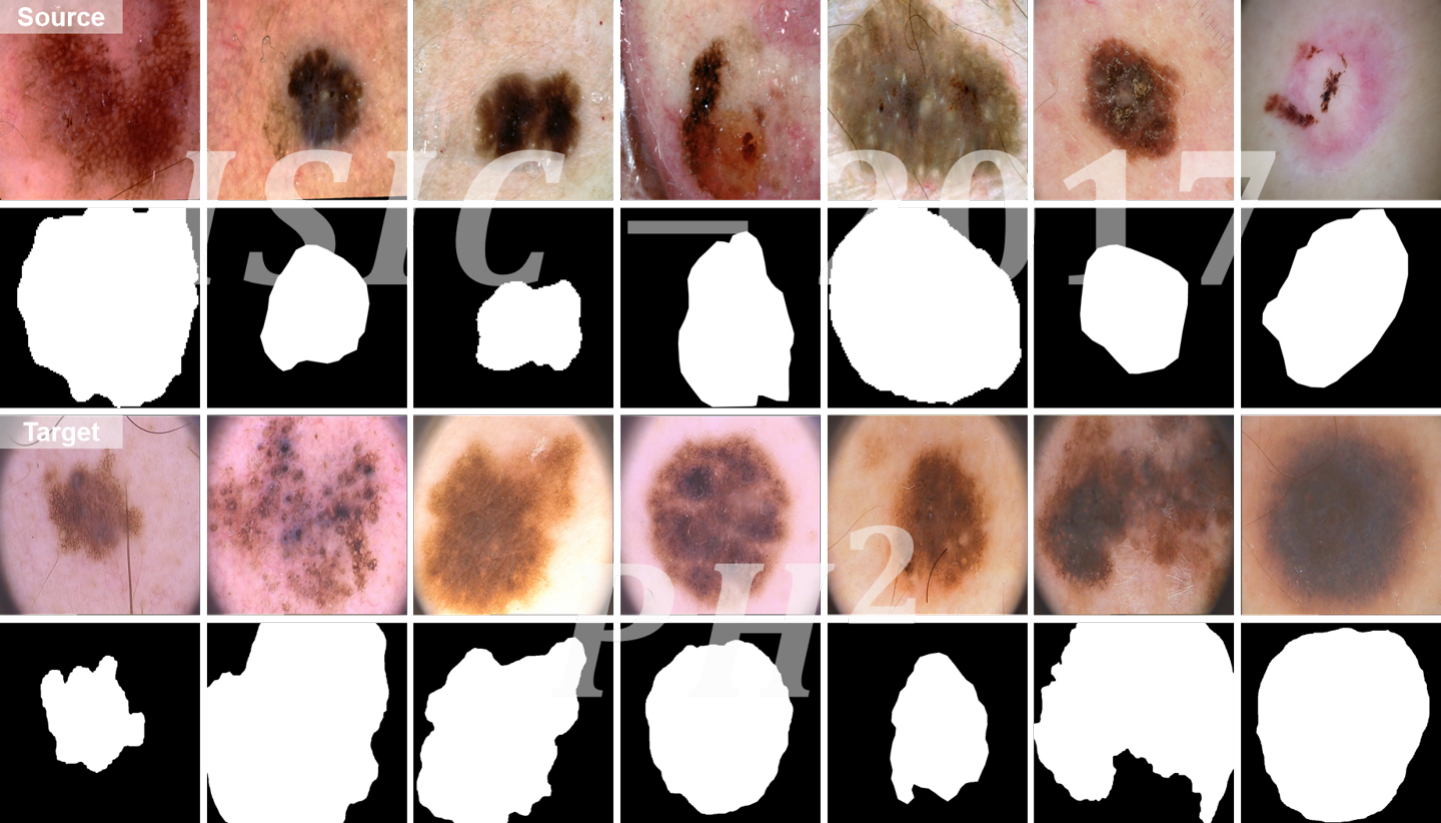}
  \caption{\textbf{Dermoscopy images for melanoma segmentation.} Cases from the source dataset ISIC-2017 and the target dataset PH². The first row displays dermoscopy images and melanoma lesion annotations; the second row presents the corresponding annotation masks. The bottom rows show the equivalent dermoscopy images and annotations for the target dataset. }
  \label{fig: DERdataset}
\end{figure}

Source domain: ISIC-2017~\cite{8363547}, Target domain: PH\textsuperscript{2}~\cite{mendoncca2015ph2}, as shown in Fig.~\ref{fig: DERdataset}.
ISIC-2017, released by the International Skin Imaging Collaboration, is a large-scale and diverse dermoscopy dataset comprising 2,000 training images, 150 validation images, and 600 test images. We use its 2,000 training images as the source domain. PH\textsuperscript{2}, collected at a hospital in Lisbon, Portugal, includes 200 dermoscopy images with meticulous annotations, obtained from a single device and a narrower patient base. The two datasets show stable differences in camera settings, illumination, skin tones, and lesion spectra, leading to pronounced inter-dataset shifts despite sharing the same task: lesion segmentation. This pair provides a benchmark to assess the benefits and risks of TTA in small target domains.

\subsubsection{Chest X-ray-based Lung Segmentation Datasets}
\begin{figure}[h!]
  \centering
  \includegraphics[width=\linewidth]{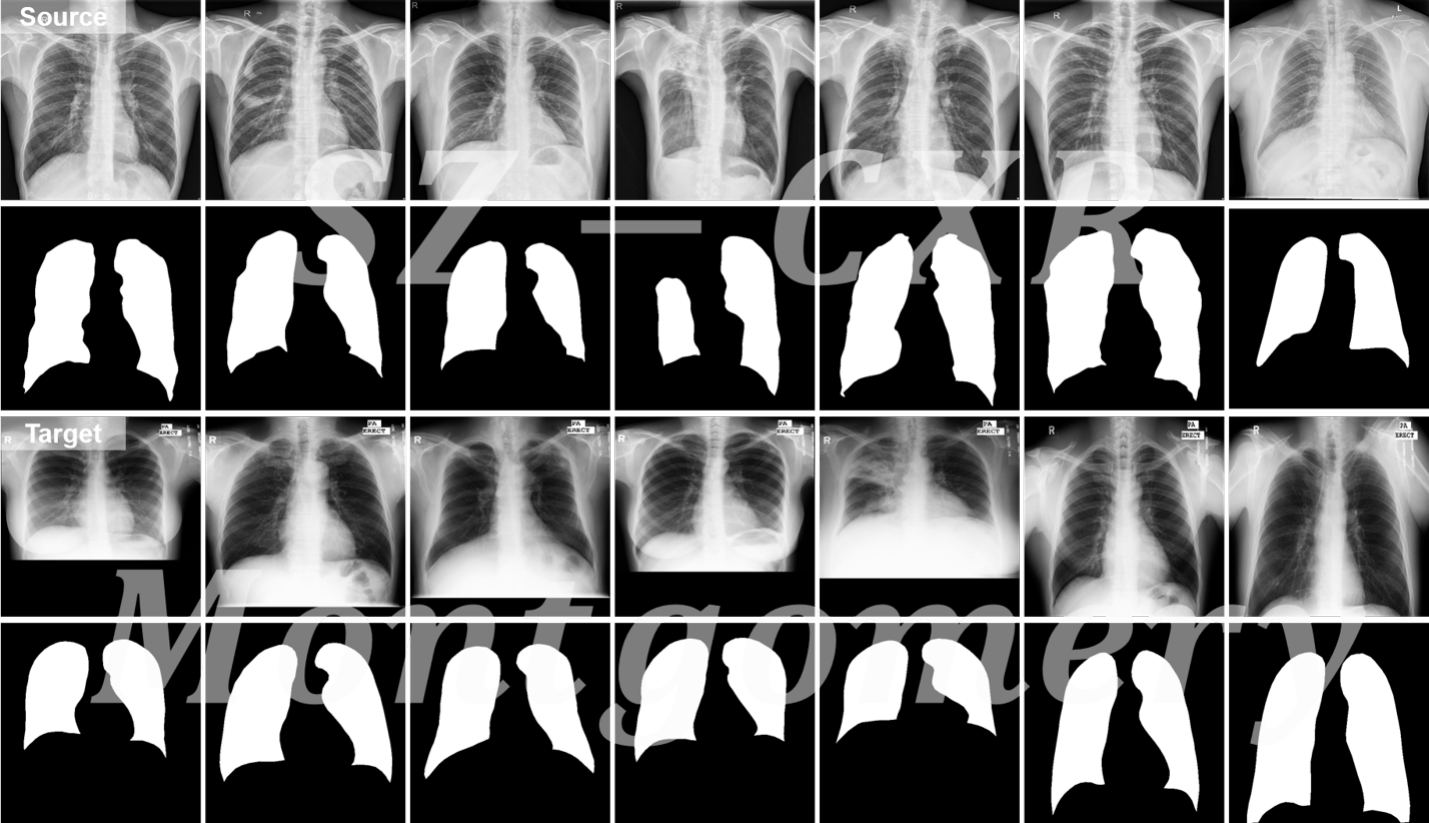}
  \caption{\textbf{Chest X-ray images for lung segmentation.} Cases from the source dataset SZ-CXR and the target dataset Montgomery County CXR Set. For the source dataset area, the first row displays posteroanterior chest radiographs and lung field annotations; the second row presents the corresponding annotation masks. The bottom section shows the equivalent chest X-ray images and annotations for the target dataset.}
  \label{fig: CXRdataset}
\end{figure}

Source domain: Shenzhen (SZ-CXR)~\cite{stirenko2018chest}, Target domain: Montgomery CXR Set~\cite{jaeger2014two}, as shown in Fig.~\ref{fig: CXRdataset}.
Source domain: Shenzhen (SZ-CXR)~\cite{stirenko2018chest}, Target domain: Montgomery CXR Set~\cite{jaeger2014two}, as shown in Fig.~\ref{fig: CXRdataset}.
The SZ-CXR dataset, collected at Shenzhen Third People’s Hospital, comprises 662 chest radiographs with lung-field masks, all of which are used as source-domain training data. The Montgomery dataset, curated by the Montgomery County Department of Health and NIH, contains 138 posteroanterior chest radiographs with lung masks, used as target test data. Pronounced differences exist in population demographics, imaging devices, and resolution distributions between Chinese and American cohorts. Under identical modality and anatomical focus, this dataset pair provides cross-national and cross-device shifts, forming a canonical setup for TTA evaluation in chest X-ray segmentation.

\subsubsection{Summary of Datasets}
Fig.~\ref{fig: datasetv1} illustrates representative examples of the seven domain-shift pairs, while Table~\ref{tab:tta_benchmark_datasets} summarizes their modalities, tasks, dimensions, and repartitioning details. Collectively, these seven source-to-target combinations adhere strictly to the principles of modality and task consistency with matched categories and annotation granularity, while simultaneously exhibiting systematic discrepancies across institutions, devices, and protocols, populations, or preparation pipelines. Moreover, they reflect the practical deployment setting characterized by large-sample source domains and small-sample target domains.

\begin{figure}[h!]
  \centering
  \includegraphics[width=\linewidth]{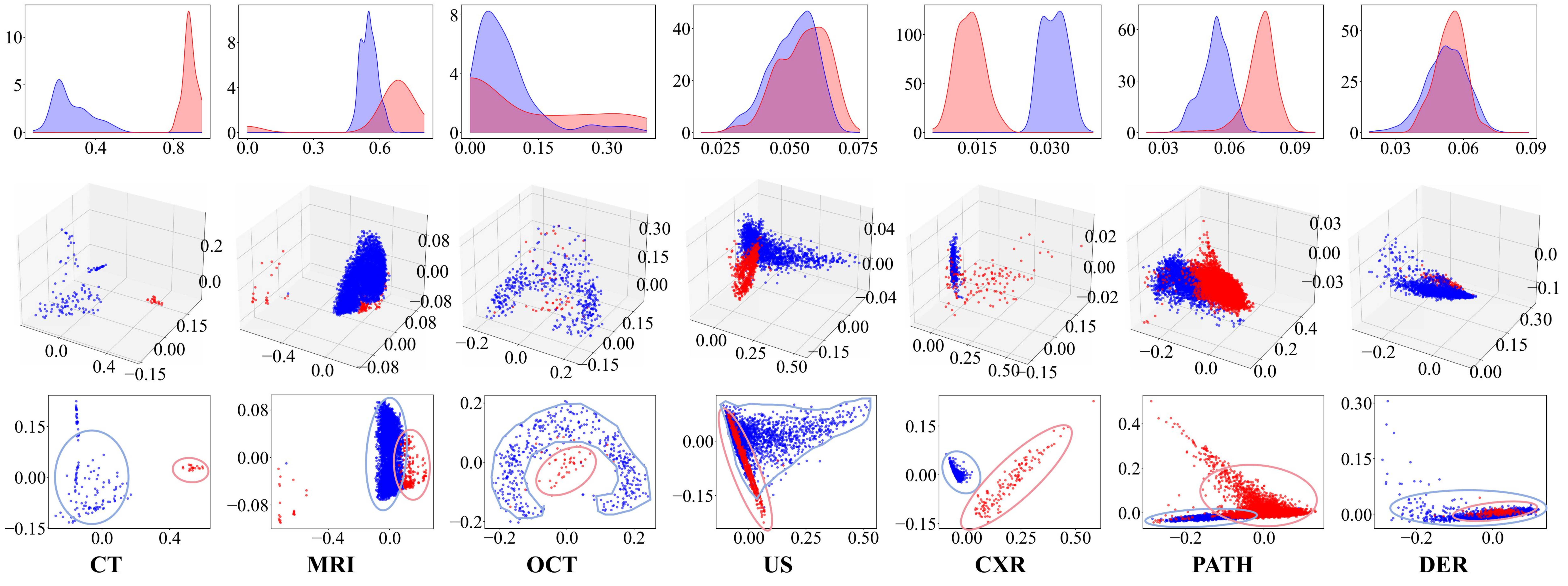}
  \caption{\textbf{Multi-perspective visualization of feature distributions across heterogeneous domains.} Row 1 illustrates the marginal frequency distribution of feature values, with the abscissa and ordinate denoting feature magnitude and frequency, respectively. Rows 2 and 3 present the spatial distribution of features in the low-dimensional embedding space obtained via dimensionality reduction. Source-domain and target-domain samples are encoded in red and blue, respectively.}
  \label{fig: Domain}
\end{figure}

\begin{table}[h!]
\centering
\resizebox{\linewidth}{!}{%
\begin{threeparttable}
\caption{\textbf{Source and target dataset pairs for the TTA benchmark in medical image segmentation.} The Category indicates the number of segmentation classes after aligning the annotation categories between the source and target domain datasets. "Binary" refers to binary segmentation tasks, while "4-Class" refers to four-class segmentation tasks, which include the background class. The Reprocess column uses \ding{51} to indicate reprocessed datasets and \ding{55} to indicate datasets used in their original form.}
\label{tab:tta_benchmark_datasets}
\scriptsize

\renewcommand{\arraystretch}{1.4} 
\begin{tabular}{>{\centering\arraybackslash}p{1.8cm}
                >{\centering\arraybackslash}p{2.8cm}
                >{\centering\arraybackslash}p{1cm}
                >{\centering\arraybackslash}p{1.3cm}
                >{\centering\arraybackslash}p{1.1cm}
                >{\centering\arraybackslash}p{0.8cm}
                >{\centering\arraybackslash}p{1.1cm}
                >{\centering\arraybackslash}p{1.2cm}}
\hline
  \rowcolor[HTML]{D8D6C0}
\textbf{Modal} & \textbf{Dataset} & \textbf{Domain} & \textbf{Category} & \textbf{Quantity} & \textbf{Year} & \textbf{Reprocess} & \textbf{Source} \\
\hline
\multirow{2}{*}{\textbf{MRI}} & BraTS-GLI2024~\cite{de2024brats} & Source & 4-Class & 1k$\sim$2k & 2024 & \textcolor{red}{\ding{55}} & \href{https://www.synapse.org/Synapse:syn59059776}{Link} \\
\cdashline{2-8} 
& BraTS-SSA~\cite{adewole2023brain} & Target & 4-Class & 60 & 2023 & \textcolor{red}{\ding{55}} & \href{https://www.synapse.org/\#!Synapse:syn51514109}{Link} \\
\midrule
\multirow{2}{*}{\textbf{CT}} & LiTS~\cite{bilic2023liver} & Source & Binary & $<$0.5k & 2017 & \textcolor{green}{\ding{51}} & \href{https://competitions.codalab.org/competitions/17094}{Link} \\
\cdashline{2-8}
& 3D-IRCADB~\cite{soler20103d} & Target & Binary & 20 & 2010 & \textcolor{green}{\ding{51}} & \href{https://cloud.ircad.fr/index.php/s/JN3z7EynBiwYyjy/download}{Link} \\
\midrule
\multirow{2}{*}{\textbf{DER}} & ISIC-2017~\cite{8363547} & Source & Binary & 2k$\sim$3k & 2017 & \textcolor{red}{\ding{55}} & \href{https://challenge.isic-archive.com/data/\#2017}{Link} \\
\cdashline{2-8}
& PH\textsuperscript{2}~\cite{mendoncca2015ph2} & Target & Binary & $<$0.5k & 2015 & \textcolor{red}{\ding{55}} & \href{https://www.dropbox.com/s/k88qukc20ljnbuo/PH2Dataset.rar}{Link} \\
\midrule
\multirow{2}{*}{\textbf{Ultrasound}} & TN3K~\cite{gong2021multi} & Source & Binary & 2k$\sim$3k & 2021 & \textcolor{red}{\ding{55}} & \href{https://github.com/haifangong/TRFE-Net-for-thyroid-nodule-segmentation}{Link} \\
\cdashline{2-8}
& DDTI~\cite{pedraza2015open} & Target & Binary & 0.5k$\sim$1k & 2015 & \textcolor{red}{\ding{55}} & \href{http://cimalab.intec.co/applications/thyroid/}{Link} \\
\midrule
\multirow{2}{*}{\textbf{X-Ray}} & SZ-CXR~\cite{stirenko2018chest} & Source & Binary & 0.5k$\sim$1k & 2018 & \textcolor{red}{\ding{55}} & \href{https://www.kaggle.com/datasets/raddar/tuberculosis-chest-xrays-shenzhen}{Link} \\
\cdashline{2-8}
& Montgomery~\cite{jaeger2014two} & Target & Binary & $<$0.5k & 2021 & \textcolor{red}{\ding{55}} & \href{https://openi.nlm.nih.gov/imgs/collections/NLM-MontgomeryCXRSet.zip}{Link} \\
\midrule
\multirow{2}{*}{\textbf{Fundus}} & RIGA$\text{$\text{+}$}$(MES)~\cite{hu2022domain}& Source & Binary & $<$0.5k & 2021 & \textcolor{green}{\ding{51}} & \href{https://github.com/mohaEs/RIGA-segmentation-masks/raw/main/RIGA_masks.zip}{Link} \\
\cdashline{2-8}
& RIGA$\text{$\text{+}$}$(MB)~\cite{hu2022domain}& Target & Binary & $<$0.5k & 2021 & \textcolor{green}{\ding{51}} & \href{https://github.com/mohaEs/RIGA-segmentation-masks/raw/main/RIGA_masks.zip}{Link} \\
\midrule
\multirow{2}{*}{\textbf{Histopathology}} & CRAG~\cite{graham2019mild}& Source & Binary & $<$0.5k & 2019 & \textcolor{green}{\ding{51}} & \href{https://github.com/XiaoyuZHK/CRAG-Dataset_Aug_ToCOCO}{Link} \\
\cdashline{2-8}
& Glas~\cite{SIRINUKUNWATTANA2017489} & Target & Binary & $<$0.5k & 2017 & \textcolor{green}{\ding{51}} & \href{https://academictorrents.com/details/208814dd113c2b0a242e74e832ccac28fcff74e5}{Link} \\
\bottomrule
\end{tabular}
\end{threeparttable}
}
\end{table}

The evaluation is organized into two categories. For MRI brain tumor segmentation, the metrics are computed separately for ET, CT, and WT regions in accordance with the BraTS evaluation protocol, allowing performance to be captured across distinct tumor components. For other modalities such as CT, OCT, and US, which predominantly involve binary segmentation of foreground and background, the metrics are computed on the foreground regions to assess both target recognition and boundary stability.
This unified and fine-grained metric design enables a more precise characterization of the adaptability and robustness of different TTA methods across diverse modalities and tasks.

\subsection{Implementation Details}
\label{subsec3.2}
\subsubsection{Unified Pretraining for Medical Image Segmentation}
To eliminate confounding factors introduced by backbone differences, all source domain models are pretrained using a standard UNet architecture. The 2D and 3D configurations, including input sizes, encoder depths, channel widths, output channels, and BraTS-specific settings, are listed in Table~\ref{tab:unet_config}. These configurations are consistently applied across modalities and remain reusable, with only minimal adjustments required for dimensionality in 2D or 3D, as well as the number of classes. This ensures that TTA comparisons have equivalent starting points.

\begin{table}[h!]
  \centering
  \caption{\textbf{Unified architectural configurations of UNet backbones for pretraining. }
UNet2d and UNet3d share the same encoder depth and base width \(C_0\), while the BraTS-specific UNet3d variant uses four input and output channels for multi-class tumor segmentation.}
  \vspace{0.5em}
  \label{tab:unet_config}
  \resizebox{\textwidth}{!}{%
   \begin{tabular}{lccc}
   \hline
     \rowcolor[HTML]{D8D6C0}
    \textbf{Components} & \textbf{UNet2d} & \textbf{UNet3d} & \textbf{UNet3d (BraTS)} \\
    \hline
    Input size          & \( 256 \times 256 \)        & \( 128 \times 128 \times 128 \) & \( 128 \times 128 \times 128 \) \\
    Input channel       & 1                           & 1                              & 4 \\
    
    Output channel      & 2                           & 2                              & 4 \\
    Encoder blocks      & 5                           & 5                              & 5 \\
    Block depth \( D \) & 256                         & 256                            & 256 \\
    Basic width \( C_0 \)& 32                         & 32                             & 32 \\
    Channels of encoders& \( C_0 \)-\( 2C_0 \)-\( 4C_0 \)-\( 8C_0 \)-\( 16C_0 \) & \( C_0 \)-\( 2C_0 \)-\( 4C_0 \)-\( 8C_0 \)-\( 16C_0 \) & \( C_0 \)-\( 2C_0 \)-\( 4C_0 \)-\( 8C_0 \)-\( 16C_0 \) \\
    \bottomrule
  \end{tabular}
  }
\end{table}

\subsubsection{Unified Implementation of the TTA for Medical Image Segmentation}
To enable a fair comparison among various TTA approaches, we adopt UNet as the unified, pre-trained backbone, without altering its network topology. Adjustments are made solely during testing, following each method’s prescribed strategy to update permissible parameters, statistics, and adaptation modules. All methods share the following standard settings: (1) identical source domain initialization and data preprocessing. (2) identical target domain input sequences with streaming or mini-batch configurations. (3) a unified evaluation interface and standardized metric system. (4) a strict prohibition on accessing source domain data or labels, thereby preventing both explicit and implicit leakage.

\section{Experiments}

To systematically and comprehensively evaluate TTA strategies for medical image segmentation, we benchmarked twenty representative methods across seven imaging modalities. These methods encompass four major paradigms: Input-level Transformation, Feature-level Alignment, Output-level Regularization, and Prior Estimation, each with five representative methods. Such broad coverage ensures that the experimental findings possess strong generalizability and reference value across diverse imaging modalities and segmentation tasks. It is worth noting that many of these methods were initially designed and optimized for a specific data dimensionality. For instance, FSM~\cite{yang2022source} primarily targets 2D images, whereas DANN~\cite{ganin2016domain} is tailored for 3D data. Given that our benchmark encompasses both 2D and 3D medical images, we extended these single-dimensional methods to support both settings, enabling fair and comprehensive cross-dimensional evaluation. Tables~\ref {tab:tta_basic} and~\ref {tab:tta_cross_dimensional} summarize all investigated methods, along with the key principles behind their dimensional extensions.

\begin{table*}[h!]
  \centering
  \caption{\textbf{Implementation details of compared TTA methods.} Covering their paradigms, original modalities, and dimensions, with paradigms marked by symbols: \textcolor{black}{$\spadesuit$} for Input-level Translation, \textcolor{green}{$\clubsuit$} for Feature-level Alignment, \textcolor{red}{$\varheartsuit$} for Output-level Regularization, and \textcolor{blue}{$\vardiamondsuit$} for Prior Estimation.}
  \vspace{6pt}
  \label{tab:tta_basic}
  \resizebox{\textwidth}{!}{%
  \begin{tabular}{l l c c}
    \hline
     \rowcolor[HTML]{D8D6C0}
    \textbf{Method} & \textbf{Paradigm} & \textbf{Original Modal} & \textbf{Original Imension} \\
    \hline

    \textcolor{black}{$\spadesuit$} SFDA-FSM~\cite{yang2022source} & Input-level Translation & Endoscope & 2D \\
    \textcolor{black}{$\spadesuit$} RSA~\cite{zeng2024reliable} & Input-level Translation & MRI & 2D \\
    \textcolor{black}{$\spadesuit$} DL-TTA~\cite{yang2022dltta} & Input-level Translation & PATH & 2D \\
    \textcolor{black}{$\spadesuit$} STDR~\cite{wang2024dual} & Input-level Translation & MRI & 2D \\
    \textcolor{black}{$\spadesuit$} AIF-SFDA~\cite{li2025aif} & Input-level Translation & OCT & 2D \\
    \midrule

    \textcolor{green}{$\clubsuit$} DANN~\cite{ganin2016domain} & Feature-level Alignment & MRI & 3D \\
    \textcolor{green}{$\clubsuit$} Testfit~\cite{zhang2024testfit} & Feature-level Alignment & CT/PATH & General \\
    \textcolor{green}{$\clubsuit$} GraTa~\cite{chen2025gradient} & Feature-level Alignment & OCT & 2D \\
    \textcolor{green}{$\clubsuit$} UDA-MIMA~\cite{hu2024unsupervised} & Feature-level Alignment & MRI/CT & 3D \\
    \textcolor{green}{$\clubsuit$} DeTTA~\cite{wen2024denoising} & Feature-level Alignment & CT & 2D \\
    \midrule

    \textcolor{red}{$\varheartsuit$} TENT~\cite{wang2021tent} & Output-level Regularization & General Image & General \\
    \textcolor{red}{$\varheartsuit$} DG-TTA~\cite{weihsbach2023dg} & Output-level Regularization & MRI/CT & 3D \\
    \textcolor{red}{$\varheartsuit$} SaTTCA~\cite{li2023scale} & Output-level Regularization & CT & 3D \\
    \textcolor{red}{$\varheartsuit$} UPL-SFDA~\cite{wu2023upl} & Output-level Regularization & CMR/MRI & General \\
    \textcolor{red}{$\varheartsuit$} SmaRT~\cite{Wang2025SmaRTSR} & Output-level Regularization & MRI & 3D \\
    \midrule

    \textcolor{blue}{$\vardiamondsuit$} PASS~\cite{zhang2024pass} & Prior Estimation & OCT & 2D \\
    \textcolor{blue}{$\vardiamondsuit$} ProSFDA~\cite{hu2025source} & Prior Estimation & OCT & 2D \\
    \textcolor{blue}{$\vardiamondsuit$} AdaMI~\cite{bateson2022test} & Prior Estimation & MRI/CT & 3D \\
    \textcolor{blue}{$\vardiamondsuit$} ExploringTTA~\cite{omolegan2025exploring} & Prior Estimation & US & 3D \\
    \textcolor{blue}{$\vardiamondsuit$} VPTTA~\cite{chen2024each} & Prior Estimation & OCT & 2D \\
    \bottomrule
  \end{tabular}%
  }
\end{table*}

\begin{table*}[h!]
  \centering
  \caption{\textbf{Implementation details of compared TTA methods.} Covering method and cross-dimensional extension, marked by symbols: \textcolor{black}{$\spadesuit$} for Input-level Translation, \textcolor{green}{$\clubsuit$} for Feature-level Alignment, \textcolor{red}{$\varheartsuit$} for Output-level Regularization, \textcolor{blue}{$\vardiamondsuit$} for Prior Estimation.}
  \vspace{6pt}
  \label{tab:tta_cross_dimensional}
  \resizebox{\textwidth}{!}{%
  \footnotesize 
  \begin{tabular}{l p{12.5cm}} 
     \hline
    \rowcolor[HTML]{D8D6C0}
    \textbf{Method} & \textbf{Cross-Dimensional Method Extension} \\
    \hline
    \textcolor{black}{$\spadesuit$} SFDA-FSM~\cite{yang2022source} & 2D to 3D: The \texttt{fft2}/\texttt{ifft2} operations are replaced with \texttt{fftn}/\texttt{ifftn}. The low-frequency square replacement is upgraded to a cubic structure. Case-level mask sharing and sliding-window fusion are introduced to ensure inter-slice consistency. \\
     \rowcolor[HTML]{ECEADE}
    \textcolor{black}{$\spadesuit$} RSA~\cite{zeng2024reliable} & 2D to 3D: The diffusion U-Net is extended to a 3D architecture, with noise and timestep embeddings augmented along the depth dimension. Training is on volumetric patches, and inference incorporates consistency constraints to ensure coherence. \\
    \textcolor{black}{$\spadesuit$} DL-TTA~\cite{yang2022dltta} & 2D to 3D: The reconstructor is replaced by a 3D autoencoder or generator. Patch-level reconstructions are normalized and aggregated into volumetric outputs, which are then fed into the pretrained segmentation network. \\
    \rowcolor[HTML]{ECEADE}
    \textcolor{black}{$\spadesuit$} STDR~\cite{wang2024dual} & 2D to 3D: Style and statistical alignment are extended from image level to volume level, with statistics shared to ensure inter-slice consistency and correspondence. \\
    \textcolor{black}{$\spadesuit$} AIF-SFDA~\cite{li2025aif} & 2D to 3D: Adaptive frequencies are computed and aligned in the 3D spectrum, frequency-band cubes are mixed and stabilized through case-level normalization to maintain volumetric appearance consistency. \\
     \rowcolor[HTML]{ECEADE}
    \textcolor{green}{$\clubsuit$} DANN~\cite{ganin2016domain} & 3D to 2D: The domain discriminator and feature extractor are downgraded from 3D to 2D convolutions. Domain labels are shared across slices, with case-level feature aggregation and gradient reversal applied. \\
    \textcolor{green}{$\clubsuit$} Testfit~\cite{zhang2024testfit} & None: Native support for both 2D and 3D. \\
     \rowcolor[HTML]{ECEADE}
    \textcolor{green}{$\clubsuit$} GraTa~\cite{chen2025gradient} & 2D to 3D: Gradient alignment across multiple enhanced branches on the 3D volume, update normalized affine parameters, and accumulate case-level stable direction estimates. \\
    \textcolor{green}{$\clubsuit$} UDA-MIMA~\cite{hu2024unsupervised} & 3D to 2D: Mutual information estimation is reduced from volumetric to slice level. A 2.5D neighborhood mechanism is introduced to preserve depth context. \\
    \rowcolor[HTML]{ECEADE}
    \textcolor{green}{$\clubsuit$} DeTTA~\cite{wen2024denoising} & 2D to 3D: Single-slice masks are expanded into patch-level random masks. Multiple generated inputs are fused volumetrically, with case-level parameter resets preventing cross-domain interference. \\
    \textcolor{red}{$\varheartsuit$} TENT~\cite{wang2021tent} & None: Natively compatible with both 2D and 3D settings. \\
   \rowcolor[HTML]{ECEADE}
    \textcolor{red}{$\varheartsuit$} DG-TTA~\cite{weihsbach2023dg} & 3D to 2D: Dual-branch consistency is reduced from volumetric to slice level. EMA teachers and consistency masks are shared to enforce smoothness across slices. \\
    \textcolor{red}{$\varheartsuit$} SaTTCA~\cite{li2023scale} & 3D to 2D: 3D click prompts are projected into 2D Gaussian heatmaps. Radii are rescaled based on the pixel spacing and jointly optimized with entropy regularization. \\
    \rowcolor[HTML]{ECEADE}
    \textcolor{red}{$\varheartsuit$} UPL-SFDA~\cite{wu2023upl} & None: Natively compatible with both 2D and 3D settings. \\
    \textcolor{red}{$\varheartsuit$} SmaRT~\cite{Wang2025SmaRTSR} & 3D to 2D: The loss function is adapted from a $3 \times 3 \times 3$ region to a $3 \times 3$ region, and the connectivity criterion is redefined for 2D scenarios. \\
     \rowcolor[HTML]{ECEADE}
    \textcolor{blue}{$\vardiamondsuit$} PASS~\cite{zhang2024pass} & 2D to 3D: Shape and style prompts are elevated to 3D tokens, shared across three orthogonal planes, and injected into the volumetric encoder through cross-attention. \\
    \textcolor{blue}{$\vardiamondsuit$} ProSFDA~\cite{hu2025source} & 2D to 3D: 3D class prototypes are constructed from high-confidence voxels. Volumetric prototype recalibration and logits reweighting are performed, followed by case-level aggregation to enhance stability. \\
     \rowcolor[HTML]{ECEADE}
    \textcolor{blue}{$\vardiamondsuit$} AdaMI~\cite{bateson2022test} & 3D to 2D: The original network processes 2D slices from 3D images and can be directly adapted for inference on standalone 2D inputs. \\
    \textcolor{blue}{$\vardiamondsuit$} ExploringTTA~\cite{omolegan2025exploring} & 3D to 2D: Structural and topological priors are reduced from volumetric to slice-level boundary and connectivity priors, with inter-slice smoothing ensuring consistency. \\
    \rowcolor[HTML]{ECEADE}
    \textcolor{blue}{$\vardiamondsuit$} VPTTA~\cite{chen2024each} & 2D to 3D: The 2D FFT is replaced with 3D FFT. Batch normalization statistics are extended to 3D, and prompts are redesigned as 3D tensors. \\
    \bottomrule
  \end{tabular}%
  }
\end{table*}
\subsection{Overall Performance Analysis}
To assess domain quality gaps and distribution shifts, we split the source domain into training and validation at an 8:2 ratio for in-domain evaluation and used the entire target domain for out-of-domain testing. Under a unified UNet backbone, all four TTA paradigms achieved higher average segmentation accuracy than the no-adaptation baseline across modalities, confirming the practical value of TTA for medical image segmentation. As summarized in Table~\ref{tab:tta_intra}, Input-level Transformation methods achieved the highest Dice in most modalities. In ultrasound, Prior Estimation was slightly better; yet, Input-level methods still outperformed the remaining paradigms by a clear margin. From the perspective of HD95, a boundary-sensitive complement to Dice, the top performers did not concentrate within a single paradigm. In OCT, pathology, and MRI, the leading methods often came from different categories. These patterns suggest that appearance alignment effectively reduces intensity and texture discrepancies but cannot, by itself, resolve complex deformations and boundary noise. We also observe task-dependent failures: certain methods degrade on specific segmentation tasks and can even underperform the no-adaptation baseline, revealing limited generalization and underscoring the need for careful paradigm selection.

\begin{table}[h!]
  \centering
  \caption{\textbf{Paradigm taxonomy and applicability lineage across modalities.} We categorize TTA methods by their locus of operation and, using evaluations across modalities, organs, and tasks, derive a \emph{lineage map} that highlights effective, partially effective, and ineffective regimes, with effectiveness defined as Dice increasing and HD95 decreasing. The table reports Dice and HD95 with significance at $p<0.05$. }
  \vspace{0.5em}
  \label{tab:tta_intra}
  \resizebox{\textwidth}{!}{%
  \begin{tabular}{l >{\columncolor{red!10}}c c >{\columncolor{red!10}}c c >{\columncolor{red!10}}c c >{\columncolor{red!10}}c c >{\columncolor{red!10}}c c >{\columncolor{red!10}}c c >{\columncolor{red!10}}c c >{\columncolor{red!10}}c c}
    \toprule
    \multirow{2}{*}{\textbf{Method}} & \multicolumn{2}{c}{\textbf{MRI}} & \multicolumn{2}{c}{\textbf{CT}} & \multicolumn{2}{c}{\textbf{OCT}} & \multicolumn{2}{c}{\textbf{US}} & \multicolumn{2}{c}{\textbf{PATH}} & \multicolumn{2}{c}{\textbf{DER}} & \multicolumn{2}{c}{\textbf{CXR}} \\
    \cmidrule(lr){2-3} \cmidrule(lr){4-5} \cmidrule(lr){6-7} \cmidrule(lr){8-9} \cmidrule(lr){10-11} \cmidrule(lr){12-13} \cmidrule(lr){14-15}
    & \textbf{Dice$\uparrow$} & \textbf{HD95$\downarrow$} & \textbf{Dice$\uparrow$} & \textbf{HD95$\downarrow$} & \textbf{Dice$\uparrow$} & \textbf{HD95$\downarrow$} & \textbf{Dice$\uparrow$} & \textbf{HD95$\downarrow$} & \textbf{Dice$\uparrow$} & \textbf{HD95$\downarrow$} & \textbf{Dice$\uparrow$} & \textbf{HD95$\downarrow$} & \textbf{Dice$\uparrow$} & \textbf{HD95$\downarrow$} \\
    \midrule
    \textbf{Intra-Domain} & 0.7049 & 3.33 & 0.8949 & 1.63 & 0.9550 & 0.41 & 0.7215 & 31.01 & 0.8360 & 32.66 & 0.9280 & 2.33 & 0.9557 & 1.02 \\
    \midrule
    \textbf{Target-Domain (w/o TTA)} & 0.5951 & 23.04 & 0.6595 & 13.91 & 0.8951 & 13.32 & 0.4811 & 62.78 & 0.8026 & 33.66 & 0.8532 & 13.78 & 0.9093 & 5.15 \\
    \midrule
    \textcolor{black}{$\spadesuit$} \textbf{Input-level Translation} & 0.6411 & 13.72 & 0.7156 & 10.32 & 0.9075 & 9.77 & 0.5757 & 49.26 & 0.8107 & 31.63 & 0.8672 & 12.68 & 0.9308 & 3.36 \\
    \textcolor{green}{$\clubsuit$} \textbf{Feature-level Alignment} & 0.6269 & 9.67 & 0.6874 & 13.56 & 0.9000 & 10.28 & 0.5539 & 50.83 & 0.7930 & 35.39 & 0.8404 & 15.59 & 0.9194 & 4.69 \\
     \textcolor{red}{$\varheartsuit$} \textbf{Output-level Regularization} & 0.6399 & 7.98 & 0.7240 & 14.24 & 0.9051 & 8.02 & 0.5235 & 58.03 & 0.8008 & 32.14 & 0.8606 & 13.92 & 0.9284 & 3.87 \\
    \textcolor{blue}{$\vardiamondsuit$} \textbf{Prior Estimation} & 0.6395 & 6.96 & 0.6969 & 14.01 & 0.8981 & 11.17 & 0.5776 & 46.95 & 0.7981 & 33.98 & 0.8526 & 15.78 & 0.9285 & 3.66 \\
    \bottomrule
  \end{tabular}%
  }
\end{table}

More specifically, Tables~\ref {tab:oct} through~\ref{tab:ct} report the performance of twenty methods across seven imaging modalities: MRI, CT, OCT, US, PATH, DER, and CXR. While the mean performance of each paradigm exceeds the non-adaptive baseline, within each paradigm, some methods drop below the baseline on specific modality-task pairs, revealing the limits of these TTA methods and their inability to generalize uniformly across medical segmentation problems. Overall, Input-level Transformation shows the most outstanding stability and reproducibility across modalities. For example, SFDA-FSM achieves the top Dice score in multiple modalities, including MRI, CXR, DER, OCT, PATH, and US. Although it does not lead in CT, it still surpasses the non-adaptive baseline without TTA. On average, SFDA-FSM delivers an absolute Dice increase of 0.118, with the most significant gain observed in the US modality, where the Dice rises by approximately 0.35, and a steady improvement in CT, where the increase is close to 0.12. In a similar vein, RSA and TestFit provide consistent gains across several modalities. While less stable than FSM, they achieve top or near-top HD95 on boundary-sensitive tasks such as CXR and DER, highlighting the effectiveness of the strategy that first aligns appearance and then performs inference.

\subsection{TTA Under Mild Domain Shift}

In tasks characterized by relatively mild domain shifts, such as OCT, PATH, DER, and CXR, Input-level methods generally achieved superior results, often excelling across both Dice and HD95 metrics. In addition, Prior Estimation methods exhibited advantages in scenarios with stable structural priors, indicating that anatomical regularities can enhance model adaptability.

\subsubsection{Optical Coherence Tomography  Modality}

\begin{figure}[tbp]
  \centering
  \includegraphics[width=\linewidth]{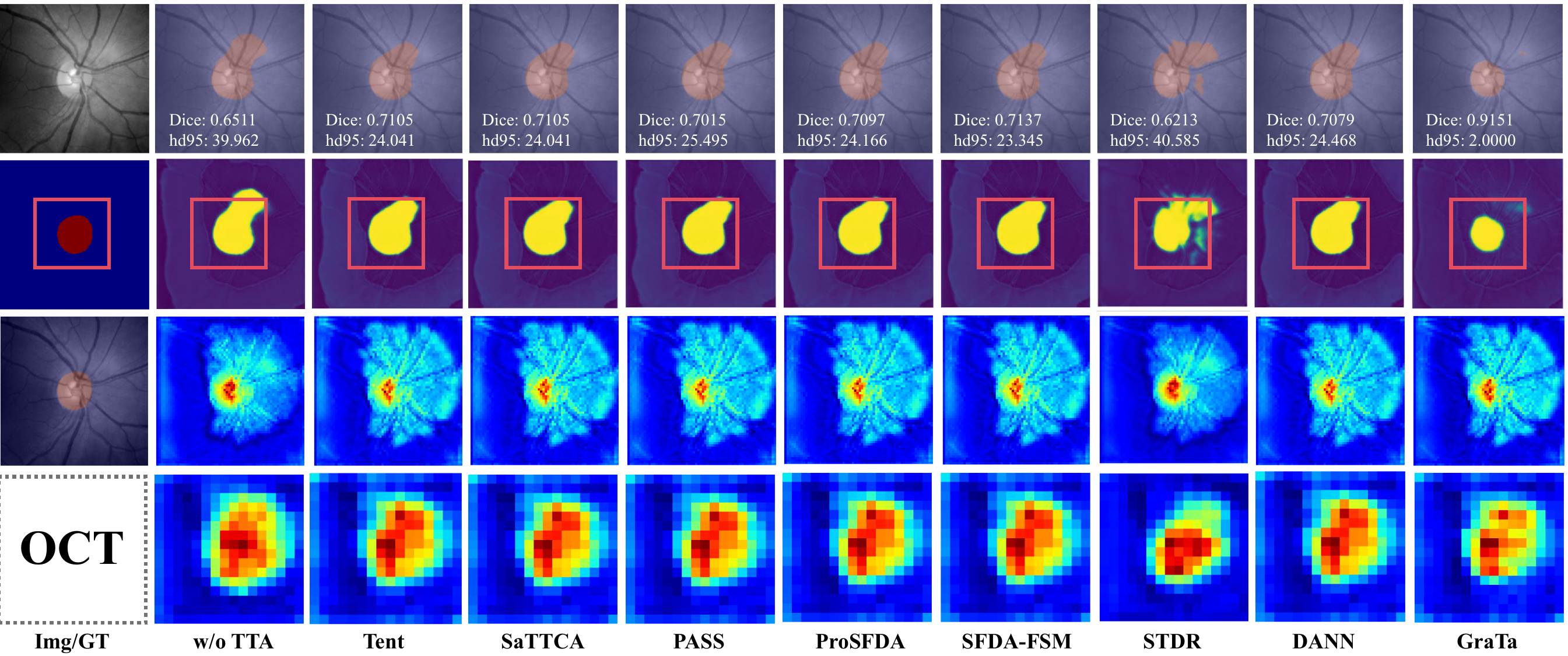}
  \caption{\textbf{Visual comparison of TTA efficacy across different methods on OCT datasets.} Rows 1 and 2 illustrate the segmentation outcomes. Rows 3 and 4 depict changes in features extracted by the Encoder after TTA. GT denotes the ground-truth.}
  \label{fig: OCT_seg}
\end{figure}

As shown in Table~\ref{tab:oct}, in the OCT task, 15 out of 20 methods outperformed the baseline Dice score of 0.8951, and 17 improved the HD95. Among them, SFDA-FSM achieved the best performance, with Dice increasing to 0.9319, a 4.1\% gain, and HD95 decreasing to 9.87, a 3.45\% reduction, indicating that appearance and frequency-domain alignment effectively alleviate intensity and texture discrepancies. In contrast, UPL-SFDA excelled in boundary accuracy, with HD95 reduced to 6.40, a 52 percent decrease, demonstrating the efficacy of Output-level Regularization in suppressing coarse deviations in predictive distributions. Feature alignment methods, such as UDA-MIMA and TestFit, and partially mismatched Prior Estimation approaches, such as VPTTA, suffer from performance degradation in this modality due to noisy pseudo-labels, suboptimal learning rates, and prior misalignment, which lead to oversmoothing and boundary degradation. Fig.~\ref{fig: OCT_seg} illustrates the effects of various TTA methods on the OCT task, as well as their influence on the encoder. Most models demonstrated positive outcomes.


\begin{table}[h!]
  \centering
  \caption{\textbf{Comprehensive performance comparison of TTA methods on OCT Fundus Image Segmentation.} Results colored in \colorbox{red!30}{\makebox[1.3em][c]{\raisebox{0pt}[0.5em][0em]{red}}}, \colorbox{orange!30}{\makebox[2.8em][c]{\raisebox{0pt}[0.5em][0em]{orange}}}, and \colorbox{yellow!30}{\makebox[2.5em][c]{\raisebox{0pt}[0.5em][0em]{yellow}}} denote the best, second-best, and third-best performances, respectively. Rows with a \colorbox{gray!30}{\makebox[1.6em][c]{\raisebox{0pt}[0.5em][0em]{gray}}} background indicate Dice values lower than that of the Target-Domain(w/o TTA). Symbols denote paradigms: \textcolor{black}{\(\spadesuit\)} for Input-level Translation, \textcolor{green}{\(\clubsuit\)} for Feature-level Alignment, \textcolor{red}{\(\varheartsuit\)} for Output-level Regularization, \textcolor{blue}{\(\vardiamondsuit\)} for Prior Estimation. All improvements are statistically significant at \( p < 0.05 \).}
  \vspace{0.5em}
  \label{tab:oct}
  \resizebox{\textwidth}{!}{%
  \begin{tabular}{lccccc} 
    \arrayrulecolor{black}\hline
    \rowcolor[HTML]{D8D6C0} 
    \textbf{Method} & \textbf{Dice $\uparrow$} & \textbf{HD95 $\downarrow$} & \textbf{JI $\uparrow$} & \textbf{Sen $\uparrow$} & \textbf{PPV $\uparrow$} \arraybackslash \\
    \arrayrulecolor{black}\hline
    \textbf{Intra-Domain}  & \textbf{0.9550{\scriptsize\textcolor{customPurple}{$\pm$0.0007}}} & \textbf{0.41{\scriptsize\textcolor{customPurple}{$\pm$0.01}}} & \textbf{0.9145{\scriptsize\textcolor{customPurple}{$\pm$0.0012}}} & \textbf{0.9607{\scriptsize\textcolor{customPurple}{$\pm$0.0033}}} & \textbf{0.9512{\scriptsize\textcolor{customPurple}{$\pm$0.0023}}}\\
    \arrayrulecolor{black}\hline
    \textbf{Target-Domain(w/o TTA)} & \textbf{0.8951{\scriptsize\textcolor{customPurple}{$\pm$0.0951}}} & \textbf{13.32{\scriptsize\textcolor{customPurple}{$\pm$28.89}}} & \textbf{0.8217{\scriptsize\textcolor{customPurple}{$\pm$0.1392}}} & \textbf{0.9611{\scriptsize\textcolor{customPurple}{$\pm$0.0387}}} & \textbf{0.8529{\scriptsize\textcolor{customPurple}{$\pm$0.1516}}}\\
    \arrayrulecolor{black}\hline
    \textcolor{black}{$\spadesuit$} RSA (2021, MICCAI)~\cite{zeng2024reliable}          & 0.8981{\scriptsize\textcolor{customPurple}{$\pm$0.0879}} & 13.01{\scriptsize\textcolor{customPurple}{$\pm$23.45}} & 0.8362{\scriptsize\textcolor{customPurple}{$\pm$0.1303}} & 0.9447{\scriptsize\textcolor{customPurple}{$\pm$0.0531}} & 0.8621{\scriptsize\textcolor{customPurple}{$\pm$0.1329}} \\
    
    \rowcolor[HTML]{F0F0F0}
    \textcolor{black}{$\spadesuit$} DL-TTA (2022, TMI)~\cite{yang2022dltta}       & 0.8927{\scriptsize\textcolor{customPurple}{$\pm$0.1033}} & 10.01{\scriptsize\textcolor{customPurple}{$\pm$14.25}} & 0.8247{\scriptsize\textcolor{customPurple}{$\pm$0.1441}} & 0.9523{\scriptsize\textcolor{customPurple}{$\pm$0.0415}} & 0.8583{\scriptsize\textcolor{customPurple}{$\pm$0.1421}} \\
    \textcolor{black}{$\spadesuit$} SFDA-FSM (2022, MIA)~\cite{yang2022source}     & \cellcolor{best}0.9319{\scriptsize\textcolor{customPurple}{$\pm$0.0483}} & 9.87{\scriptsize\textcolor{customPurple}{$\pm$26.50}} & \cellcolor{best}0.8761{\scriptsize\textcolor{customPurple}{$\pm$0.0797}} & \cellcolor{second}0.9652{\scriptsize\textcolor{customPurple}{$\pm$0.0364}} & \cellcolor{best}0.9051{\scriptsize\textcolor{customPurple}{$\pm$0.0797}} \\
    \textcolor{black}{$\spadesuit$} STDR (2024, TMI)~\cite{wang2024dual}         & \cellcolor{second}0.9171{\scriptsize\textcolor{customPurple}{$\pm$0.0862}} & 7.85{\scriptsize\textcolor{customPurple}{$\pm$14.68}} & \cellcolor{second}0.8589{\scriptsize\textcolor{customPurple}{$\pm$0.1312}} & 0.9010{\scriptsize\textcolor{customPurple}{$\pm$0.0495}} & 0.8617{\scriptsize\textcolor{customPurple}{$\pm$0.1441}} \\
    \textcolor{black}{$\spadesuit$} AIF-SFDA (2025, AAAI)~\cite{li2025aif}     & 0.8978{\scriptsize\textcolor{customPurple}{$\pm$0.1059}} & 8.12{\scriptsize\textcolor{customPurple}{$\pm$24.09}} & 0.8245{\scriptsize\textcolor{customPurple}{$\pm$0.1466}} & 0.9605{\scriptsize\textcolor{customPurple}{$\pm$0.0774}} & 0.8569{\scriptsize\textcolor{customPurple}{$\pm$0.1495}} \\
    \arrayrulecolor{black}\hline
    \textcolor{green}{$\clubsuit$} DANN (2016, JMLR)~\cite{ganin2016domain}         & 0.9015{\scriptsize\textcolor{customPurple}{$\pm$0.0915}} & 9.30{\scriptsize\textcolor{customPurple}{$\pm$22.15}} & 0.8315{\scriptsize\textcolor{customPurple}{$\pm$0.1313}} & 0.9550{\scriptsize\textcolor{customPurple}{$\pm$0.0608}} & 0.8664{\scriptsize\textcolor{customPurple}{$\pm$0.1373}} \\
    \textcolor{green}{$\clubsuit$} DeTTA (2024, WACV)~\cite{wen2024denoising}        & 0.9069{\scriptsize\textcolor{customPurple}{$\pm$0.0855}} & 9.52{\scriptsize\textcolor{customPurple}{$\pm$18.35}} & 0.8392{\scriptsize\textcolor{customPurple}{$\pm$0.1233}} & 0.9467{\scriptsize\textcolor{customPurple}{$\pm$0.0676}} & 0.8836{\scriptsize\textcolor{customPurple}{$\pm$0.1305}} \\
    \rowcolor[HTML]{F0F0F0}
    \textcolor{green}{$\clubsuit$} TestFit (2024, MIA)~\cite{zhang2024testfit}      & 0.8930{\scriptsize\textcolor{customPurple}{$\pm$0.0951}} & 13.83{\scriptsize\textcolor{customPurple}{$\pm$28.88}} & 0.8186{\scriptsize\textcolor{customPurple}{$\pm$0.1391}} & 0.9601{\scriptsize\textcolor{customPurple}{$\pm$0.0854}} & 0.8505{\scriptsize\textcolor{customPurple}{$\pm$0.1516}} \\
    \rowcolor[HTML]{F0F0F0}
    \textcolor{green}{$\clubsuit$} UDA-MIMA (2024, CMPB)~\cite{hu2024unsupervised}     & 0.8911{\scriptsize\textcolor{customPurple}{$\pm$0.1074}} & 11.99{\scriptsize\textcolor{customPurple}{$\pm$25.17}} & 0.8170{\scriptsize\textcolor{customPurple}{$\pm$0.1414}} & 0.9498{\scriptsize\textcolor{customPurple}{$\pm$0.0734}} & 0.8521{\scriptsize\textcolor{customPurple}{$\pm$0.1502}} \\
    \textcolor{green}{$\clubsuit$} GraTa (2025, AAAI)~\cite{chen2025gradient}        & 0.9073{\scriptsize\textcolor{customPurple}{$\pm$0.0965}} & \cellcolor{second}6.75{\scriptsize\textcolor{customPurple}{$\pm$18.78}} & 0.8410{\scriptsize\textcolor{customPurple}{$\pm$0.1232}} & 0.9278{\scriptsize\textcolor{customPurple}{$\pm$0.0822}} & \cellcolor{second}0.8981{\scriptsize\textcolor{customPurple}{$\pm$0.1307}} \\
    \arrayrulecolor{black}\hline
    \textcolor{red}{$\varheartsuit$} TENT (2021, ICLR)~\cite{wang2021tent}         & 0.9022{\scriptsize\textcolor{customPurple}{$\pm$0.0921}} & \cellcolor{third}7.22{\scriptsize\textcolor{customPurple}{$\pm$19.12}} & 0.8330{\scriptsize\textcolor{customPurple}{$\pm$0.1325}} & 0.9554{\scriptsize\textcolor{customPurple}{$\pm$0.0598}} & 0.8682{\scriptsize\textcolor{customPurple}{$\pm$0.1392}} \\
    \textcolor{red}{$\varheartsuit$} UPL-SFDA (2023, TMI)~\cite{wu2023upl}     & \cellcolor{third}0.9099{\scriptsize\textcolor{customPurple}{$\pm$0.0849}} & \cellcolor{best}6.40{\scriptsize\textcolor{customPurple}{$\pm$18.33}} & \cellcolor{third}0.8441{\scriptsize\textcolor{customPurple}{$\pm$0.1226}} & 0.9435{\scriptsize\textcolor{customPurple}{$\pm$0.0690}} & 0.8911{\scriptsize\textcolor{customPurple}{$\pm$0.1285}} \\
    \textcolor{red}{$\varheartsuit$} DG-TTA (2023, arXiv)~\cite{weihsbach2023dg}        & 0.9053{\scriptsize\textcolor{customPurple}{$\pm$0.0748}} & 9.96{\scriptsize\textcolor{customPurple}{$\pm$23.44}} & 0.8347{\scriptsize\textcolor{customPurple}{$\pm$0.1140}} & 0.9295{\scriptsize\textcolor{customPurple}{$\pm$0.0665}} & \cellcolor{third}0.8967{\scriptsize\textcolor{customPurple}{$\pm$0.1285}} \\
    \textcolor{red}{$\varheartsuit$} SaTTCA (2023, MICCAI)~\cite{li2023scale}        & 0.9030{\scriptsize\textcolor{customPurple}{$\pm$0.0823}} & 8.52{\scriptsize\textcolor{customPurple}{$\pm$8.03}} & 0.8330{\scriptsize\textcolor{customPurple}{$\pm$0.0389}} & 0.9452{\scriptsize\textcolor{customPurple}{$\pm$0.0364}} & 0.8782{\scriptsize\textcolor{customPurple}{$\pm$0.0541}} \\
    \textcolor{red}{$\varheartsuit$} SmaRT (2025, arXiv)~\cite{Wang2025SmaRTSR}        & 0.8993{\scriptsize\textcolor{customPurple}{$\pm$0.0924}} & 7.49{\scriptsize\textcolor{customPurple}{$\pm$16.54}} & 0.8280{\scriptsize\textcolor{customPurple}{$\pm$0.1327}} & \cellcolor{best}0.9738{\scriptsize\textcolor{customPurple}{$\pm$0.0348}} & 0.8485{\scriptsize\textcolor{customPurple}{$\pm$0.1401}} \\
    \arrayrulecolor{black}\hline
    \textcolor{blue}{$\vardiamondsuit$} AdaMI (2022, MICCAI)~\cite{bateson2022test}        & 0.9013{\scriptsize\textcolor{customPurple}{$\pm$0.0943}} & 10.25{\scriptsize\textcolor{customPurple}{$\pm$22.16}} & 0.8285{\scriptsize\textcolor{customPurple}{$\pm$0.1337}} & 0.9631{\scriptsize\textcolor{customPurple}{$\pm$0.0425}} & 0.8624{\scriptsize\textcolor{customPurple}{$\pm$0.1461}} \\
    \textcolor{blue}{$\vardiamondsuit$} PASS (2024, TMI)~\cite{zhang2024pass}         & 0.9009{\scriptsize\textcolor{customPurple}{$\pm$0.0920}} & 9.33{\scriptsize\textcolor{customPurple}{$\pm$22.15}} & 0.8307{\scriptsize\textcolor{customPurple}{$\pm$0.1320}} & 0.9552{\scriptsize\textcolor{customPurple}{$\pm$0.0609}} & 0.8645{\scriptsize\textcolor{customPurple}{$\pm$0.1380}} \\
    \textcolor{blue}{$\vardiamondsuit$} VPTTA (2024, CVPR)~\cite{chen2024each}        & 0.8975{\scriptsize\textcolor{customPurple}{$\pm$0.0993}} & 13.83{\scriptsize\textcolor{customPurple}{$\pm$28.89}} & 0.8229{\scriptsize\textcolor{customPurple}{$\pm$0.1432}} & \cellcolor{third}0.9643{\scriptsize\textcolor{customPurple}{$\pm$0.0426}} & 0.8548{\scriptsize\textcolor{customPurple}{$\pm$0.1554}} \\
    \rowcolor[HTML]{F0F0F0}
    \textcolor{blue}{$\vardiamondsuit$} ExploringTTA (2025, ISBI)~\cite{omolegan2025exploring} & 0.8934{\scriptsize\textcolor{customPurple}{$\pm$0.1017}} & 11.11{\scriptsize\textcolor{customPurple}{$\pm$24.09}} & 0.8203{\scriptsize\textcolor{customPurple}{$\pm$0.1427}} & 0.9567{\scriptsize\textcolor{customPurple}{$\pm$0.0737}} & 0.8527{\scriptsize\textcolor{customPurple}{$\pm$0.1458}} \\
    \textcolor{blue}{$\vardiamondsuit$} ProSFDA (2026, PR)~\cite{hu2025source}      & 0.8973{\scriptsize\textcolor{customPurple}{$\pm$0.0943}} & 11.33{\scriptsize\textcolor{customPurple}{$\pm$24.18}} & 0.8254{\scriptsize\textcolor{customPurple}{$\pm$0.1360}} & 0.9616{\scriptsize\textcolor{customPurple}{$\pm$0.0458}} & 0.8557{\scriptsize\textcolor{customPurple}{$\pm$0.1460}} \\
    \arrayrulecolor{black}\hline
  \end{tabular}%
  }
\end{table}

\subsubsection{Histopathology Modality}

\begin{figure}[tbp]
  \centering
  \includegraphics[width=\linewidth]{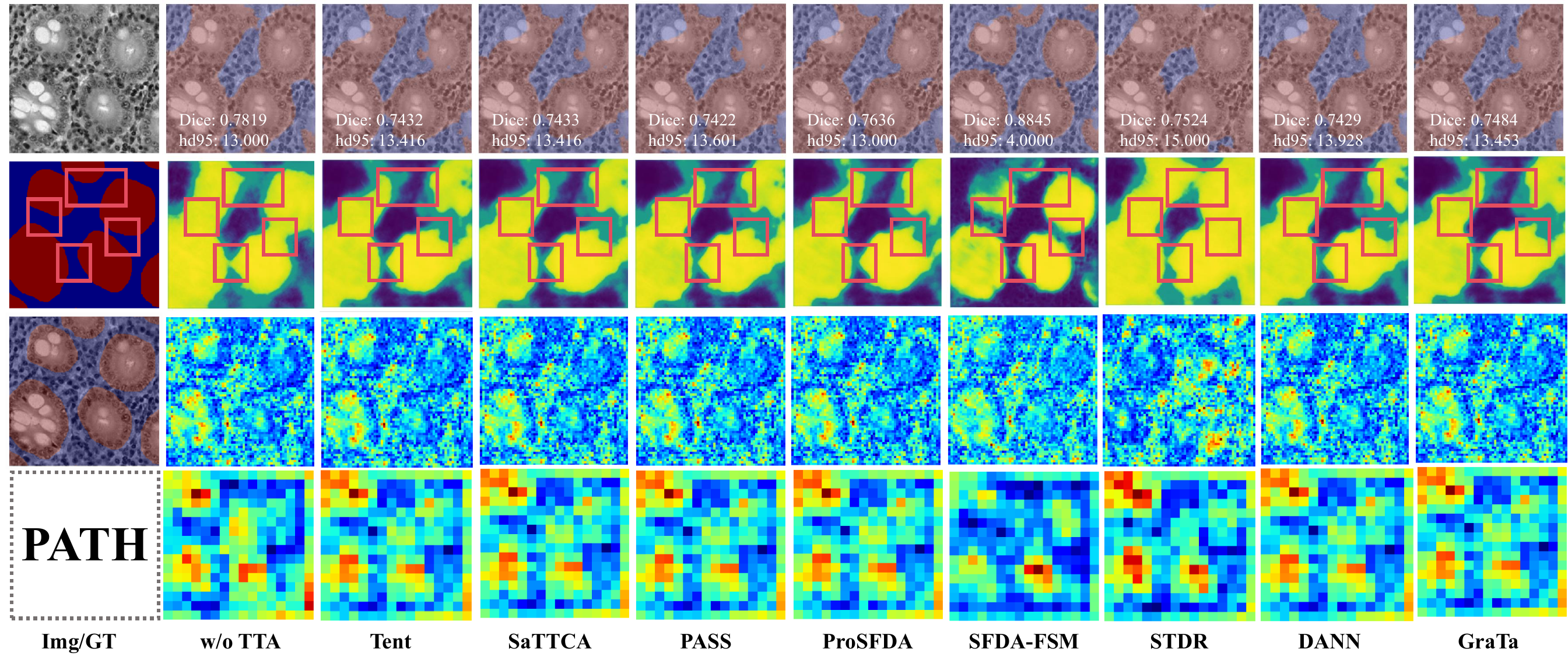}
  \caption{\textbf{Visual comparison of TTA efficacy across different methods on DER datasets.} Rows 1 and 2 illustrate the segmentation outcomes. Rows 3 and 4 depict changes in features extracted by the Encoder after TTA. GT denotes the ground-truth.}
  \label{fig: PATH_seg}
\end{figure}

As reported in the Table~\ref{tab:PATH}, results on the Histopathology task showed greater divergence: only 6 of 20 methods improved Dice scores, while 13 methods experienced degradation. SFDA-FSM again achieved the best Dice score of 0.8713, representing an 8.6\% increase, whereas UPL-SFDA yielded the most remarkable HD95 improvement, a 47.1\% reduction. However, most Prior Estimation methods, such as PASS, AdaMI, and VPTTA, performed poorly in this modality, highlighting the substantial impact of staining variations and scanning heterogeneity on prior-dependent strategies. Notably, our benchmark employs a patch-level evaluation protocol, which effectively circumvents the limitations and instability associated with WSI-level adaptation and GPU memory constraints. This also underscores that WSI-level TTA remains a critical open challenge. Fig.~\ref{fig: PATH_seg} presents a visualization of the effects of the TTA method on the Histopathology task, illustrating the changes in the feature extraction capabilities of the encoder. The segmentation results indicate a general refinement of the boundaries.
\begin{table}[h!]
  \centering
  \caption{\textbf{Comprehensive performance comparison of TTA methods on Colorectal Gland Segmentation in Histopathology Images.}
 Results colored in \colorbox{red!30}{\makebox[1.3em][c]{\raisebox{0pt}[0.5em][0em]{red}}}, \colorbox{orange!30}{\makebox[2.8em][c]{\raisebox{0pt}[0.5em][0em]{orange}}}, and \colorbox{yellow!30}{\makebox[2.5em][c]{\raisebox{0pt}[0.5em][0em]{yellow}}} denote the best, second-best, and third-best performances, respectively. Rows with a \colorbox{gray!30}{\makebox[1.6em][c]{\raisebox{0pt}[0.5em][0em]{gray}}} background indicate Dice values lower than that of the Target-Domain(w/o TTA). Symbols denote paradigms: \textcolor{black}{\(\spadesuit\)} for Input-level Translation, \textcolor{green}{\(\clubsuit\)} for Feature-level Alignment, \textcolor{red}{\(\varheartsuit\)} for Output-level Regularization, \textcolor{blue}{\(\vardiamondsuit\)} for Prior Estimation. All improvements are statistically significant at \( p < 0.05 \).}
  \vspace{0.5em}
  \label{tab:PATH}
  \resizebox{\textwidth}{!}{%
  \begin{tabular}{lccccc}
    \arrayrulecolor{black}\hline
    \rowcolor[HTML]{D8D6C0}
    \textbf{Method} & \textbf{Dice $\uparrow$} & \textbf{HD95 $\downarrow$} & \textbf{JI $\uparrow$} & \textbf{Sen $\uparrow$} & \textbf{PPV $\uparrow$} \arraybackslash \\
    \arrayrulecolor{black}\hline
    \textbf{Intra-Domain} & \textbf{0.8360{\scriptsize\textcolor{customPurple}{$\pm$0.1966}}} & \textbf{32.66{\scriptsize\textcolor{customPurple}{$\pm$56.88}}} & \textbf{0.7562{\scriptsize\textcolor{customPurple}{$\pm$0.2288}}} & \textbf{0.9583{\scriptsize\textcolor{customPurple}{$\pm$0.0826}}} & \textbf{0.7834{\scriptsize\textcolor{customPurple}{$\pm$0.2342}}} \\
    \arrayrulecolor{black}\hline
    \textbf{Target-Domain (w/o TTA)} & \textbf{0.8026{\scriptsize\textcolor{customPurple}{$\pm$0.1782}}} & \textbf{33.66{\scriptsize\textcolor{customPurple}{$\pm$49.01}}} & \textbf{0.7001{\scriptsize\textcolor{customPurple}{$\pm$0.2034}}} & \textbf{0.9422{\scriptsize\textcolor{customPurple}{$\pm$0.1119}}} & \textbf{0.7307{\scriptsize\textcolor{customPurple}{$\pm$0.2063}}} \\
    \arrayrulecolor{black}\hline
    \rowcolor[HTML]{F0F0F0}
    \textcolor{black}{$\spadesuit$} RSA (2021, MICCAI)~\cite{zeng2024reliable}          & 0.7810{\scriptsize\textcolor{customPurple}{$\pm$0.1745}} & 35.72{\scriptsize\textcolor{customPurple}{$\pm$46.39}} & 0.6691{\scriptsize\textcolor{customPurple}{$\pm$0.2016}} & 0.9314{\scriptsize\textcolor{customPurple}{$\pm$0.1196}} & 0.7092{\scriptsize\textcolor{customPurple}{$\pm$0.2060}} \\
    \rowcolor[HTML]{F0F0F0}
    \textcolor{black}{$\spadesuit$} DL-TTA (2022, TMI)~\cite{yang2022dltta}       & 0.7935{\scriptsize\textcolor{customPurple}{$\pm$0.1748}} & 34.62{\scriptsize\textcolor{customPurple}{$\pm$47.45}} & 0.6867{\scriptsize\textcolor{customPurple}{$\pm$0.2035}} & 0.9336{\scriptsize\textcolor{customPurple}{$\pm$0.1095}} & 0.7262{\scriptsize\textcolor{customPurple}{$\pm$0.2087}} \\
        \textcolor{black}{$\spadesuit$} SFDA-FSM (2022, MIA)~\cite{yang2022source}     & \cellcolor{best}0.8713{\scriptsize\textcolor{customPurple}{$\pm$0.1460}} & \cellcolor{second}19.56{\scriptsize\textcolor{customPurple}{$\pm$41.32}} & \cellcolor{best}0.7944{\scriptsize\textcolor{customPurple}{$\pm$0.1763}} & 0.9142{\scriptsize\textcolor{customPurple}{$\pm$0.1173}} & \cellcolor{best}0.8526{\scriptsize\textcolor{customPurple}{$\pm$0.1709}} \\
    \rowcolor[HTML]{F0F0F0}
    \textcolor{black}{$\spadesuit$} STDR (2024, TMI)~\cite{wang2024dual}         & 0.7993{\scriptsize\textcolor{customPurple}{$\pm$0.1795}} & 34.81{\scriptsize\textcolor{customPurple}{$\pm$46.23}} & 0.6952{\scriptsize\textcolor{customPurple}{$\pm$0.2002}} & \cellcolor{best}0.9579{\scriptsize\textcolor{customPurple}{$\pm$0.1152}} & 0.7118{\scriptsize\textcolor{customPurple}{$\pm$0.2247}} \\

    \textcolor{black}{$\spadesuit$} AIF-SFDA (2025, AAAI)~\cite{li2025aif}     & 0.8082{\scriptsize\textcolor{customPurple}{$\pm$0.1801}} & 33.46{\scriptsize\textcolor{customPurple}{$\pm$48.66}} & 0.7053{\scriptsize\textcolor{customPurple}{$\pm$0.2057}} & 0.9397{\scriptsize\textcolor{customPurple}{$\pm$0.1098}} & 0.7342{\scriptsize\textcolor{customPurple}{$\pm$0.2086}} \\
    \arrayrulecolor{black}\hline
    \rowcolor[HTML]{F0F0F0}
    \textcolor{green}{$\clubsuit$} DANN (2016, JMLR)~\cite{ganin2016domain}         & 0.7794{\scriptsize\textcolor{customPurple}{$\pm$0.1900}} & 38.37{\scriptsize\textcolor{customPurple}{$\pm$51.65}} & 0.6758{\scriptsize\textcolor{customPurple}{$\pm$0.2132}} & 0.9079{\scriptsize\textcolor{customPurple}{$\pm$0.1262}} & 0.7301{\scriptsize\textcolor{customPurple}{$\pm$0.2302}} \\
        \textcolor{green}{$\clubsuit$} DeTTA (2024, WACV)~\cite{wen2024denoising}        & 0.8035{\scriptsize\textcolor{customPurple}{$\pm$0.1617}} & 35.49{\scriptsize\textcolor{customPurple}{$\pm$49.71}} & 0.7005{\scriptsize\textcolor{customPurple}{$\pm$0.1842}} & 0.9238{\scriptsize\textcolor{customPurple}{$\pm$0.1650}} & 0.7342{\scriptsize\textcolor{customPurple}{$\pm$0.2209}} \\
            \textcolor{green}{$\clubsuit$} TestFit (2024, MIA)~\cite{zhang2024testfit}      & 0.8037{\scriptsize\textcolor{customPurple}{$\pm$0.1758}} & 33.45{\scriptsize\textcolor{customPurple}{$\pm$48.65}} & 0.7010{\scriptsize\textcolor{customPurple}{$\pm$0.2015}} & \cellcolor{third}0.9454{\scriptsize\textcolor{customPurple}{$\pm$0.1056}} & 0.7300{\scriptsize\textcolor{customPurple}{$\pm$0.2045}} \\
    \rowcolor[HTML]{F0F0F0}
    \textcolor{green}{$\clubsuit$} UDA-MIMA (2024, CMPB)~\cite{hu2024unsupervised}     & 0.7890{\scriptsize\textcolor{customPurple}{$\pm$0.1732}} & 33.16{\scriptsize\textcolor{customPurple}{$\pm$46.01}} & 0.6795{\scriptsize\textcolor{customPurple}{$\pm$0.1984}} & 0.8462{\scriptsize\textcolor{customPurple}{$\pm$0.1685}} & 0.7920{\scriptsize\textcolor{customPurple}{$\pm$0.2055}} \\
    \rowcolor[HTML]{F0F0F0}
    \textcolor{green}{$\clubsuit$} GraTa (2025, AAAI)~\cite{chen2025gradient}        & 0.7892{\scriptsize\textcolor{customPurple}{$\pm$0.1832}} & 36.49{\scriptsize\textcolor{customPurple}{$\pm$49.71}} & 0.6832{\scriptsize\textcolor{customPurple}{$\pm$0.2107}} & 0.9139{\scriptsize\textcolor{customPurple}{$\pm$0.1232}} & 0.7372{\scriptsize\textcolor{customPurple}{$\pm$0.2193}} \\

    \arrayrulecolor{black}\hline
    \rowcolor[HTML]{F0F0F0}
    \textcolor{red}{$\varheartsuit$} TENT (2021, ICLR)~\cite{wang2021tent}         & 0.7890{\scriptsize\textcolor{customPurple}{$\pm$0.1833}} & 36.50{\scriptsize\textcolor{customPurple}{$\pm$49.68}} & 0.6830{\scriptsize\textcolor{customPurple}{$\pm$0.2108}} & 0.9136{\scriptsize\textcolor{customPurple}{$\pm$0.1236}} & 0.7371{\scriptsize\textcolor{customPurple}{$\pm$0.2193}} \\
        \textcolor{red}{$\varheartsuit$} UPL-SFDA (2023, TMI)~\cite{wu2023upl}     & \cellcolor{second}0.8438{\scriptsize\textcolor{customPurple}{$\pm$0.1820}} & \cellcolor{best}17.81{\scriptsize\textcolor{customPurple}{$\pm$43.01}} & \cellcolor{second}0.7629{\scriptsize\textcolor{customPurple}{$\pm$0.2132}} & 0.9435{\scriptsize\textcolor{customPurple}{$\pm$0.1120}} & \cellcolor{second}0.8038{\scriptsize\textcolor{customPurple}{$\pm$0.2331}} \\
        
    \rowcolor[HTML]{F0F0F0}
    \textcolor{red}{$\varheartsuit$} DG-TTA (2023, arXiv)~\cite{weihsbach2023dg}        & 0.8018{\scriptsize\textcolor{customPurple}{$\pm$0.1772}} & 33.76{\scriptsize\textcolor{customPurple}{$\pm$48.88}} & 0.6987{\scriptsize\textcolor{customPurple}{$\pm$0.2026}} & \cellcolor{second}0.9455{\scriptsize\textcolor{customPurple}{$\pm$0.1074}} & 0.7276{\scriptsize\textcolor{customPurple}{$\pm$0.2054}} \\
    \rowcolor[HTML]{F0F0F0}
    \textcolor{red}{$\varheartsuit$} SaTTCA (2023, MICCAI)~\cite{li2023scale}        & 0.7686{\scriptsize\textcolor{customPurple}{$\pm$0.1186}} & 40.49{\scriptsize\textcolor{customPurple}{$\pm$32.23}} & 0.6566{\scriptsize\textcolor{customPurple}{$\pm$0.1358}} & 0.8958{\scriptsize\textcolor{customPurple}{$\pm$0.0770}} & 0.7207{\scriptsize\textcolor{customPurple}{$\pm$0.1568}} \\
    \rowcolor[HTML]{F0F0F0}
    \textcolor{red}{$\varheartsuit$} SmaRT (2025, arXiv)~\cite{Wang2025SmaRTSR}        & 0.7385{\scriptsize\textcolor{customPurple}{$\pm$0.1978}} & 43.84{\scriptsize\textcolor{customPurple}{$\pm$52.81}} & 0.6183{\scriptsize\textcolor{customPurple}{$\pm$0.2129}} & 0.9160{\scriptsize\textcolor{customPurple}{$\pm$0.1082}} & 0.6613{\scriptsize\textcolor{customPurple}{$\pm$0.2298}} \\

    \arrayrulecolor{black}\hline
    \rowcolor[HTML]{F0F0F0}
    \textcolor{blue}{$\vardiamondsuit$} AdaMI (2022, MICCAI)~\cite{bateson2022test}        & 0.7930{\scriptsize\textcolor{customPurple}{$\pm$0.1743}} & 34.21{\scriptsize\textcolor{customPurple}{$\pm$48.18}} & 0.6854{\scriptsize\textcolor{customPurple}{$\pm$0.2001}} & 0.8708{\scriptsize\textcolor{customPurple}{$\pm$0.1518}} & 0.7773{\scriptsize\textcolor{customPurple}{$\pm$0.2109}} \\
        \rowcolor[HTML]{F0F0F0}
    \textcolor{blue}{$\vardiamondsuit$} PASS (2024, TMI)~\cite{zhang2024pass}         & 0.7938{\scriptsize\textcolor{customPurple}{$\pm$0.1743}} & 34.12{\scriptsize\textcolor{customPurple}{$\pm$48.34}} & 0.6866{\scriptsize\textcolor{customPurple}{$\pm$0.2002}} & 0.8712{\scriptsize\textcolor{customPurple}{$\pm$0.1511}} & 0.7781{\scriptsize\textcolor{customPurple}{$\pm$0.2108}} \\
    \rowcolor[HTML]{F0F0F0}
    \textcolor{blue}{$\vardiamondsuit$} VPTTA (2024, CVPR)~\cite{chen2024each}        & 0.7938{\scriptsize\textcolor{customPurple}{$\pm$0.1875}} & 38.45{\scriptsize\textcolor{customPurple}{$\pm$42.91}} & 0.6876{\scriptsize\textcolor{customPurple}{$\pm$0.2149}} & 0.9182{\scriptsize\textcolor{customPurple}{$\pm$0.1274}} & 0.7415{\scriptsize\textcolor{customPurple}{$\pm$0.2234}} \\

    \rowcolor[HTML]{F0F0F0}
    \textcolor{blue}{$\vardiamondsuit$} ExploringTTA (2025, ISBI)~\cite{omolegan2025exploring} & 0.7835{\scriptsize\textcolor{customPurple}{$\pm$0.1818}} & 36.16{\scriptsize\textcolor{customPurple}{$\pm$48.91}} & 0.6745{\scriptsize\textcolor{customPurple}{$\pm$0.2069}} & 0.8743{\scriptsize\textcolor{customPurple}{$\pm$0.1387}} & 0.7573{\scriptsize\textcolor{customPurple}{$\pm$0.2231}} \\
    \textcolor{blue}{$\vardiamondsuit$} ProSFDA (2026, PR)~\cite{hu2025source}      & \cellcolor{third}0.8263{\scriptsize\textcolor{customPurple}{$\pm$0.1647}} & \cellcolor{third}26.95{\scriptsize\textcolor{customPurple}{$\pm$43.57}} & \cellcolor{third}0.7308{\scriptsize\textcolor{customPurple}{$\pm$0.1944}} & 0.8772{\scriptsize\textcolor{customPurple}{$\pm$0.1457}} & \cellcolor{third}0.8173{\scriptsize\textcolor{customPurple}{$\pm$0.1893}} \\
    \arrayrulecolor{black}\hline
  \end{tabular}%
  }
\end{table}

\subsubsection{Dermoscopy Modality}

\begin{figure}[h!]
  \centering
  \includegraphics[width=\linewidth]{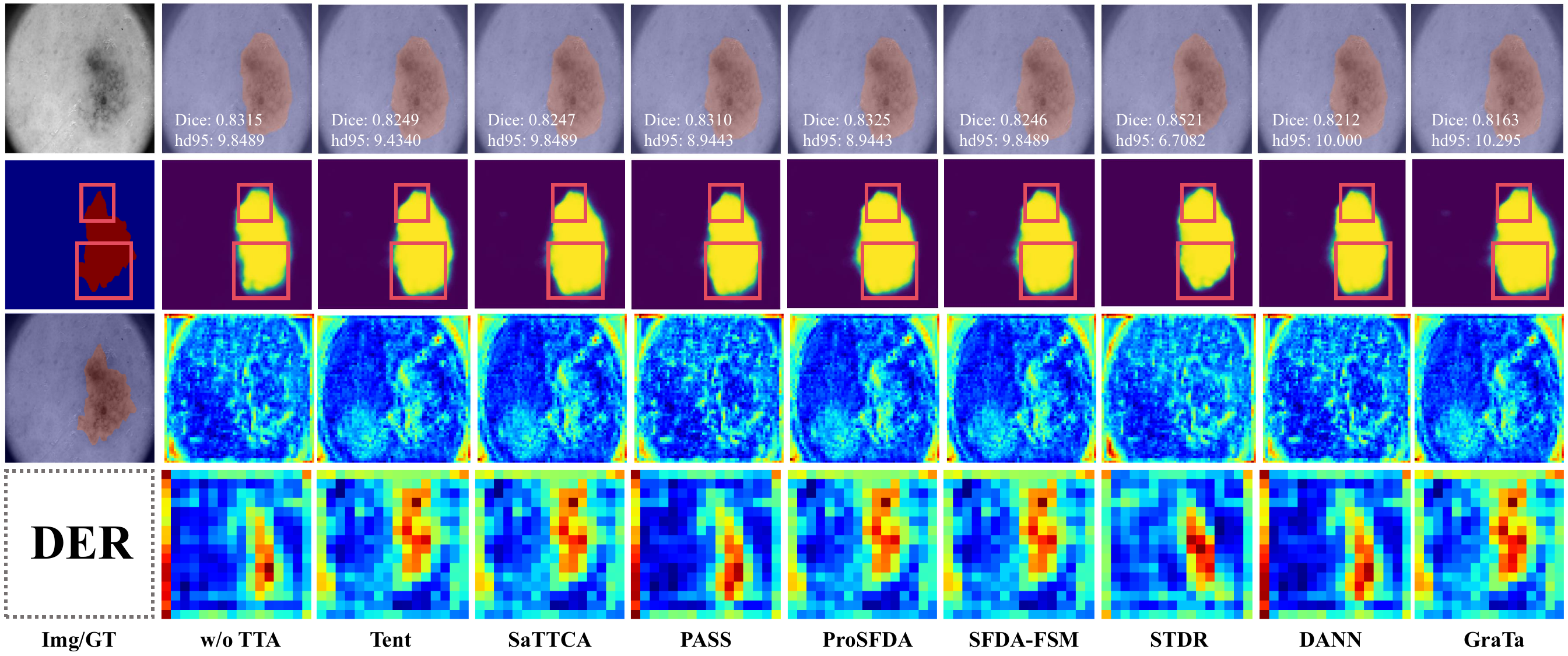}
  \caption{\textbf{Visual comparison of TTA efficacy across different methods on DER datasets.} Rows 1 and 2 illustrate the segmentation outcomes. Rows 3 and 4 depict changes in features extracted by the Encoder after TTA. GT denotes the ground-truth.}
  \label{fig: DER_seg}
\end{figure}

In DER, as summarized in the Table~\ref{tab:der}, SFDA-FSM continued to excel with a Dice score of 0.9258, an improvement of 8.5 percent, and HD95 of 3.58, a reduction of 74 percent, followed closely by SmaRT. In contrast, UDA-MIMA exhibited substantial degradation, with the Dice score decreasing by 7.3\% and the HD95 score increasing by 59\%. Overall, Input-level Transformation and Output-level Regularization showed clear advantages in both Dice and HD95, while feature alignment and Prior Estimation tended to fail. This stems from the domain-specific characteristics of dermoscopic imaging, including variations in color calibration and illumination, device-dependent imaging artifacts, and confounders such as hair and glare, which intensify boundary ambiguity. Additionally, the high proportion of small, low-contrast lesions amplifies pseudo-label noise, causing representational drift and undermining fine-grained alignment strategies. Fig.~\ref{fig: DER_seg} demonstrates the effects of various TTA methods on the DER task, as well as their influence on the encoder. The performance of the different methods varies, exhibiting both positive and negative outcomes.
\begin{table}[h!]
  \centering
  \caption{\textbf{Comprehensive performance comparison of TTA methods on Melanoma Segmentation in DER Images.}
 Results colored in \colorbox{red!30}{\makebox[1.3em][c]{\raisebox{0pt}[0.5em][0em]{red}}}, \colorbox{orange!30}{\makebox[2.8em][c]{\raisebox{0pt}[0.5em][0em]{orange}}}, and \colorbox{yellow!30}{\makebox[2.5em][c]{\raisebox{0pt}[0.5em][0em]{yellow}}} denote the best, second-best, and third-best performances, respectively. Rows with a \colorbox{gray!30}{\makebox[1.6em][c]{\raisebox{0pt}[0.5em][0em]{gray}}} background indicate Dice values lower than that of the Target-Domain(w/o TTA). Symbols denote paradigms: \textcolor{black}{\(\spadesuit\)} for Input-level Translation, \textcolor{green}{\(\clubsuit\)} for Feature-level Alignment, \textcolor{red}{\(\varheartsuit\)} for Output-level Regularization, \textcolor{blue}{\(\vardiamondsuit\)} for Prior Estimation. All improvements are statistically significant at \( p < 0.05 \).}
  \vspace{0.5em}
  \label{tab:der}
  \resizebox{\textwidth}{!}{%
  \begin{tabular}{lccccc}
    \arrayrulecolor{black}\hline
    \rowcolor[HTML]{D8D6C0}
    \textbf{Method} & \textbf{Dice $\uparrow$} & \textbf{HD95 $\downarrow$} & \textbf{JI $\uparrow$} & \textbf{Sen $\uparrow$} & \textbf{PPV $\uparrow$} \\
    \arrayrulecolor{black}\hline
    \textbf{Intra-Domain} &\textbf{ 0.9280{\scriptsize\textcolor{customPurple}{$\pm$0.0767}}} & \textbf{2.33{\scriptsize\textcolor{customPurple}{$\pm$3.47}}} & \textbf{0.8723{\scriptsize\textcolor{customPurple}{$\pm$0.0979}}} & \textbf{0.9397{\scriptsize\textcolor{customPurple}{$\pm$0.1006}}} & \textbf{0.9270{\scriptsize\textcolor{customPurple}{$\pm$0.0797}}} \\
  \hline
    \textbf{Target Domain (w/o TTA)} & \textbf{0.8532{\scriptsize\textcolor{customPurple}{$\pm$0.0864}}} & \textbf{13.78{\scriptsize\textcolor{customPurple}{$\pm$13.69}}} & \textbf{0.7716{\scriptsize\textcolor{customPurple}{$\pm$0.1052}}} & \textbf{0.8527{\scriptsize\textcolor{customPurple}{$\pm$0.1323}}} & \textbf{0.9120{\scriptsize\textcolor{customPurple}{$\pm$0.0556}}} \\
   \hline
    \rowcolor[HTML]{F0F0F0}
    \textcolor{black}{$\spadesuit$} RSA (2021, MICCAI)~\cite{zeng2024reliable} & 0.8517{\scriptsize\textcolor{customPurple}{$\pm$0.1747}} & 16.85{\scriptsize\textcolor{customPurple}{$\pm$42.47}} & 0.7724{\scriptsize\textcolor{customPurple}{$\pm$0.2053}} & 0.8179{\scriptsize\textcolor{customPurple}{$\pm$0.2256}} & \cellcolor{second}0.9387{\scriptsize\textcolor{customPurple}{$\pm$0.1134}} \\
    \rowcolor[HTML]{F0F0F0}
    \textcolor{black}{$\spadesuit$} DL-TTA (2022, TMI)~\cite{yang2022dltta} & 0.8478{\scriptsize\textcolor{customPurple}{$\pm$0.1648}} & 13.64{\scriptsize\textcolor{customPurple}{$\pm$25.37}} & 0.7672{\scriptsize\textcolor{customPurple}{$\pm$0.1840}} & 0.8576{\scriptsize\textcolor{customPurple}{$\pm$0.2318}} & \cellcolor{third}0.9341{\scriptsize\textcolor{customPurple}{$\pm$0.1326}} \\
    \textcolor{black}{$\spadesuit$} SFDA-FSM (2022, MIA)~\cite{yang2022source} & \cellcolor{best}0.9258{\scriptsize\textcolor{customPurple}{$\pm$0.0604}} & \cellcolor{best}3.58{\scriptsize\textcolor{customPurple}{$\pm$9.86}} & \cellcolor{best}0.8670{\scriptsize\textcolor{customPurple}{$\pm$0.0904}} & \cellcolor{second}0.8979{\scriptsize\textcolor{customPurple}{$\pm$0.0937}} & \cellcolor{best}0.9667{\scriptsize\textcolor{customPurple}{$\pm$0.0619}} \\
    \rowcolor[HTML]{F0F0F0}
    \textcolor{black}{$\spadesuit$} STDR (2024, TMI)~\cite{wang2024dual} & 0.8526{\scriptsize\textcolor{customPurple}{$\pm$0.1682}} & 15.65{\scriptsize\textcolor{customPurple}{$\pm$36.71}} & 0.7692{\scriptsize\textcolor{customPurple}{$\pm$0.1883}} & 0.8540{\scriptsize\textcolor{customPurple}{$\pm$0.2106}} & 0.8623{\scriptsize\textcolor{customPurple}{$\pm$0.1139}} \\
    \textcolor{black}{$\spadesuit$} AIF-SFDA (2025, AAAI)~\cite{li2025aif} & 0.8538{\scriptsize\textcolor{customPurple}{$\pm$0.1735}} & 13.66{\scriptsize\textcolor{customPurple}{$\pm$36.71}} & 0.7749{\scriptsize\textcolor{customPurple}{$\pm$0.1936}} & 0.8597{\scriptsize\textcolor{customPurple}{$\pm$0.2159}} & 0.8698{\scriptsize\textcolor{customPurple}{$\pm$0.1192}} \\
    \midrule
    \rowcolor[HTML]{F0F0F0}
    \textcolor{green}{$\clubsuit$} DANN (2016, JMLR)~\cite{ganin2016domain} & 0.8425{\scriptsize\textcolor{customPurple}{$\pm$0.1715}} & 14.71{\scriptsize\textcolor{customPurple}{$\pm$27.68}} & 0.7568{\scriptsize\textcolor{customPurple}{$\pm$0.1994}} & 0.8489{\scriptsize\textcolor{customPurple}{$\pm$0.2276}} & 0.9098{\scriptsize\textcolor{customPurple}{$\pm$0.1042}} \\
    \textcolor{green}{$\clubsuit$} TestFit (2024, MIA)~\cite{zhang2024testfit} & 0.8686{\scriptsize\textcolor{customPurple}{$\pm$0.1428}} & \cellcolor{third}11.52{\scriptsize\textcolor{customPurple}{$\pm$23.84}} & 0.7856{\scriptsize\textcolor{customPurple}{$\pm$0.1686}} & \cellcolor{third}0.8925{\scriptsize\textcolor{customPurple}{$\pm$0.1916}} & 0.8890{\scriptsize\textcolor{customPurple}{$\pm$0.0943}} \\
    \rowcolor[HTML]{F0F0F0}
    \textcolor{green}{$\clubsuit$} DeTTA (2024, WACV)~\cite{wen2024denoising} & 0.8479{\scriptsize\textcolor{customPurple}{$\pm$0.1785}} & 15.84{\scriptsize\textcolor{customPurple}{$\pm$42.27}} & 0.7645{\scriptsize\textcolor{customPurple}{$\pm$0.2182}} & 0.8489{\scriptsize\textcolor{customPurple}{$\pm$0.2347}} & 0.9018{\scriptsize\textcolor{customPurple}{$\pm$0.1385}} \\
    \rowcolor[HTML]{F0F0F0}
    \textcolor{green}{$\clubsuit$} UDA-MIMA (2024, CMPB)~\cite{hu2024unsupervised} & 0.7913{\scriptsize\textcolor{customPurple}{$\pm$0.1458}} & 21.91{\scriptsize\textcolor{customPurple}{$\pm$26.37}} & 0.6767{\scriptsize\textcolor{customPurple}{$\pm$0.1842}} & 0.8260{\scriptsize\textcolor{customPurple}{$\pm$0.2347}} & 0.8426{\scriptsize\textcolor{customPurple}{$\pm$0.1579}} \\
    \rowcolor[HTML]{F0F0F0}
    \textcolor{green}{$\clubsuit$} GraTa (2025, AAAI)~\cite{chen2025gradient} & 0.8516{\scriptsize\textcolor{customPurple}{$\pm$0.1665}} & 13.96{\scriptsize\textcolor{customPurple}{$\pm$27.20}} & 0.7694{\scriptsize\textcolor{customPurple}{$\pm$0.1956}} & 0.8533{\scriptsize\textcolor{customPurple}{$\pm$0.2215}} & 0.9090{\scriptsize\textcolor{customPurple}{$\pm$0.0945}} \\
    \midrule
    \textcolor{red}{$\varheartsuit$} TENT (2021, ICLR)~\cite{wang2021tent} & 0.8558{\scriptsize\textcolor{customPurple}{$\pm$0.1627}} & 14.33{\scriptsize\textcolor{customPurple}{$\pm$35.04}} & 0.7794{\scriptsize\textcolor{customPurple}{$\pm$0.1911}} & 0.8565{\scriptsize\textcolor{customPurple}{$\pm$0.2141}} & 0.9107{\scriptsize\textcolor{customPurple}{$\pm$0.1195}} \\
    \textcolor{red}{$\varheartsuit$} UPL-SFDA (2023, TMI)~\cite{wu2023upl} & 0.8683{\scriptsize\textcolor{customPurple}{$\pm$0.1534}} & 13.26{\scriptsize\textcolor{customPurple}{$\pm$30.34}} & 0.7891{\scriptsize\textcolor{customPurple}{$\pm$0.1739}} & 0.8672{\scriptsize\textcolor{customPurple}{$\pm$0.1941}} & 0.8990{\scriptsize\textcolor{customPurple}{$\pm$0.1045}} \\
    \textcolor{red}{$\varheartsuit$} DG-TTA (2023, arXiv)~\cite{weihsbach2023dg} & \cellcolor{third}0.8725{\scriptsize\textcolor{customPurple}{$\pm$0.1481}} & 13.37{\scriptsize\textcolor{customPurple}{$\pm$40.52}} & \cellcolor{third}0.7962{\scriptsize\textcolor{customPurple}{$\pm$0.1744}} & 0.8586{\scriptsize\textcolor{customPurple}{$\pm$0.1938}} & 0.8947{\scriptsize\textcolor{customPurple}{$\pm$0.1311}} \\
    \rowcolor[HTML]{F0F0F0}
    \textcolor{red}{$\varheartsuit$} SaTTCA (2023, MICCAI)~\cite{li2023scale} & 0.8426{\scriptsize\textcolor{customPurple}{$\pm$0.0968}} & 14.71{\scriptsize\textcolor{customPurple}{$\pm$15.67}} & 0.7570{\scriptsize\textcolor{customPurple}{$\pm$0.1175}} & 0.8491{\scriptsize\textcolor{customPurple}{$\pm$0.1547}} & 0.9009{\scriptsize\textcolor{customPurple}{$\pm$0.0767}} \\
    \textcolor{red}{$\varheartsuit$} SmaRT (2025, arXiv)~\cite{Wang2025SmaRTSR} & \cellcolor{second}0.9108{\scriptsize\textcolor{customPurple}{$\pm$0.0688}} & \cellcolor{second}6.19{\scriptsize\textcolor{customPurple}{$\pm$13.86}} & \cellcolor{second}0.8427{\scriptsize\textcolor{customPurple}{$\pm$0.1048}} & \cellcolor{best}0.9679{\scriptsize\textcolor{customPurple}{$\pm$0.0628}} & 0.8721{\scriptsize\textcolor{customPurple}{$\pm$0.1112}} \\
    \midrule
    \rowcolor[HTML]{F0F0F0}
    \textcolor{blue}{$\vardiamondsuit$} AdaMI (2022, MICCAI)~\cite{bateson2022test} & 0.8425{\scriptsize\textcolor{customPurple}{$\pm$0.1738}} & 16.84{\scriptsize\textcolor{customPurple}{$\pm$42.26}} & 0.7597{\scriptsize\textcolor{customPurple}{$\pm$0.2136}} & 0.8438{\scriptsize\textcolor{customPurple}{$\pm$0.2300}} & 0.8970{\scriptsize\textcolor{customPurple}{$\pm$0.1342}} \\
        \rowcolor[HTML]{F0F0F0}
    \textcolor{blue}{$\vardiamondsuit$} PASS (2024, TMI)~\cite{zhang2024pass} & 0.8426{\scriptsize\textcolor{customPurple}{$\pm$0.1736}} & 15.90{\scriptsize\textcolor{customPurple}{$\pm$35.99}} & 0.7577{\scriptsize\textcolor{customPurple}{$\pm$0.2014}} & 0.8445{\scriptsize\textcolor{customPurple}{$\pm$0.2296}} & 0.9013{\scriptsize\textcolor{customPurple}{$\pm$0.1188}} \\
    \textcolor{blue}{$\vardiamondsuit$} VPTTA (2024, CVPR)~\cite{chen2024each} & 0.8562{\scriptsize\textcolor{customPurple}{$\pm$0.1708}} & 16.96{\scriptsize\textcolor{customPurple}{$\pm$27.20}} & 0.7738{\scriptsize\textcolor{customPurple}{$\pm$0.1997}} & 0.8576{\scriptsize\textcolor{customPurple}{$\pm$0.2253}} & 0.9132{\scriptsize\textcolor{customPurple}{$\pm$0.0982}} \\

    \rowcolor[HTML]{F0F0F0}
    \textcolor{blue}{$\vardiamondsuit$} ExploringTTA (2025, ISBI)~\cite{omolegan2025exploring} & 0.8519{\scriptsize\textcolor{customPurple}{$\pm$0.1687}} & 14.81{\scriptsize\textcolor{customPurple}{$\pm$35.29}} & 0.7704{\scriptsize\textcolor{customPurple}{$\pm$0.1971}} & 0.8533{\scriptsize\textcolor{customPurple}{$\pm$0.2221}} & 0.9112{\scriptsize\textcolor{customPurple}{$\pm$0.1147}} \\
    \textcolor{blue}{$\vardiamondsuit$} ProSFDA (2026, PR)~\cite{hu2025source} & 0.8696{\scriptsize\textcolor{customPurple}{$\pm$0.1579}} & 14.41{\scriptsize\textcolor{customPurple}{$\pm$41.43}} & 0.7947{\scriptsize\textcolor{customPurple}{$\pm$0.1852}} & 0.8593{\scriptsize\textcolor{customPurple}{$\pm$0.2055}} & 0.9191{\scriptsize\textcolor{customPurple}{$\pm$0.1199}} \\
    \bottomrule
  \end{tabular}%
  }
\end{table}

\subsubsection{Chest X-ray Modality}

\begin{figure}[h!]
  \centering
  \includegraphics[width=\linewidth]{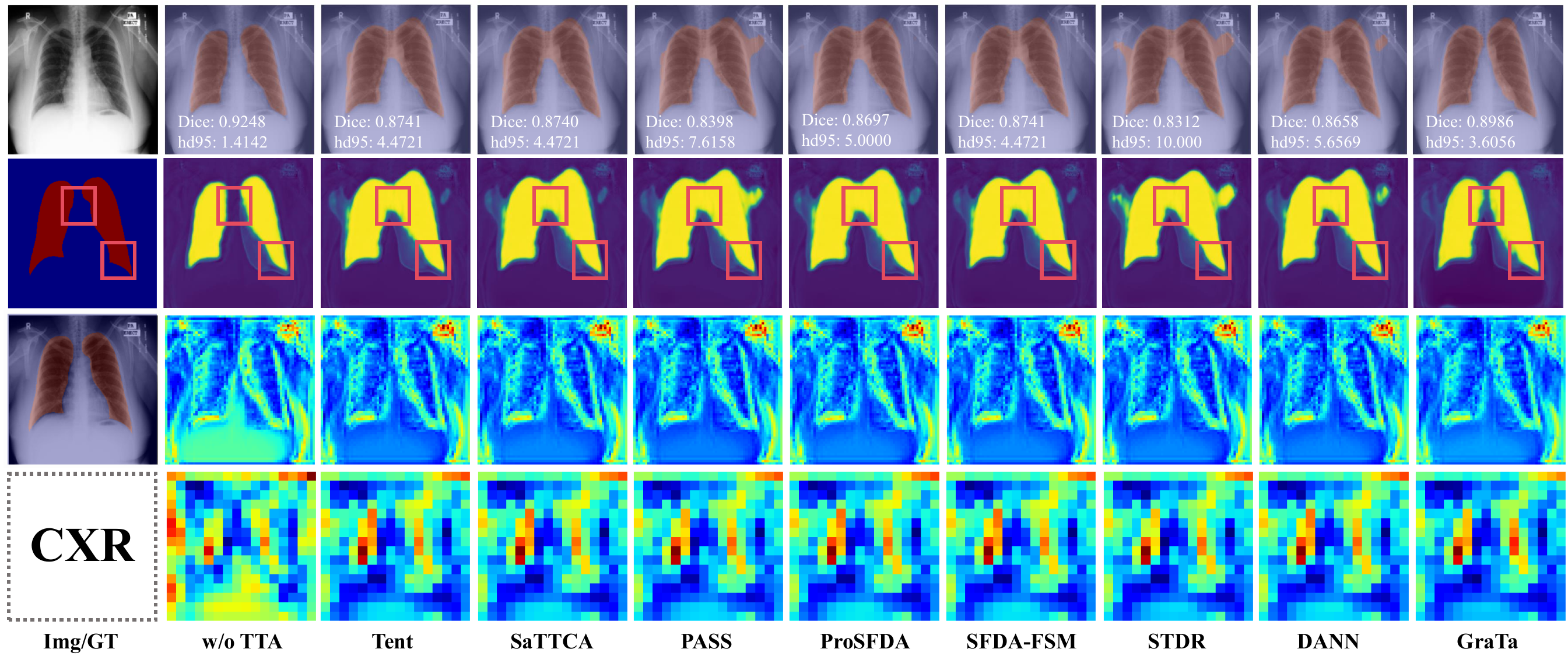}
  \caption{\textbf{Visual comparison of TTA efficacy across different methods on CXR datasets.} Rows 1 and 2 illustrate the segmentation outcomes. Rows 3 and 4 depict changes in features extracted by the Encoder after TTA. GT denotes the ground-truth.}
  \label{fig: CXR_seg}
\end{figure}

In the Chest X-ray task shown in Table~\ref{tab:cxr}, most methods achieved substantial improvements, with 17 out of 20 surpassing the baseline Dice score on this challenging benchmark. SFDA-FSM again led the pack with a Dice score of 0.9663, representing an improvement of 6.3 percent, and an HD95 of 0.5576, a reduction of 89 percent. This was followed by ProSFDA, DG-TTA, AdaMI, and SaTTCA. Nevertheless, a few methods exhibited instability. For example, VPTTA slightly underperformed the baseline in Dice, while TestFit and UPL-SFDA showed worse HD95 values. Overall, all four paradigms demonstrated adaptive capability on CXR, though Prior Estimation and feature alignment approaches were more sensitive to parameter settings and data distribution. Fig~\ref{fig: CXR_seg} illustrates the effects of various TTA methods on the CXR task and their influence on the encoder. The majority of models exhibit adverse effects.

\begin{table}[h!]

  \centering
  \caption{\textbf{Comprehensive performance comparison of TTA methods on Lung Segmentation in Chest X-ray Images.}
 Results colored in \colorbox{red!30}{\makebox[1.3em][c]{\raisebox{0pt}[0.5em][0em]{red}}}, \colorbox{orange!30}{\makebox[2.8em][c]{\raisebox{0pt}[0.5em][0em]{orange}}}, and \colorbox{yellow!30}{\makebox[2.5em][c]{\raisebox{0pt}[0.5em][0em]{yellow}}} denote the best, second-best, and third-best performances, respectively. Rows with a \colorbox{gray!30}{\makebox[1.6em][c]{\raisebox{0pt}[0.5em][0em]{gray}}} background indicate Dice values lower than that of the Target-Domain(w/o TTA). Symbols denote paradigms: \textcolor{black}{\(\spadesuit\)} for Input-level Translation, \textcolor{green}{\(\clubsuit\)} for Feature-level Alignment, \textcolor{red}{\(\varheartsuit\)} for Output-level Regularization, \textcolor{blue}{\(\vardiamondsuit\)} for Prior Estimation. All improvements are statistically significant at \( p < 0.05 \).}
  \vspace{0.5em}
  \label{tab:cxr}
  \resizebox{\textwidth}{!}{%
  \begin{tabular}{lccccc}
    \arrayrulecolor{black}\hline
    \rowcolor[HTML]{D8D6C0}
    \textbf{Method} & \textbf{Dice $\uparrow$} & \textbf{HD95 $\downarrow$} & \textbf{JI $\uparrow$} & \textbf{Sen $\uparrow$} & \textbf{PPV $\uparrow$} \arraybackslash \\
    \arrayrulecolor{black}\hline
    \textbf{Intra-Domain} & \textbf{0.9557{\scriptsize\textcolor{customPurple}{$\pm$0.0057}}} & \textbf{1.02{\scriptsize\textcolor{customPurple}{$\pm$0.30}}} & \textbf{0.9170{\scriptsize\textcolor{customPurple}{$\pm$0.0095}}} & \textbf{0.9573{\scriptsize\textcolor{customPurple}{$\pm$0.0077}}} & \textbf{0.9569{\scriptsize\textcolor{customPurple}{$\pm$0.0023}}} \\
  \hline
    \textbf{Target-Domain (w/o TTA)} & \textbf{0.9093{\scriptsize\textcolor{customPurple}{$\pm$0.0695}}} & \textbf{5.15{\scriptsize\textcolor{customPurple}{$\pm$9.11}}} & \textbf{0.8414{\scriptsize\textcolor{customPurple}{$\pm$0.1038}}} & \textbf{0.8534{\scriptsize\textcolor{customPurple}{$\pm$0.1059}}} & \textbf{0.9846{\scriptsize\textcolor{customPurple}{$\pm$0.0272}}} \\
    \midrule
    \textcolor{black}{$\spadesuit$} RSA (2021, MICCAI)~\cite{zeng2024reliable} & 0.9183{\scriptsize\textcolor{customPurple}{$\pm$0.0892}} & 4.64{\scriptsize\textcolor{customPurple}{$\pm$9.46}} & 0.8781{\scriptsize\textcolor{customPurple}{$\pm$0.0848}} & 0.9531{\scriptsize\textcolor{customPurple}{$\pm$0.0423}} & 0.9245{\scriptsize\textcolor{customPurple}{$\pm$0.1048}} \\
    \textcolor{black}{$\spadesuit$} DL-TTA (2022, TMI)~\cite{yang2022dltta} & 0.9162{\scriptsize\textcolor{customPurple}{$\pm$0.0782}} & 4.36{\scriptsize\textcolor{customPurple}{$\pm$10.38}} & 0.8863{\scriptsize\textcolor{customPurple}{$\pm$0.0818}} & 0.9408{\scriptsize\textcolor{customPurple}{$\pm$0.0513}} & 0.9131{\scriptsize\textcolor{customPurple}{$\pm$0.0865}} \\
    \textcolor{black}{$\spadesuit$} SFDA-FSM (2022, MIA)~\cite{yang2022source} & \cellcolor{best}0.9663{\scriptsize\textcolor{customPurple}{$\pm$0.0223}} & \cellcolor{best}0.56{\scriptsize\textcolor{customPurple}{$\pm$3.01}} & \cellcolor{best}0.9357{\scriptsize\textcolor{customPurple}{$\pm$0.0390}} & \cellcolor{best}0.9648{\scriptsize\textcolor{customPurple}{$\pm$0.0398}} & \cellcolor{second}0.9690{\scriptsize\textcolor{customPurple}{$\pm$0.0151}} \\
    \textcolor{black}{$\spadesuit$} STDR (2024, TMI)~\cite{wang2024dual} & 0.9260{\scriptsize\textcolor{customPurple}{$\pm$0.0674}} & 3.87{\scriptsize\textcolor{customPurple}{$\pm$8.13}} & 0.8662{\scriptsize\textcolor{customPurple}{$\pm$0.0913}} & 0.9478{\scriptsize\textcolor{customPurple}{$\pm$0.0429}} & 0.9100{\scriptsize\textcolor{customPurple}{$\pm$0.0876}} \\
    \textcolor{black}{$\spadesuit$} AIF-SFDA (2025, AAAI)~\cite{li2025aif} & 0.9274{\scriptsize\textcolor{customPurple}{$\pm$0.0205}} & 3.39{\scriptsize\textcolor{customPurple}{$\pm$4.20}} & 0.8755{\scriptsize\textcolor{customPurple}{$\pm$0.0324}} & 0.9531{\scriptsize\textcolor{customPurple}{$\pm$0.0135}} & 0.9101{\scriptsize\textcolor{customPurple}{$\pm$0.0377}} \\
    \midrule
    \textcolor{green}{$\clubsuit$} DANN (2016, JMLR)~\cite{ganin2016domain} & 0.9344{\scriptsize\textcolor{customPurple}{$\pm$0.0632}} & 3.51{\scriptsize\textcolor{customPurple}{$\pm$10.14}} & 0.8824{\scriptsize\textcolor{customPurple}{$\pm$0.0932}} & \cellcolor{third}0.9608{\scriptsize\textcolor{customPurple}{$\pm$0.0418}} & 0.9167{\scriptsize\textcolor{customPurple}{$\pm$0.0940}} \\
        \textcolor{green}{$\clubsuit$} DeTTA (2024, WACV)~\cite{wen2024denoising} & 0.9095{\scriptsize\textcolor{customPurple}{$\pm$0.0716}} & 4.35{\scriptsize\textcolor{customPurple}{$\pm$10.37}} & 0.8796{\scriptsize\textcolor{customPurple}{$\pm$0.0751}} & 0.9341{\scriptsize\textcolor{customPurple}{$\pm$0.0384}} & 0.9064{\scriptsize\textcolor{customPurple}{$\pm$0.0798}} \\
    \textcolor{green}{$\clubsuit$} TestFit (2024, MIA)~\cite{zhang2024testfit} & 0.9120{\scriptsize\textcolor{customPurple}{$\pm$0.0849}} & 5.59{\scriptsize\textcolor{customPurple}{$\pm$11.62}} & 0.8477{\scriptsize\textcolor{customPurple}{$\pm$0.1230}} & 0.9208{\scriptsize\textcolor{customPurple}{$\pm$0.1117}} & 0.9409{\scriptsize\textcolor{customPurple}{$\pm$0.1022}} \\

    \textcolor{green}{$\clubsuit$} UDA-MIMA (2024, CMPB)~\cite{hu2024unsupervised} & 0.9199{\scriptsize\textcolor{customPurple}{$\pm$0.0802}} & 5.26{\scriptsize\textcolor{customPurple}{$\pm$12.70}} & 0.8602{\scriptsize\textcolor{customPurple}{$\pm$0.1159}} & 0.9567{\scriptsize\textcolor{customPurple}{$\pm$0.0415}} & 0.8964{\scriptsize\textcolor{customPurple}{$\pm$0.1214}} \\
    \textcolor{green}{$\clubsuit$} GraTa (2025, AAAI)~\cite{chen2025gradient} & 0.9212{\scriptsize\textcolor{customPurple}{$\pm$0.0733}} & 4.73{\scriptsize\textcolor{customPurple}{$\pm$11.46}} & 0.8611{\scriptsize\textcolor{customPurple}{$\pm$0.1074}} & 0.9478{\scriptsize\textcolor{customPurple}{$\pm$0.0440}} & 0.9061{\scriptsize\textcolor{customPurple}{$\pm$0.1145}} \\
    \midrule
    \textcolor{red}{$\varheartsuit$} TENT (2021, ICLR)~\cite{wang2021tent} & 0.9325{\scriptsize\textcolor{customPurple}{$\pm$0.0630}} & 3.69{\scriptsize\textcolor{customPurple}{$\pm$10.20}} & 0.8791{\scriptsize\textcolor{customPurple}{$\pm$0.0942}} & 0.9607{\scriptsize\textcolor{customPurple}{$\pm$0.0400}} & 0.9133{\scriptsize\textcolor{customPurple}{$\pm$0.0961}} \\
    \rowcolor[HTML]{F0F0F0}
    \textcolor{red}{$\varheartsuit$} UPL-SFDA (2023, TMI)~\cite{wu2023upl} & 0.9042{\scriptsize\textcolor{customPurple}{$\pm$0.0831}} & 5.47{\scriptsize\textcolor{customPurple}{$\pm$10.45}} & 0.8557{\scriptsize\textcolor{customPurple}{$\pm$0.0953}} & 0.9243{\scriptsize\textcolor{customPurple}{$\pm$0.0536}} & 0.9147{\scriptsize\textcolor{customPurple}{$\pm$0.0857}} \\
    \textcolor{red}{$\varheartsuit$} DG-TTA (2023, arXiv)~\cite{weihsbach2023dg} & 0.9420{\scriptsize\textcolor{customPurple}{$\pm$0.0458}} & 2.92{\scriptsize\textcolor{customPurple}{$\pm$7.72}} & 0.8936{\scriptsize\textcolor{customPurple}{$\pm$0.0757}} & 0.9175{\scriptsize\textcolor{customPurple}{$\pm$0.0685}} & \cellcolor{best}0.9720{\scriptsize\textcolor{customPurple}{$\pm$0.0434}} \\
    \textcolor{red}{$\varheartsuit$} SaTTCA (2023, MICCAI)~\cite{li2023scale} & 0.9348{\scriptsize\textcolor{customPurple}{$\pm$0.0272}} & 3.40{\scriptsize\textcolor{customPurple}{$\pm$4.21}} & 0.8829{\scriptsize\textcolor{customPurple}{$\pm$0.0397}} & 0.9605{\scriptsize\textcolor{customPurple}{$\pm$0.0208}} & 0.9175{\scriptsize\textcolor{customPurple}{$\pm$0.0450}} \\
    \textcolor{red}{$\varheartsuit$} SmaRT (2025, arXiv)~\cite{Wang2025SmaRTSR} & \cellcolor{second}0.9501{\scriptsize\textcolor{customPurple}{$\pm$0.0288}} & \cellcolor{second}1.77{\scriptsize\textcolor{customPurple}{$\pm$6.10}} & \cellcolor{second}0.9064{\scriptsize\textcolor{customPurple}{$\pm$0.0496}} & 0.9572{\scriptsize\textcolor{customPurple}{$\pm$0.0413}} & 0.9449{\scriptsize\textcolor{customPurple}{$\pm$0.0364}} \\
    \midrule
    \textcolor{blue}{$\vardiamondsuit$} AdaMI (2022, MICCAI)~\cite{bateson2022test} & 0.9363{\scriptsize\textcolor{customPurple}{$\pm$0.0628}} & 3.49{\scriptsize\textcolor{customPurple}{$\pm$10.09}} & 0.8857{\scriptsize\textcolor{customPurple}{$\pm$0.0926}} & 0.9576{\scriptsize\textcolor{customPurple}{$\pm$0.0425}} & 0.9232{\scriptsize\textcolor{customPurple}{$\pm$0.0933}} \\
        \textcolor{blue}{$\vardiamondsuit$} PASS (2024, TMI)~\cite{zhang2024pass} & 0.9346{\scriptsize\textcolor{customPurple}{$\pm$0.0632}} & 3.50{\scriptsize\textcolor{customPurple}{$\pm$10.13}} & 0.8827{\scriptsize\textcolor{customPurple}{$\pm$0.0931}} & 0.9606{\scriptsize\textcolor{customPurple}{$\pm$0.0418}} & 0.9173{\scriptsize\textcolor{customPurple}{$\pm$0.0939}} \\
    \rowcolor[HTML]{F0F0F0}
    \textcolor{blue}{$\vardiamondsuit$} VPTTA (2024, CVPR)~\cite{chen2024each} & 0.8978{\scriptsize\textcolor{customPurple}{$\pm$0.0765}} & 4.87{\scriptsize\textcolor{customPurple}{$\pm$10.44}} & 0.8492{\scriptsize\textcolor{customPurple}{$\pm$0.0886}} & 0.9177{\scriptsize\textcolor{customPurple}{$\pm$0.0469}} & 0.9081{\scriptsize\textcolor{customPurple}{$\pm$0.0791}} \\

    \textcolor{blue}{$\vardiamondsuit$} ExploringTTA (2025, ISBI)~\cite{omolegan2025exploring} & 0.9286{\scriptsize\textcolor{customPurple}{$\pm$0.0711}} & 3.89{\scriptsize\textcolor{customPurple}{$\pm$10.67}} & 0.8734{\scriptsize\textcolor{customPurple}{$\pm$0.1014}} & \cellcolor{second}0.9625{\scriptsize\textcolor{customPurple}{$\pm$0.0402}} & 0.9058{\scriptsize\textcolor{customPurple}{$\pm$0.1040}} \\
    \textcolor{blue}{$\vardiamondsuit$} ProSFDA (2026, PR)~\cite{hu2025source} & \cellcolor{third}0.9451{\scriptsize\textcolor{customPurple}{$\pm$0.0445}} & \cellcolor{third}2.56{\scriptsize\textcolor{customPurple}{$\pm$7.64}} & \cellcolor{third}0.8990{\scriptsize\textcolor{customPurple}{$\pm$0.0716}} & 0.9402{\scriptsize\textcolor{customPurple}{$\pm$0.0563}} & \cellcolor{third}0.9549{\scriptsize\textcolor{customPurple}{$\pm$0.0603}} \\
    \bottomrule
  \end{tabular}%
  }
\end{table}

\subsection{TTA Under Strong Domain Shift}

In scenarios with substantial domain shifts or severe noise interference, such as those involving US, MRI, and CT, Input-level Transformation retained its overall advantage in terms of Dice scores, whereas the best HD95 performance typically came from feature alignment or output regularization. In particular, in 3D MRI and CT, appearance alignment alone was insufficient to ensure cross-patch consistency, and generative model-based approaches incurred high computational costs and low inference efficiency. By contrast, Output-level Regularization strategies, such as entropy minimization, showed greater robustness to heavy noise.

\subsubsection{Magnetic Resonance Imaging Modality}

\begin{figure}[h!]
  \centering
  \includegraphics[width=\linewidth]{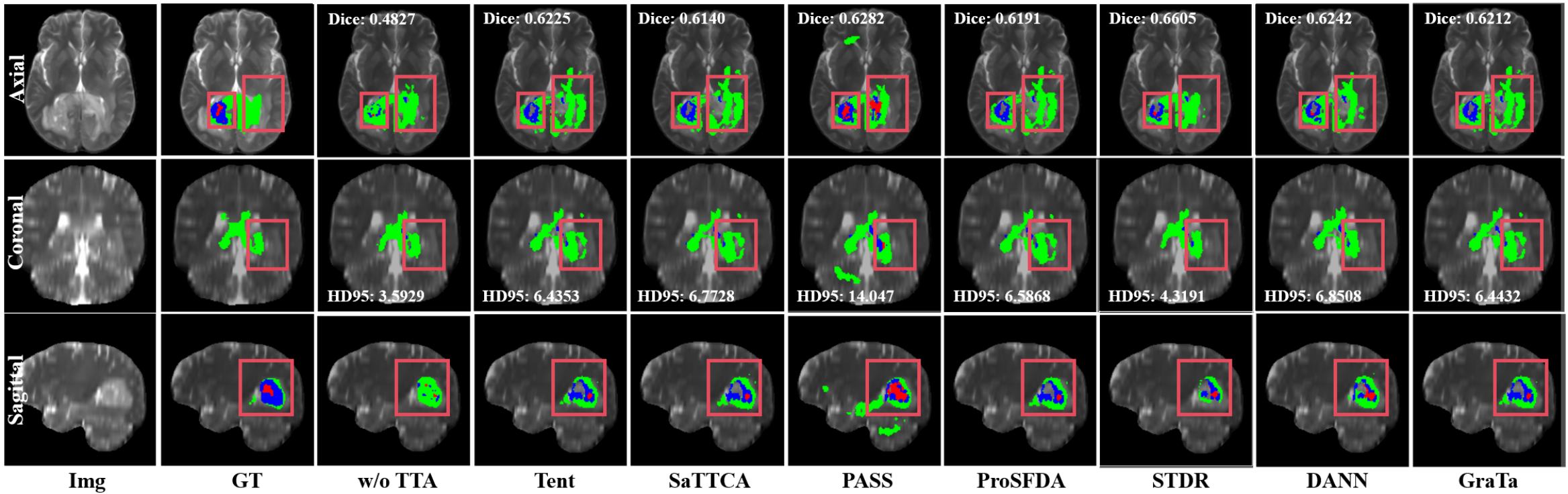}
  \caption{\textbf{Visual comparison of TTA efficacy across different methods on MRI datasets.} The three rows represent slices taken from three different orientations. GT denotes the ground-truth. The colors \textcolor{red}{red}, \textcolor{green}{green}, and \textcolor{blue}{blue} represent categories 1, 2, and 3, respectively.}
  \label{fig: MRI_seg}
\end{figure}

In the Magnetic Resonance Imaging task shown in the Table~\ref{tab:mri-dice-hd95} and Table~\ref{tab:mri-sen-ppv}, Input-level Transformation methods achieved the most notable improvement in average Dice, with DL-TTA performing the best. The average Dice reached 0.6675, representing a 12.2\% increase. Within subregions, ET and TC improved by 0.1209 and 0.1007, respectively, while WT decreased slightly by 0.0045. SmaRT achieved the largest reduction in HD95, bringing it to 4.4402, an 80.7\% decrease. Notably, only half of the methods outperformed the baseline in the WT subregion. This is primarily because WT relies heavily on the high FLAIR signal, which exhibits diffuse boundaries and significant inter-center variability. This reliance leads entropy-minimization and uncertainty-filtering approaches to focus on core regions while contracting peripheral edema areas, resulting in mild under-segmentation. In contrast, ET and TC regions exhibit stronger contrast and more compact morphology, yielding more stable pseudo-labels and more consistent convergence.
However, it is essential to note that the baseline model used for the TTA methods is the standard UNet3D, which shows limited representation ability in more complex task scenarios, such as BraTS. As a result, its performance is subpar in the source domain. Nevertheless, when directly tested on the target domain, a clear domain shift is observed, with significant declines in metrics such as Dice and HD95. The performance improvements observed with other methods listed on the leaderboard further validate the effectiveness of the TTA approach. Fig.~\ref{fig: MRI_seg} presents the effects of various TTA methods on the MRI task. Most models demonstrate a positive impact on the Dice metric while negatively affecting the HD95 metric.

\begin{table}[h!]
  \centering
  \fontsize{7.5}{8}\selectfont 
  \setlength{\tabcolsep}{1.5mm}
  \renewcommand{\arraystretch}{1.3}  
  \caption{\textbf{Comprehensive performance comparison of TTA methods on MRI brain tumor segmentation: Dice and HD95 across ET/TC/WT.}
 Results colored in \colorbox{red!30}{\makebox[1.3em][c]{\raisebox{0pt}[0.5em][0em]{red}}}, \colorbox{orange!30}{\makebox[2.8em][c]{\raisebox{0pt}[0.5em][0em]{orange}}}, and \colorbox{yellow!30}{\makebox[2.5em][c]{\raisebox{0pt}[0.5em][0em]{yellow}}} denote the best, second-best, and third-best performances, respectively. Rows with a \colorbox{gray!30}{\makebox[1.6em][c]{\raisebox{0pt}[0.5em][0em]{gray}}} background indicate Dice values lower than that of the Target-Domain(w/o TTA). Symbols denote paradigms: \textcolor{black}{\(\spadesuit\)} for Input-level Translation, \textcolor{green}{\(\clubsuit\)} for Feature-level Alignment, \textcolor{red}{\(\varheartsuit\)} for Output-level Regularization, \textcolor{blue}{\(\vardiamondsuit\)} for Prior Estimation. All improvements are statistically significant at \( p < 0.05 \).}
  \label{tab:mri-dice-hd95}
  \resizebox{\textwidth}{!}{%
  \begin{tabular}{lcccccc}  
   \hline  
    \rowcolor{customHeader}
    \multicolumn{1}{l}{\cellcolor{customHeader}} & \multicolumn{3}{c}{\cellcolor{customHeader}\textbf{Dice$\uparrow$}} & \multicolumn{3}{c}{\cellcolor{customHeader}\textbf{HD95$\downarrow$}} \\  
    \cline{2-7}  
    \rowcolor{customHeader}
    \multicolumn{1}{l}{\multirow{-2}{*}{\cellcolor{customHeader}\textbf{Method}}} & \textbf{ET} & \textbf{TC} & \textbf{WT} & \textbf{ET} & \textbf{TC} & \textbf{WT} \\  
   \hline
    \textbf{Intra-Domain}
      & \textbf{0.6320{\tiny\textcolor{customPurple}{$\pm$0.1664}}} & \textbf{0.6379{\tiny\textcolor{customPurple}{$\pm$0.1706}}} & \textbf{0.8480{\tiny\textcolor{customPurple}{$\pm$0.0542}}}
      & \textbf{4.40{\tiny\textcolor{customPurple}{$\pm$9.63}}} & \textbf{4.40{\tiny\textcolor{customPurple}{$\pm$9.58}}} & \textbf{1.20{\tiny\textcolor{customPurple}{$\pm$0.74}}} \\
    \hline
    \textbf{Target-Domain (w/o TTA)}
      & \textbf{0.5232{\tiny\textcolor{customPurple}{$\pm$0.1279}}} & \textbf{0.5105{\tiny\textcolor{customPurple}{$\pm$0.1217}}} & \textbf{0.7468{\tiny\textcolor{customPurple}{$\pm$0.0834}}}
      & \textbf{26.85{\tiny\textcolor{customPurple}{$\pm$2.31}}} & \textbf{37.04{\tiny\textcolor{customPurple}{$\pm$3.27}}} & \textbf{15.22{\tiny\textcolor{customPurple}{$\pm$8.34}}} \\
    \hline
    \textcolor{black}{$\spadesuit$} RSA (2021, MICCAI)~\cite{zeng2024reliable} & 0.5527{\tiny\textcolor{customPurple}{$\pm$0.1681}} & 0.5061{\tiny\textcolor{customPurple}{$\pm$0.1267}} & \cellcolor{best}0.7593{\tiny\textcolor{customPurple}{$\pm$0.1580}} & 17.65{\tiny\textcolor{customPurple}{$\pm$4.31}} & 17.67{\tiny\textcolor{customPurple}{$\pm$4.14}} & 15.80{\tiny\textcolor{customPurple}{$\pm$13.03}} \\
    \textcolor{black}{$\spadesuit$} DL-TTA (2022, TMI)~\cite{yang2022dltta} & \cellcolor{best}0.6441{\tiny\textcolor{customPurple}{$\pm$0.1921}} & \cellcolor{best}0.6160{\tiny\textcolor{customPurple}{$\pm$0.1834}} & 0.7423{\tiny\textcolor{customPurple}{$\pm$0.0731}} & 14.31{\tiny\textcolor{customPurple}{$\pm$8.39}} & 14.28{\tiny\textcolor{customPurple}{$\pm$8.94}} & 9.35{\tiny\textcolor{customPurple}{$\pm$7.31}} \\
    \textcolor{black}{$\spadesuit$} SFDA-FSM (2022, MIA)~\cite{yang2022source} & 0.6118{\tiny\textcolor{customPurple}{$\pm$0.2369}} & 0.5936{\tiny\textcolor{customPurple}{$\pm$0.2754}} & 0.6764{\tiny\textcolor{customPurple}{$\pm$0.1287}} & 18.39{\tiny\textcolor{customPurple}{$\pm$48.30}} & 16.82{\tiny\textcolor{customPurple}{$\pm$48.07}} & 12.27{\tiny\textcolor{customPurple}{$\pm$12.87}} \\
    \textcolor{black}{$\spadesuit$} STDR (2024, TMI)~\cite{wang2024dual} & 0.5438{\tiny\textcolor{customPurple}{$\pm$0.1749}} & 0.5187{\tiny\textcolor{customPurple}{$\pm$0.1743}} & 0.7052{\tiny\textcolor{customPurple}{$\pm$0.1289}} & 15.17{\tiny\textcolor{customPurple}{$\pm$4.93}} & 14.88{\tiny\textcolor{customPurple}{$\pm$4.46}} & 12.29{\tiny\textcolor{customPurple}{$\pm$4.86}} \\
    \textcolor{black}{$\spadesuit$} AIF-SFDA (2025, AAAI)~\cite{li2025aif} & \cellcolor{third}0.6375{\tiny\textcolor{customPurple}{$\pm$0.1856}} & \cellcolor{second}0.6092{\tiny\textcolor{customPurple}{$\pm$0.1768}} & 0.7342{\tiny\textcolor{customPurple}{$\pm$0.0734}} & 10.30{\tiny\textcolor{customPurple}{$\pm$8.38}} & 9.27{\tiny\textcolor{customPurple}{$\pm$4.93}} & 7.34{\tiny\textcolor{customPurple}{$\pm$7.30}} \\
    \hline
    \textcolor{green}{$\clubsuit$} DANN (2016, JMLR)~\cite{ganin2016domain} & 0.5394{\tiny\textcolor{customPurple}{$\pm$0.1522}} & 0.5006{\tiny\textcolor{customPurple}{$\pm$0.0838}} & 0.7056{\tiny\textcolor{customPurple}{$\pm$0.1420}} & 12.65{\tiny\textcolor{customPurple}{$\pm$10.38}} & 12.49{\tiny\textcolor{customPurple}{$\pm$6.00}} & 14.20{\tiny\textcolor{customPurple}{$\pm$7.36}} \\
        \textcolor{green}{$\clubsuit$} DeTTA (2024, WACV)~\cite{wen2024denoising} & 0.5975{\tiny\textcolor{customPurple}{$\pm$0.1371}} & 0.5742{\tiny\textcolor{customPurple}{$\pm$0.0953}} & 0.7489{\tiny\textcolor{customPurple}{$\pm$0.0843}} & 12.54{\tiny\textcolor{customPurple}{$\pm$5.59}} & 11.35{\tiny\textcolor{customPurple}{$\pm$5.52}} & 8.43{\tiny\textcolor{customPurple}{$\pm$3.93}} \\
            \textcolor{green}{$\clubsuit$} TestFit (2024, MIA)~\cite{zhang2024testfit} & 0.5669{\tiny\textcolor{customPurple}{$\pm$0.1083}} & 0.5756{\tiny\textcolor{customPurple}{$\pm$0.0837}} & \cellcolor{third}0.7508{\tiny\textcolor{customPurple}{$\pm$0.0251}} & 16.11{\tiny\textcolor{customPurple}{$\pm$3.40}} & \cellcolor{third}6.87{\tiny\textcolor{customPurple}{$\pm$3.60}} & 2.51{\tiny\textcolor{customPurple}{$\pm$1.82}} \\
    \textcolor{green}{$\clubsuit$} UDA-MIMA (2024, CMPB)~\cite{hu2024unsupervised} & 0.5609{\tiny\textcolor{customPurple}{$\pm$0.2135}} & 0.5505{\tiny\textcolor{customPurple}{$\pm$0.1488}} & 0.7373{\tiny\textcolor{customPurple}{$\pm$0.0763}} & 16.65{\tiny\textcolor{customPurple}{$\pm$9.41}} & 15.56{\tiny\textcolor{customPurple}{$\pm$9.99}} & 7.63{\tiny\textcolor{customPurple}{$\pm$9.93}} \\
    \textcolor{green}{$\clubsuit$} GraTa (2025, AAAI)~\cite{chen2025gradient} & 0.5996{\tiny\textcolor{customPurple}{$\pm$0.1413}} & 0.5842{\tiny\textcolor{customPurple}{$\pm$0.1242}} & 0.7379{\tiny\textcolor{customPurple}{$\pm$0.0396}} & 6.84{\tiny\textcolor{customPurple}{$\pm$5.00}} & 7.29{\tiny\textcolor{customPurple}{$\pm$4.14}} & 3.96{\tiny\textcolor{customPurple}{$\pm$3.95}} \\
    \hline
    \textcolor{red}{$\varheartsuit$} TENT (2021, ICLR)~\cite{wang2021tent} & 0.6145{\tiny\textcolor{customPurple}{$\pm$0.1164}} & \cellcolor{third}0.6010{\tiny\textcolor{customPurple}{$\pm$0.0971}} & 0.7291{\tiny\textcolor{customPurple}{$\pm$0.0279}} & \cellcolor{second}5.49{\tiny\textcolor{customPurple}{$\pm$2.44}} & \cellcolor{second}5.77{\tiny\textcolor{customPurple}{$\pm$2.43}} & \cellcolor{best}2.79{\tiny\textcolor{customPurple}{$\pm$1.73}} \\
    \textcolor{red}{$\varheartsuit$} UPL-SFDA (2023, TMI)~\cite{wu2023upl} & 0.5416{\tiny\textcolor{customPurple}{$\pm$0.1842}} & 0.5065{\tiny\textcolor{customPurple}{$\pm$0.1337}} & 0.7398{\tiny\textcolor{customPurple}{$\pm$0.0957}} & 17.30{\tiny\textcolor{customPurple}{$\pm$14.83}} & 17.15{\tiny\textcolor{customPurple}{$\pm$8.30}} & 9.58{\tiny\textcolor{customPurple}{$\pm$6.79}} \\
    \textcolor{red}{$\varheartsuit$} DG-TTA (2023, arXiv)~\cite{weihsbach2023dg} & 0.5995{\tiny\textcolor{customPurple}{$\pm$0.1409}} & 0.5839{\tiny\textcolor{customPurple}{$\pm$0.1140}} & 0.7384{\tiny\textcolor{customPurple}{$\pm$0.0394}} & 6.95{\tiny\textcolor{customPurple}{$\pm$2.00}} & 7.38{\tiny\textcolor{customPurple}{$\pm$4.14}} & 3.94{\tiny\textcolor{customPurple}{$\pm$4.00}} \\
    \textcolor{red}{$\varheartsuit$} SaTTCA (2023, MICCAI)~\cite{li2023scale} & 0.5929{\tiny\textcolor{customPurple}{$\pm$0.1399}} & 0.5717{\tiny\textcolor{customPurple}{$\pm$0.1423}} & 0.7395{\tiny\textcolor{customPurple}{$\pm$0.0448}} & 7.35{\tiny\textcolor{customPurple}{$\pm$5.10}} & 7.64{\tiny\textcolor{customPurple}{$\pm$5.22}} & 4.48{\tiny\textcolor{customPurple}{$\pm$4.09}} \\
    \textcolor{red}{$\varheartsuit$} SmaRT (2025, arXiv)~\cite{Wang2025SmaRTSR} & \cellcolor{second}0.6403{\tiny\textcolor{customPurple}{$\pm$0.2392}} & 0.6002{\tiny\textcolor{customPurple}{$\pm$0.2794}} & \cellcolor{second}0.7520{\tiny\textcolor{customPurple}{$\pm$0.0279}} & \cellcolor{best}4.80{\tiny\textcolor{customPurple}{$\pm$4.52}} & \cellcolor{best}5.36{\tiny\textcolor{customPurple}{$\pm$4.98}} & \cellcolor{second}3.16{\tiny\textcolor{customPurple}{$\pm$2.79}} \\
    \hline
    \textcolor{blue}{$\vardiamondsuit$} AdaMI (2022, MICCAI)~\cite{bateson2022test} & 0.5906{\tiny\textcolor{customPurple}{$\pm$0.1481}} & 0.5631{\tiny\textcolor{customPurple}{$\pm$0.1057}} & 0.7481{\tiny\textcolor{customPurple}{$\pm$0.0367}} & 6.66{\tiny\textcolor{customPurple}{$\pm$5.14}} & 7.64{\tiny\textcolor{customPurple}{$\pm$5.22}} & 3.62{\tiny\textcolor{customPurple}{$\pm$4.53}} \\
       \textcolor{blue}{$\vardiamondsuit$} PASS (2024, TMI)~\cite{zhang2024pass} & 0.5983{\tiny\textcolor{customPurple}{$\pm$0.1083}} & 0.5736{\tiny\textcolor{customPurple}{$\pm$0.0977}} & 0.7365{\tiny\textcolor{customPurple}{$\pm$0.0347}} & 6.64{\tiny\textcolor{customPurple}{$\pm$5.60}} & 7.46{\tiny\textcolor{customPurple}{$\pm$5.41}} & 3.47{\tiny\textcolor{customPurple}{$\pm$4.04}} \\
    \textcolor{blue}{$\vardiamondsuit$} VPTTA (2024, CVPR)~\cite{chen2024each} & 0.5962{\tiny\textcolor{customPurple}{$\pm$0.1365}} & 0.5724{\tiny\textcolor{customPurple}{$\pm$0.0946}} & 0.7475{\tiny\textcolor{customPurple}{$\pm$0.0834}} & 13.54{\tiny\textcolor{customPurple}{$\pm$5.59}} & 10.35{\tiny\textcolor{customPurple}{$\pm$5.19}} & 8.43{\tiny\textcolor{customPurple}{$\pm$3.93}} \\
    \textcolor{blue}{$\vardiamondsuit$} ExploringTTA (2025, ISBI)~\cite{omolegan2025exploring} & 0.6038{\tiny\textcolor{customPurple}{$\pm$0.1433}} & 0.5805{\tiny\textcolor{customPurple}{$\pm$0.1473}} & 0.7443{\tiny\textcolor{customPurple}{$\pm$0.0396}} & \cellcolor{third}6.55{\tiny\textcolor{customPurple}{$\pm$5.60}} & 7.36{\tiny\textcolor{customPurple}{$\pm$5.15}} & \cellcolor{third}3.44{\tiny\textcolor{customPurple}{$\pm$3.94}} \\
    \textcolor{blue}{$\vardiamondsuit$} ProSFDA (2026, PR)~\cite{hu2025source} & 0.5883{\tiny\textcolor{customPurple}{$\pm$0.1801}} & 0.5796{\tiny\textcolor{customPurple}{$\pm$0.2058}} & 0.7412{\tiny\textcolor{customPurple}{$\pm$0.0583}} & 6.78{\tiny\textcolor{customPurple}{$\pm$5.13}} & 7.05{\tiny\textcolor{customPurple}{$\pm$5.53}} & 5.35{\tiny\textcolor{customPurple}{$\pm$5.83}} \\
    \hline
  \end{tabular}%
  }
\end{table}

\begin{table}[h!]
  \centering
  \fontsize{7.5}{8}\selectfont 
  \setlength{\tabcolsep}{1.5mm}
  \renewcommand{\arraystretch}{1.3}
  \caption{\textbf{Comprehensive performance comparison of TTA methods on MRI brain tumor segmentation: Sen and PPV across ET/TC/WT.}
 Results colored in \colorbox{red!30}{\makebox[1.3em][c]{\raisebox{0pt}[0.5em][0em]{red}}}, \colorbox{orange!30}{\makebox[2.8em][c]{\raisebox{0pt}[0.5em][0em]{orange}}}, and \colorbox{yellow!30}{\makebox[2.5em][c]{\raisebox{0pt}[0.5em][0em]{yellow}}} denote the best, second-best, and third-best performances, respectively. Rows with a \colorbox{gray!30}{\makebox[1.6em][c]{\raisebox{0pt}[0.5em][0em]{gray}}} background indicate Dice values lower than that of the Target-Domain(w/o TTA). Symbols denote paradigms: \textcolor{black}{\(\spadesuit\)} for Input-level Translation, \textcolor{green}{\(\clubsuit\)} for Feature-level Alignment, \textcolor{red}{\(\varheartsuit\)} for Output-level Regularization, \textcolor{blue}{\(\vardiamondsuit\)} for Prior Estimation. All improvements are statistically significant at \( p < 0.05 \).}
  \label{tab:mri-sen-ppv}
  \resizebox{\textwidth}{!}{%
  \begin{tabular}{lcccccc}
   \hline
    \rowcolor{customHeader}
    \multicolumn{1}{l}{\cellcolor{customHeader}} & \multicolumn{3}{c}{\cellcolor{customHeader}\textbf{Sen$\uparrow$}} & \multicolumn{3}{c}{\cellcolor{customHeader}\textbf{PPV$\uparrow$}} \\
    \cline{2-7}
    \rowcolor{customHeader}
    \multicolumn{1}{l}{\multirow{-2}{*}{\cellcolor{customHeader}\textbf{Method}}} & \textbf{ET} & \textbf{TC} & \textbf{WT} & \textbf{ET} & \textbf{TC} & \textbf{WT} \\
   \hline
   \textbf{Intra-Domain}
      & \textbf{0.6862{\tiny\textcolor{customPurple}{$\pm$0.0181}}}
      & \textbf{0.6978{\tiny\textcolor{customPurple}{$\pm$0.0209}}}
      & \textbf{0.7094{\tiny\textcolor{customPurple}{$\pm$0.0083}}}
      & \textbf{0.7912{\tiny\textcolor{customPurple}{$\pm$0.1511}}}
      & \textbf{0.7758{\tiny\textcolor{customPurple}{$\pm$0.1677}}}
      & \textbf{0.8518{\tiny\textcolor{customPurple}{$\pm$0.0559}}} \\
   \hline
    \textbf{Target-Domain (w/o TTA)}
      & \textbf{0.6328{\tiny\textcolor{customPurple}{$\pm$0.0398}}} & \textbf{0.6302{\tiny\textcolor{customPurple}{$\pm$0.0435}}} & \textbf{0.6771{\tiny\textcolor{customPurple}{$\pm$0.0387}}}
      & \textbf{0.7472{\tiny\textcolor{customPurple}{$\pm$0.0826}}} & \textbf{0.7802{\tiny\textcolor{customPurple}{$\pm$0.0722}}} & \textbf{0.7881{\tiny\textcolor{customPurple}{$\pm$0.0892}}} \\
    \hline
    \textcolor{black}{$\spadesuit$} RSA (2021, MICCAI)~\cite{zeng2024reliable} & 0.6455{\tiny\textcolor{customPurple}{$\pm$0.0231}} & 0.6559{\tiny\textcolor{customPurple}{$\pm$0.1071}} & 0.6828{\tiny\textcolor{customPurple}{$\pm$0.0195}} & 0.6162{\tiny\textcolor{customPurple}{$\pm$0.1242}} & 0.7021{\tiny\textcolor{customPurple}{$\pm$0.0143}} & 0.0993{\tiny\textcolor{customPurple}{$\pm$0.0993}} \\
    \textcolor{black}{$\spadesuit$} DL-TTA (2022, TMI)~\cite{yang2022dltta} & 0.6593{\tiny\textcolor{customPurple}{$\pm$0.1631}} & 0.6475{\tiny\textcolor{customPurple}{$\pm$0.0518}} & 0.6868{\tiny\textcolor{customPurple}{$\pm$0.0488}} & 0.7123{\tiny\textcolor{customPurple}{$\pm$0.1351}} & 0.7561{\tiny\textcolor{customPurple}{$\pm$0.1362}} & 0.7282{\tiny\textcolor{customPurple}{$\pm$0.1551}} \\
    \textcolor{black}{$\spadesuit$} SFDA-FSM (2022, MIA)~\cite{yang2022source} & \cellcolor{second}0.6671{\tiny\textcolor{customPurple}{$\pm$0.0523}} & 0.6553{\tiny\textcolor{customPurple}{$\pm$0.0672}} & 0.6971{\tiny\textcolor{customPurple}{$\pm$0.0552}} & 0.6312{\tiny\textcolor{customPurple}{$\pm$0.2734}} & 0.6764{\tiny\textcolor{customPurple}{$\pm$0.0287}} & 0.6518{\tiny\textcolor{customPurple}{$\pm$0.2349}} \\
    \textcolor{black}{$\spadesuit$} STDR (2024, TMI)~\cite{wang2024dual} & 0.6312{\tiny\textcolor{customPurple}{$\pm$0.0371}} & 0.6355{\tiny\textcolor{customPurple}{$\pm$0.0427}} & 0.6606{\tiny\textcolor{customPurple}{$\pm$0.0860}} & 0.5784{\tiny\textcolor{customPurple}{$\pm$0.1749}} & \cellcolor{third}0.8053{\tiny\textcolor{customPurple}{$\pm$0.0427}} & \cellcolor{second}0.8401{\tiny\textcolor{customPurple}{$\pm$0.1293}} \\
    \textcolor{black}{$\spadesuit$} AIF-SFDA (2025, AAAI)~\cite{li2025aif} & 0.6527{\tiny\textcolor{customPurple}{$\pm$0.0566}} & 0.6409{\tiny\textcolor{customPurple}{$\pm$0.1297}} & 0.6802{\tiny\textcolor{customPurple}{$\pm$0.1425}} & 0.7052{\tiny\textcolor{customPurple}{$\pm$0.1268}} & 0.7358{\tiny\textcolor{customPurple}{$\pm$0.1737}} & 0.7216{\tiny\textcolor{customPurple}{$\pm$0.1486}} \\
    \hline
    \textcolor{green}{$\clubsuit$} DANN (2016, JMLR)~\cite{ganin2016domain} & 0.6421{\tiny\textcolor{customPurple}{$\pm$0.0531}} & 0.6592{\tiny\textcolor{customPurple}{$\pm$0.0429}} & 0.6828{\tiny\textcolor{customPurple}{$\pm$0.0485}} & 0.5332{\tiny\textcolor{customPurple}{$\pm$0.0838}} & 0.6585{\tiny\textcolor{customPurple}{$\pm$0.0706}} & 0.0891{\tiny\textcolor{customPurple}{$\pm$0.0891}} \\
    \textcolor{green}{$\clubsuit$} DeTTA (2024, WACV)~\cite{wen2024denoising} & 0.6236{\tiny\textcolor{customPurple}{$\pm$0.0298}} & 0.6032{\tiny\textcolor{customPurple}{$\pm$0.0349}} & 0.6712{\tiny\textcolor{customPurple}{$\pm$0.0324}} & 0.6026{\tiny\textcolor{customPurple}{$\pm$0.0953}} & 0.7957{\tiny\textcolor{customPurple}{$\pm$0.0928}} & 0.7941{\tiny\textcolor{customPurple}{$\pm$0.1444}} \\
    \textcolor{green}{$\clubsuit$} TestFit (2024, MIA)~\cite{zhang2024testfit} & 0.6312{\tiny\textcolor{customPurple}{$\pm$0.0759}} & 0.5989{\tiny\textcolor{customPurple}{$\pm$0.0215}} & 0.6783{\tiny\textcolor{customPurple}{$\pm$0.0204}} & \cellcolor{best}0.7590{\tiny\textcolor{customPurple}{$\pm$0.0835}} & \cellcolor{best}0.8323{\tiny\textcolor{customPurple}{$\pm$0.0829}} & \cellcolor{best}0.8609{\tiny\textcolor{customPurple}{$\pm$0.0923}} \\
    \textcolor{green}{$\clubsuit$} UDA-MIMA (2024, CMPB)~\cite{hu2024unsupervised} & 0.6421{\tiny\textcolor{customPurple}{$\pm$0.0514}} & 0.6544{\tiny\textcolor{customPurple}{$\pm$0.0649}} & 0.6871{\tiny\textcolor{customPurple}{$\pm$0.0511}} & 0.7005{\tiny\textcolor{customPurple}{$\pm$0.0514}} & 0.7402{\tiny\textcolor{customPurple}{$\pm$0.1244}} & 0.7129{\tiny\textcolor{customPurple}{$\pm$0.0740}} \\
    \textcolor{green}{$\clubsuit$} GraTa (2025, AAAI)~\cite{chen2025gradient} & 0.6378{\tiny\textcolor{customPurple}{$\pm$0.0366}} & \cellcolor{second}0.6618{\tiny\textcolor{customPurple}{$\pm$0.0433}} & \cellcolor{third}0.7466{\tiny\textcolor{customPurple}{$\pm$0.0397}} & 0.6842{\tiny\textcolor{customPurple}{$\pm$0.1242}} & 0.7593{\tiny\textcolor{customPurple}{$\pm$0.1128}} & 0.7466{\tiny\textcolor{customPurple}{$\pm$0.1692}} \\
    \hline
    \textcolor{red}{$\varheartsuit$} TENT (2021, ICLR)~\cite{wang2021tent} & 0.6386{\tiny\textcolor{customPurple}{$\pm$0.0731}} & 0.6418{\tiny\textcolor{customPurple}{$\pm$0.0724}} & 0.6838{\tiny\textcolor{customPurple}{$\pm$0.0718}} & 0.7055{\tiny\textcolor{customPurple}{$\pm$0.0973}} & 0.7889{\tiny\textcolor{customPurple}{$\pm$0.0633}} & 0.7725{\tiny\textcolor{customPurple}{$\pm$0.1221}} \\
    \textcolor{red}{$\varheartsuit$} UPL-SFDA (2023, TMI)~\cite{wu2023upl} & \cellcolor{third}0.6612{\tiny\textcolor{customPurple}{$\pm$0.0487}} & \cellcolor{best}0.6691{\tiny\textcolor{customPurple}{$\pm$0.0499}} & 0.6828{\tiny\textcolor{customPurple}{$\pm$0.1172}} & 0.6495{\tiny\textcolor{customPurple}{$\pm$0.0831}} & 0.6691{\tiny\textcolor{customPurple}{$\pm$0.1557}} & 0.7417{\tiny\textcolor{customPurple}{$\pm$0.0489}} \\
    \textcolor{red}{$\varheartsuit$} DG-TTA (2023, arXiv)~\cite{weihsbach2023dg} & 0.6375{\tiny\textcolor{customPurple}{$\pm$0.0365}} & \cellcolor{third}0.6613{\tiny\textcolor{customPurple}{$\pm$0.0431}} & 0.7463{\tiny\textcolor{customPurple}{$\pm$0.0393}} & 0.6698{\tiny\textcolor{customPurple}{$\pm$0.1140}} & 0.7613{\tiny\textcolor{customPurple}{$\pm$0.1123}} & 0.7463{\tiny\textcolor{customPurple}{$\pm$0.1642}} \\
    \textcolor{red}{$\varheartsuit$} SaTTCA (2023, MICCAI)~\cite{li2023scale} & 0.6397{\tiny\textcolor{customPurple}{$\pm$0.0366}} & 0.6212{\tiny\textcolor{customPurple}{$\pm$0.0367}} & 0.6828{\tiny\textcolor{customPurple}{$\pm$0.0391}} & 0.6656{\tiny\textcolor{customPurple}{$\pm$0.0579}} & 0.7395{\tiny\textcolor{customPurple}{$\pm$0.1369}} & 0.7154{\tiny\textcolor{customPurple}{$\pm$0.1742}} \\
    \textcolor{red}{$\varheartsuit$} SmaRT (2025, arXiv)~\cite{Wang2025SmaRTSR} & 0.6374{\tiny\textcolor{customPurple}{$\pm$0.2147}} & 0.6256{\tiny\textcolor{customPurple}{$\pm$0.0704}} & 0.6719{\tiny\textcolor{customPurple}{$\pm$0.0450}} & 0.6494{\tiny\textcolor{customPurple}{$\pm$0.2794}} & 0.6929{\tiny\textcolor{customPurple}{$\pm$0.2245}} & 0.7019{\tiny\textcolor{customPurple}{$\pm$0.0941}} \\
    \hline
    \textcolor{blue}{$\vardiamondsuit$} AdaMI (2022, MICCAI)~\cite{bateson2022test} & 0.6249{\tiny\textcolor{customPurple}{$\pm$0.0375}} & 0.6033{\tiny\textcolor{customPurple}{$\pm$0.0415}} & \cellcolor{best}0.8301{\tiny\textcolor{customPurple}{$\pm$0.0842}} & \cellcolor{second}0.7474{\tiny\textcolor{customPurple}{$\pm$0.1068}} & \cellcolor{second}0.8108{\tiny\textcolor{customPurple}{$\pm$0.1124}} & \cellcolor{third}0.8301{\tiny\textcolor{customPurple}{$\pm$0.1553}} \\
    \textcolor{blue}{$\vardiamondsuit$} PASS (2024, TMI)~\cite{zhang2024pass} & 0.6320{\tiny\textcolor{customPurple}{$\pm$0.0835}} & 0.6107{\tiny\textcolor{customPurple}{$\pm$0.0813}} & 0.6766{\tiny\textcolor{customPurple}{$\pm$0.0204}} & \cellcolor{third}0.7187{\tiny\textcolor{customPurple}{$\pm$0.0970}} & 0.7058{\tiny\textcolor{customPurple}{$\pm$0.0955}} & 0.7991{\tiny\textcolor{customPurple}{$\pm$0.0922}} \\
    \textcolor{blue}{$\vardiamondsuit$} VPTTA (2024, CVPR)~\cite{chen2024each} & 0.6258{\tiny\textcolor{customPurple}{$\pm$0.0291}} & 0.6019{\tiny\textcolor{customPurple}{$\pm$0.0342}} & 0.6697{\tiny\textcolor{customPurple}{$\pm$0.0318}} & 0.6246{\tiny\textcolor{customPurple}{$\pm$0.0946}} & 0.7694{\tiny\textcolor{customPurple}{$\pm$0.0921}} & 0.7928{\tiny\textcolor{customPurple}{$\pm$0.1437}} \\
    \textcolor{blue}{$\vardiamondsuit$} ExploringTTA (2025, ISBI)~\cite{omolegan2025exploring} & \cellcolor{best}0.7322{\tiny\textcolor{customPurple}{$\pm$0.1016}} & 0.6082{\tiny\textcolor{customPurple}{$\pm$0.0412}} & \cellcolor{second}0.8008{\tiny\textcolor{customPurple}{$\pm$0.0837}} & 0.7322{\tiny\textcolor{customPurple}{$\pm$0.1472}} & 0.7554{\tiny\textcolor{customPurple}{$\pm$0.1627}} & 0.8008{\tiny\textcolor{customPurple}{$\pm$0.1507}} \\
    \textcolor{blue}{$\vardiamondsuit$} ProSFDA (2026, PR)~\cite{hu2025source} & 0.6318{\tiny\textcolor{customPurple}{$\pm$0.0423}} & 0.6197{\tiny\textcolor{customPurple}{$\pm$0.0621}} & 0.6846{\tiny\textcolor{customPurple}{$\pm$0.0240}} & 0.7074{\tiny\textcolor{customPurple}{$\pm$0.2076}} & 0.6971{\tiny\textcolor{customPurple}{$\pm$0.0553}} & 0.7675{\tiny\textcolor{customPurple}{$\pm$0.1945}} \\
    \hline
  \end{tabular}%
  }
\end{table}


\subsubsection{Computed Tomography Modality}

\begin{figure}[h!]
  \centering
  \includegraphics[width=\linewidth]{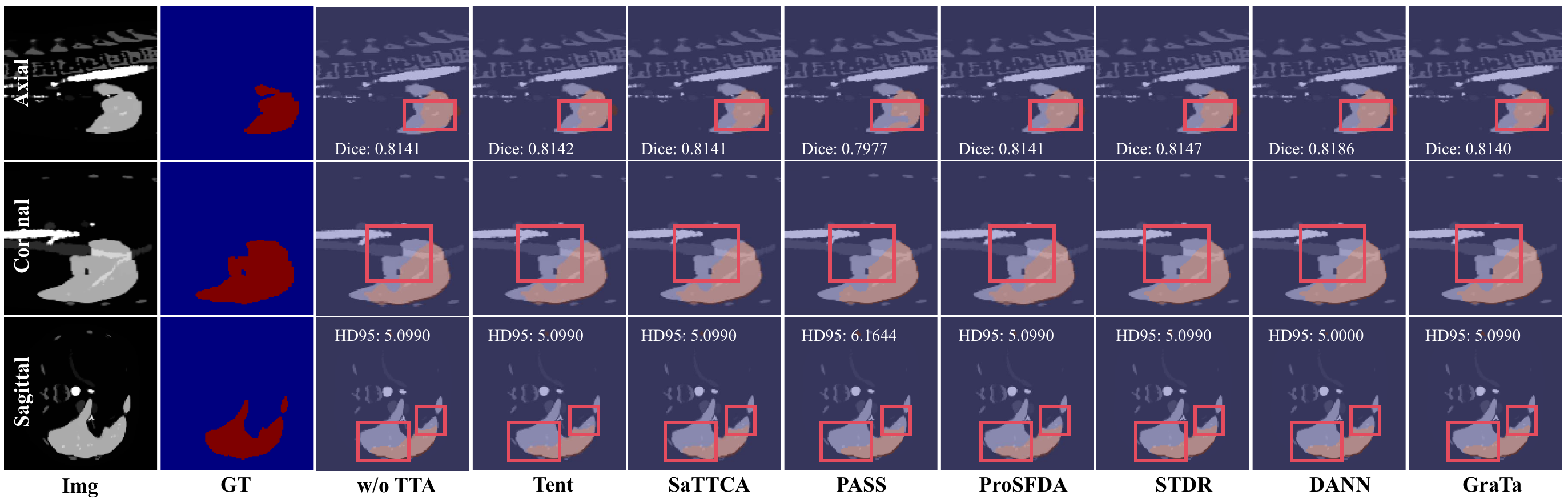}
  \caption{\textbf{Visual comparison of TTA efficacy across different methods on CT datasets.} The three rows represent slices taken from three different orientations. GT denotes the ground-truth.}
  \label{fig: CT_seg}
\end{figure}

In the Computed Tomography task reported in Table~\ref{tab:ct}, all twenty methods increased the Dice score, and fifteen of them also reduced HD95, whereas four led to higher HD95. Taken together, the CT results show consistent gains across paradigms. DG-TTA achieves the highest Dice score, reaching 0.8500 and surpassing the baseline value of 0.6595. SFDA\text{-}FSM delivers the most considerable reduction in HD95, lowering the value by 6.8811, which corresponds to an improvement of approximately 50\%. These trends align with known properties of CT reconstruction. Variability in reconstruction kernel and slice thickness induces intensity drift and artifacts, which in turn make HD95 highly sensitive to boundary outliers. Methods that adapt without explicit shape constraints, for example GraTa and TENT, are therefore more prone to boundary shifts or over\text{-}suppression during adaptation, which elevates HD95. Fig~\ref{fig: CT_seg} illustrates the performance of different TTA methods on the CT task, with most models achieving comparable results.

\begin{table}[h!]
  \centering
  \caption{\textbf{Comprehensive performance comparison of TTA methods on Liver Segmentation in Computed Tomography.}
 Results colored in \colorbox{red!30}{\makebox[1.3em][c]{\raisebox{0pt}[0.5em][0em]{red}}}, \colorbox{orange!30}{\makebox[2.8em][c]{\raisebox{0pt}[0.5em][0em]{orange}}}, and \colorbox{yellow!30}{\makebox[2.5em][c]{\raisebox{0pt}[0.5em][0em]{yellow}}} denote the best, second-best, and third-best performances, respectively. Rows with a \colorbox{gray!30}{\makebox[1.6em][c]{\raisebox{0pt}[0.5em][0em]{gray}}} background indicate Dice values lower than that of the Target-Domain(w/o TTA). Symbols denote paradigms: \textcolor{black}{\(\spadesuit\)} for Input-level Translation, \textcolor{green}{\(\clubsuit\)} for Feature-level Alignment, \textcolor{red}{\(\varheartsuit\)} for Output-level Regularization, \textcolor{blue}{\(\vardiamondsuit\)} for Prior Estimation. All improvements are statistically significant at \( p < 0.05 \).}
  \vspace{0.5em}
  \label{tab:ct}
  \resizebox{\textwidth}{!}{%
  \begin{tabular}{lccccc}
    \arrayrulecolor{black}\hline
    \rowcolor[HTML]{D8D6C0}
    \textbf{Method} & \textbf{Dice $\uparrow$} & \textbf{HD95 $\downarrow$} & \textbf{JI $\uparrow$} & \textbf{Sen $\uparrow$} & \textbf{PPV $\uparrow$} \arraybackslash \\
    \arrayrulecolor{black}\hline
    \textbf{Intra-Domain} & \textbf{0.8949\textcolor{customPurple}{\scriptsize$\pm$0.0250}} & \textbf{1.63\textcolor{customPurple}{\scriptsize$\pm$1.68}} & \textbf{0.8116\textcolor{customPurple}{\scriptsize$\pm$0.0388}} & \textbf{0.9569\textcolor{customPurple}{\scriptsize$\pm$0.0172}} & \textbf{0.8433\textcolor{customPurple}{\scriptsize$\pm$0.0382}} \\
    \arrayrulecolor{black}\hline
    \textbf{Target-Domain (w/o TTA)} & \textbf{0.6595\textcolor{customPurple}{\scriptsize$\pm$0.2833}} & \textbf{13.91\textcolor{customPurple}{\scriptsize$\pm$11.30}} & \textbf{0.5489\textcolor{customPurple}{\scriptsize$\pm$0.2723}} & \textbf{0.6376\textcolor{customPurple}{\scriptsize$\pm$0.0693}} & \textbf{0.8814\textcolor{customPurple}{\scriptsize$\pm$0.2263}} \\
    \arrayrulecolor{black}\hline
    \textcolor{black}{$\spadesuit$} RSA (2021, MICCAI)~\cite{zeng2024reliable}          & 0.6973\textcolor{customPurple}{\scriptsize$\pm$0.2583} & 12.39\textcolor{customPurple}{\scriptsize$\pm$10.28} & 0.5857\textcolor{customPurple}{\scriptsize$\pm$0.2579} & 0.6461\textcolor{customPurple}{\scriptsize$\pm$0.0655} & 0.8743\textcolor{customPurple}{\scriptsize$\pm$0.2211} \\
    \textcolor{black}{$\spadesuit$} DL-TTA (2022, TMI)~\cite{yang2022dltta}        & 0.6752\textcolor{customPurple}{\scriptsize$\pm$0.2183} & 11.58\textcolor{customPurple}{\scriptsize$\pm$9.51} & 0.6031\textcolor{customPurple}{\scriptsize$\pm$0.1947} & 0.6673\textcolor{customPurple}{\scriptsize$\pm$0.0521} & \cellcolor{second}0.9002\textcolor{customPurple}{\scriptsize$\pm$0.1248} \\
    \textcolor{black}{$\spadesuit$} SFDA-FSM (2022, MIA)~\cite{yang2022source}     & \cellcolor{third}0.7780\textcolor{customPurple}{\scriptsize$\pm$0.1147} & \cellcolor{best}6.88\textcolor{customPurple}{\scriptsize$\pm$7.17} & \cellcolor{third}0.6845\textcolor{customPurple}{\scriptsize$\pm$0.1066} & 0.6764\textcolor{customPurple}{\scriptsize$\pm$0.0271} & 0.8711\textcolor{customPurple}{\scriptsize$\pm$0.0894} \\
    \textcolor{black}{$\spadesuit$} STDR (2024, TMI)~\cite{wang2024dual}         & 0.7255\textcolor{customPurple}{\scriptsize$\pm$0.2403} & 11.34\textcolor{customPurple}{\scriptsize$\pm$9.82} & 0.6147\textcolor{customPurple}{\scriptsize$\pm$0.2458} & 0.6543\textcolor{customPurple}{\scriptsize$\pm$0.0626} & 0.8771\textcolor{customPurple}{\scriptsize$\pm$0.2176} \\
    \textcolor{black}{$\spadesuit$} AIF-SFDA (2025, AAAI)~\cite{li2025aif}     & 0.7018\textcolor{customPurple}{\scriptsize$\pm$0.2625} & \cellcolor{second}9.39\textcolor{customPurple}{\scriptsize$\pm$10.28} & 0.5900\textcolor{customPurple}{\scriptsize$\pm$0.2620} & 0.6503\textcolor{customPurple}{\scriptsize$\pm$0.0696} & 0.8786\textcolor{customPurple}{\scriptsize$\pm$0.2252} \\
    \arrayrulecolor{black}\hline
    \textcolor{green}{$\clubsuit$} DANN (2016, JMLR)~\cite{ganin2016domain}         & 0.7082\textcolor{customPurple}{\scriptsize$\pm$0.2205} & 10.99\textcolor{customPurple}{\scriptsize$\pm$8.83} & 0.5866\textcolor{customPurple}{\scriptsize$\pm$0.2285} & 0.6499\textcolor{customPurple}{\scriptsize$\pm$0.0593} & 0.8415\textcolor{customPurple}{\scriptsize$\pm$0.1911} \\
        \textcolor{green}{$\clubsuit$} DeTTA (2024, WACV)~\cite{wen2024denoising}        & 0.6838\textcolor{customPurple}{\scriptsize$\pm$0.1139} & 11.51\textcolor{customPurple}{\scriptsize$\pm$9.81} & 0.5749\textcolor{customPurple}{\scriptsize$\pm$0.0958} & \cellcolor{second}0.6570\textcolor{customPurple}{\scriptsize$\pm$0.0257} & 0.8182\textcolor{customPurple}{\scriptsize$\pm$0.1307} \\
            \textcolor{green}{$\clubsuit$} TestFit (2024, MIA)~\cite{zhang2024testfit}       & 0.6614\textcolor{customPurple}{\scriptsize$\pm$0.2155} & \cellcolor{third}9.95\textcolor{customPurple}{\scriptsize$\pm$8.27} & 0.5816\textcolor{customPurple}{\scriptsize$\pm$0.1835} & 0.6440\textcolor{customPurple}{\scriptsize$\pm$0.0492} & \cellcolor{best}0.9005\textcolor{customPurple}{\scriptsize$\pm$0.1086} \\
    \textcolor{green}{$\clubsuit$} UDA-MIMA (2024, CMPB)~\cite{hu2024unsupervised}     & 0.6919\textcolor{customPurple}{\scriptsize$\pm$0.2354} & 12.21\textcolor{customPurple}{\scriptsize$\pm$10.97} & 0.5756\textcolor{customPurple}{\scriptsize$\pm$0.2036} & 0.6423\textcolor{customPurple}{\scriptsize$\pm$0.0542} & \cellcolor{third}0.8842\textcolor{customPurple}{\scriptsize$\pm$0.2352} \\
    \textcolor{green}{$\clubsuit$} GraTa (2025, AAAI)~\cite{chen2025gradient}        & 0.6916\textcolor{customPurple}{\scriptsize$\pm$0.1485} & 23.12\textcolor{customPurple}{\scriptsize$\pm$15.27} & 0.5472\textcolor{customPurple}{\scriptsize$\pm$0.1652} & \cellcolor{third}0.6519\textcolor{customPurple}{\scriptsize$\pm$0.0334} & 0.7493\textcolor{customPurple}{\scriptsize$\pm$0.1877} \\
    \arrayrulecolor{black}\hline
    \textcolor{red}{$\varheartsuit$} TENT (2021, ICLR)~\cite{wang2021tent}         & 0.6918\textcolor{customPurple}{\scriptsize$\pm$0.1484} & 23.10\textcolor{customPurple}{\scriptsize$\pm$15.25} & 0.5474\textcolor{customPurple}{\scriptsize$\pm$0.1652} & \cellcolor{third}0.6519\textcolor{customPurple}{\scriptsize$\pm$0.0334} & 0.7495\textcolor{customPurple}{\scriptsize$\pm$0.1876} \\
    \textcolor{red}{$\varheartsuit$} UPL-SFDA (2023, TMI)~\cite{wu2023upl}      & 0.6748\textcolor{customPurple}{\scriptsize$\pm$0.1145} & 12.73\textcolor{customPurple}{\scriptsize$\pm$10.26} & 0.5657\textcolor{customPurple}{\scriptsize$\pm$0.1964} & 0.6471\textcolor{customPurple}{\scriptsize$\pm$0.0526} & 0.8023\textcolor{customPurple}{\scriptsize$\pm$0.1632} \\
    \textcolor{red}{$\varheartsuit$} DG-TTA (2023, arXiv)~\cite{weihsbach2023dg}        & \cellcolor{best}0.8500\textcolor{customPurple}{\scriptsize$\pm$0.1037} & 8.61\textcolor{customPurple}{\scriptsize$\pm$11.89} & \cellcolor{best}0.7518\textcolor{customPurple}{\scriptsize$\pm$0.1408} & 0.7112\textcolor{customPurple}{\scriptsize$\pm$0.0210} & 0.8216\textcolor{customPurple}{\scriptsize$\pm$0.1615} \\
    \textcolor{red}{$\varheartsuit$} SaTTCA (2023, MICCAI)~\cite{li2023scale}       & 0.6793\textcolor{customPurple}{\scriptsize$\pm$0.1098} & 12.51\textcolor{customPurple}{\scriptsize$\pm$9.81} & 0.5706\textcolor{customPurple}{\scriptsize$\pm$0.0919} & 0.6528\textcolor{customPurple}{\scriptsize$\pm$0.0218} & 0.8139\textcolor{customPurple}{\scriptsize$\pm$0.1268} \\
    \textcolor{red}{$\varheartsuit$} SmaRT (2025, arXiv)~\cite{Wang2025SmaRTSR}        & \cellcolor{second}0.8405\textcolor{customPurple}{\scriptsize$\pm$0.1231} & 8.91\textcolor{customPurple}{\scriptsize$\pm$12.26} & \cellcolor{second}0.7422\textcolor{customPurple}{\scriptsize$\pm$0.1634} & \cellcolor{best}0.7261\textcolor{customPurple}{\scriptsize$\pm$0.0061} & 0.7534\textcolor{customPurple}{\scriptsize$\pm$0.1649} \\
    \arrayrulecolor{black}\hline
    \textcolor{blue}{$\vardiamondsuit$} AdaMI (2022, MICCAI)~\cite{bateson2022test}        & 0.7094\textcolor{customPurple}{\scriptsize$\pm$0.2200} & 10.97\textcolor{customPurple}{\scriptsize$\pm$8.82} & 0.5879\textcolor{customPurple}{\scriptsize$\pm$0.2282} & 0.6503\textcolor{customPurple}{\scriptsize$\pm$0.0592} & 0.8407\textcolor{customPurple}{\scriptsize$\pm$0.1834} \\
        \textcolor{blue}{$\vardiamondsuit$} PASS (2024, TMI)~\cite{zhang2024pass}         & 0.7088\textcolor{customPurple}{\scriptsize$\pm$0.2204} & 11.52\textcolor{customPurple}{\scriptsize$\pm$8.80} & 0.5873\textcolor{customPurple}{\scriptsize$\pm$0.2284} & 0.6521\textcolor{customPurple}{\scriptsize$\pm$0.0397} & 0.8420\textcolor{customPurple}{\scriptsize$\pm$0.1913} \\
    \textcolor{blue}{$\vardiamondsuit$} VPTTA (2024, CVPR)~\cite{chen2024each}       & 0.6792\textcolor{customPurple}{\scriptsize$\pm$0.1187} & 14.73\textcolor{customPurple}{\scriptsize$\pm$10.26} & 0.5700\textcolor{customPurple}{\scriptsize$\pm$0.2005} & 0.6514\textcolor{customPurple}{\scriptsize$\pm$0.0567} & 0.8066\textcolor{customPurple}{\scriptsize$\pm$0.1674} \\

    \textcolor{blue}{$\vardiamondsuit$} ExploringTTA (2025, ISBI)~\cite{omolegan2025exploring}& 0.6933\textcolor{customPurple}{\scriptsize$\pm$0.1472} & 20.86\textcolor{customPurple}{\scriptsize$\pm$14.80} & 0.5489\textcolor{customPurple}{\scriptsize$\pm$0.1642} & \cellcolor{third}0.6519\textcolor{customPurple}{\scriptsize$\pm$0.0335} & 0.7525\textcolor{customPurple}{\scriptsize$\pm$0.1841} \\
    \textcolor{blue}{$\vardiamondsuit$} ProSFDA (2026, PR)~\cite{hu2025source}      & 0.6937\textcolor{customPurple}{\scriptsize$\pm$0.2003} & 11.95\textcolor{customPurple}{\scriptsize$\pm$8.54} & 0.5624\textcolor{customPurple}{\scriptsize$\pm$0.2080} & 0.6458\textcolor{customPurple}{\scriptsize$\pm$0.0542} & 0.8135\textcolor{customPurple}{\scriptsize$\pm$0.1738} \\
    \arrayrulecolor{black}\hline
  \end{tabular}%
  }
\end{table}

\subsubsection{Ultrasound Modality}

\begin{figure}[h!]
  \centering
  \includegraphics[width=\linewidth]{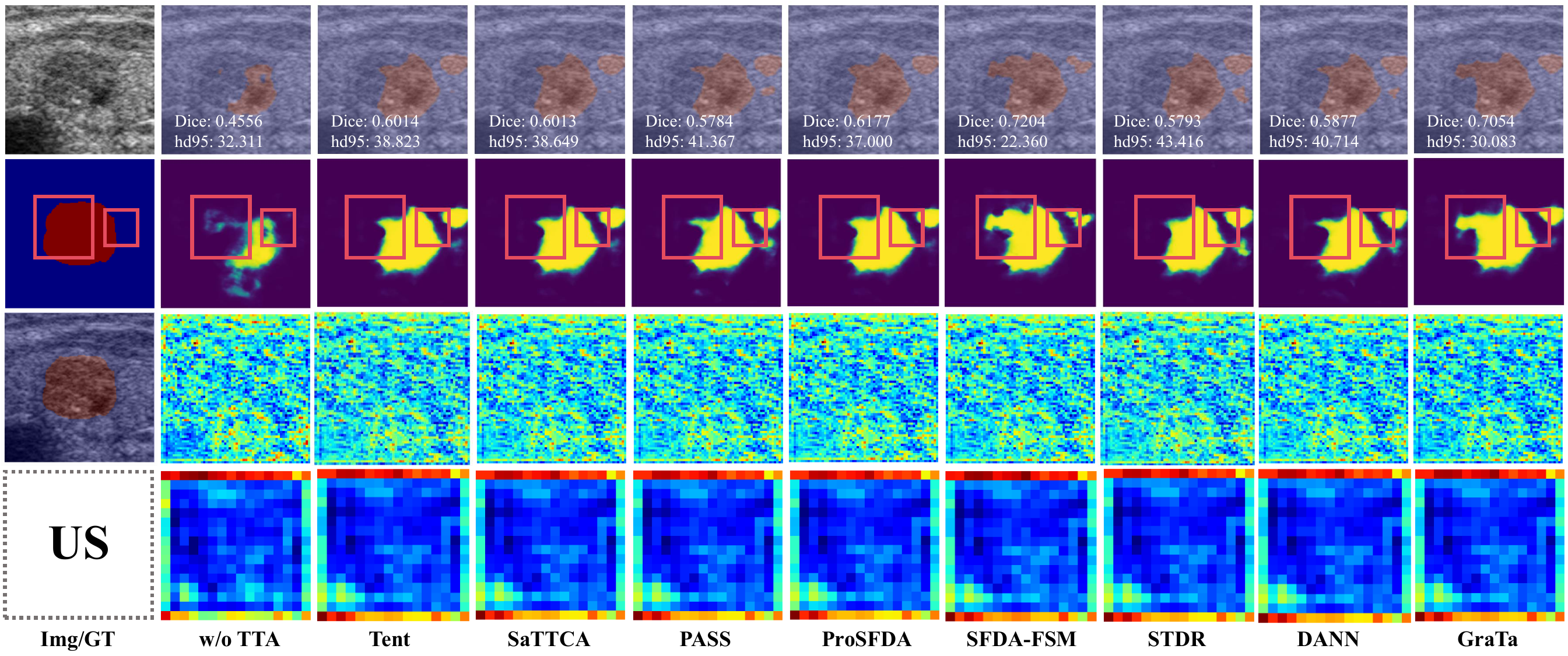}
  \caption{\textbf{Visual comparison of TTA efficacy across different methods on US datasets.} Rows 1 and 2 illustrate the segmentation outcomes. Rows 3 and 4 depict changes in features extracted by the Encoder after TTA. GT denotes the ground-truth.}
  \label{fig: US_seg}
\end{figure}

In the Ultrasound task shown in Table~\ref{tab:us}, most methods achieved improvements. SFDA-FSM achieved the most significant Dice gain, with a score of 0.8333 compared to the baseline, representing an increase of 0.3522, or 73 percent. However, its HD95 worsened to 10.6708, an 83 percent increase. Methods such as RSA also exhibited marked HD95 degradation, closely related to intrinsic ultrasound characteristics, including speckle noise, low inter-tissue contrast, acoustic shadowing, and occlusions, which cause boundary ambiguity and discontinuity. When Input-level Transformation applies strong style remapping during inference, it can induce local edge truncation or expansion, amplifying HD95 sensitivity. Fig.~\ref{fig: US_seg} illustrates the effects of various TTA methods on the US task and their influence on the encoder, where most models exhibit beneficial effects. 

In summary, experimental results indicate that input-level transformation methods offer plug-and-play adaptability and strong cross-modality and cross-center generalization, whereas feature-level and output-level strategies primarily enhance boundary consistency and 3D adaptation. Each paradigm exhibits distinct advantages, suggesting that future work should focus on their synergistic integration to achieve more robust and widely generalizable TTA.

\begin{table}[h!]
  \centering
  \caption{\textbf{Comprehensive performance comparison of TTA methods on Ultrasound Thyroid Nodule Segmentation.}
 Results colored in \colorbox{red!30}{\makebox[1.3em][c]{\raisebox{0pt}[0.5em][0em]{red}}}, \colorbox{orange!30}{\makebox[2.8em][c]{\raisebox{0pt}[0.5em][0em]{orange}}}, and \colorbox{yellow!30}{\makebox[2.5em][c]{\raisebox{0pt}[0.5em][0em]{yellow}}} denote the best, second-best, and third-best performances, respectively. Rows with a \colorbox{gray!30}{\makebox[1.6em][c]{\raisebox{0pt}[0.5em][0em]{gray}}} background indicate Dice values lower than that of the Target-Domain(w/o TTA). Symbols denote paradigms: \textcolor{black}{\(\spadesuit\)} for Input-level Translation, \textcolor{green}{\(\clubsuit\)} for Feature-level Alignment, \textcolor{red}{\(\varheartsuit\)} for Output-level Regularization, \textcolor{blue}{\(\vardiamondsuit\)} for Prior Estimation. All improvements are statistically significant at \( p < 0.05 \).}
  \label{tab:us}
  \resizebox{\textwidth}{!}{%
  \begin{tabular}{lccccc}
    \arrayrulecolor{black}\hline
    \rowcolor[HTML]{D8D6C0}
    \textbf{Method} & \textbf{Dice $\uparrow$} & \textbf{HD95 $\downarrow$} & \textbf{JI $\uparrow$} & \textbf{Sen $\uparrow$} & \textbf{PPV $\uparrow$} \arraybackslash \\
    \arrayrulecolor{black}\hline
    \textbf{Intra-Domain} & \textbf{0.7215\textcolor{customPurple}{\scriptsize$\pm$0.0266}} & \textbf{31.01\textcolor{customPurple}{\scriptsize$\pm$8.30}} & \textbf{0.6066\textcolor{customPurple}{\scriptsize$\pm$0.0313}} & \textbf{0.7812\textcolor{customPurple}{\scriptsize$\pm$0.0351}} & \textbf{0.7541\textcolor{customPurple}{\scriptsize$\pm$0.0328}} \\
    \arrayrulecolor{black}\hline
    \textbf{Target-Domain (w/o TTA)} & \textbf{0.4811\textcolor{customPurple}{\scriptsize$\pm$0.2708}} & \textbf{62.78\textcolor{customPurple}{\scriptsize$\pm$68.88}} & \textbf{0.3591\textcolor{customPurple}{\scriptsize$\pm$0.2393}} & \textbf{0.4288\textcolor{customPurple}{\scriptsize$\pm$0.2958}} & \textbf{0.7531\textcolor{customPurple}{\scriptsize$\pm$0.2801}} \\
    \arrayrulecolor{black}\hline
    \textcolor{black}{$\spadesuit$} RSA (2021, MICCAI)~\cite{zeng2024reliable}          & 0.4879\textcolor{customPurple}{\scriptsize$\pm$0.3011} & 71.24\textcolor{customPurple}{\scriptsize$\pm$92.52} & 0.3752\textcolor{customPurple}{\scriptsize$\pm$0.2677} & 0.4322\textcolor{customPurple}{\scriptsize$\pm$0.3201} & 0.7740\textcolor{customPurple}{\scriptsize$\pm$0.3008} \\
    \textcolor{black}{$\spadesuit$} DL-TTA (2022, TMI)~\cite{yang2022dltta}        & 0.4930\textcolor{customPurple}{\scriptsize$\pm$0.3193} & 60.47\textcolor{customPurple}{\scriptsize$\pm$119.30} & 0.3895\textcolor{customPurple}{\scriptsize$\pm$0.2779} & 0.4631\textcolor{customPurple}{\scriptsize$\pm$0.3365} & 0.7752\textcolor{customPurple}{\scriptsize$\pm$0.3643} \\
    \textcolor{black}{$\spadesuit$} SFDA-FSM (2022, MIA)~\cite{yang2022source}     & \cellcolor{best}0.8333\textcolor{customPurple}{\scriptsize$\pm$0.1139} & \cellcolor{best}10.67\textcolor{customPurple}{\scriptsize$\pm$9.25} & \cellcolor{best}0.7275\textcolor{customPurple}{\scriptsize$\pm$0.1385} & \cellcolor{best}0.7731\textcolor{customPurple}{\scriptsize$\pm$0.1462} & \cellcolor{best}0.9303\textcolor{customPurple}{\scriptsize$\pm$0.0853} \\
    \textcolor{black}{$\spadesuit$} STDR (2024, TMI)~\cite{wang2024dual}         & 0.4949\textcolor{customPurple}{\scriptsize$\pm$0.2537} & 61.55\textcolor{customPurple}{\scriptsize$\pm$58.31} & 0.3807\textcolor{customPurple}{\scriptsize$\pm$0.2146} & 0.5233\textcolor{customPurple}{\scriptsize$\pm$0.1997} & 0.7612\textcolor{customPurple}{\scriptsize$\pm$0.2362} \\
    \textcolor{black}{$\spadesuit$} AIF-SFDA (2025, AAAI)~\cite{li2025aif}     & 0.5692\textcolor{customPurple}{\scriptsize$\pm$0.2525} & 42.39\textcolor{customPurple}{\scriptsize$\pm$59.86} & 0.4368\textcolor{customPurple}{\scriptsize$\pm$0.2331} & 0.5373\textcolor{customPurple}{\scriptsize$\pm$0.2935} & 0.7671\textcolor{customPurple}{\scriptsize$\pm$0.2458} \\
    \arrayrulecolor{black}\hline
    \textcolor{green}{$\clubsuit$} DANN (2016, JMLR)~\cite{ganin2016domain}         & 0.5646\textcolor{customPurple}{\scriptsize$\pm$0.2482} & 50.38\textcolor{customPurple}{\scriptsize$\pm$59.86} & 0.4324\textcolor{customPurple}{\scriptsize$\pm$0.2289} & 0.5330\textcolor{customPurple}{\scriptsize$\pm$0.2893} & 0.7628\textcolor{customPurple}{\scriptsize$\pm$0.2416} \\
        \textcolor{green}{$\clubsuit$} DeTTA (2024, WACV)~\cite{wen2024denoising}        & 0.5049\textcolor{customPurple}{\scriptsize$\pm$0.2735} & 59.50\textcolor{customPurple}{\scriptsize$\pm$65.51} & 0.3807\textcolor{customPurple}{\scriptsize$\pm$0.2453} & 0.4577\textcolor{customPurple}{\scriptsize$\pm$0.3037} & 0.7637\textcolor{customPurple}{\scriptsize$\pm$0.2748} \\
            \textcolor{green}{$\clubsuit$} TestFit (2024, MIA)~\cite{zhang2024testfit}       & 0.5004\textcolor{customPurple}{\scriptsize$\pm$0.2693} & 58.50\textcolor{customPurple}{\scriptsize$\pm$65.51} & 0.3764\textcolor{customPurple}{\scriptsize$\pm$0.2411} & 0.4534\textcolor{customPurple}{\scriptsize$\pm$0.2995} & 0.7594\textcolor{customPurple}{\scriptsize$\pm$0.2706} \\
    \textcolor{green}{$\clubsuit$} UDA-MIMA (2024, CMPB)~\cite{hu2024unsupervised}     & 0.6359\textcolor{customPurple}{\scriptsize$\pm$0.1406} & 35.15\textcolor{customPurple}{\scriptsize$\pm$17.99} & 0.4809\textcolor{customPurple}{\scriptsize$\pm$0.1445} & 0.5715\textcolor{customPurple}{\scriptsize$\pm$0.1672} & \cellcolor{second}0.7811\textcolor{customPurple}{\scriptsize$\pm$0.2014} \\

    \textcolor{green}{$\clubsuit$} GraTa (2025, AAAI)        & 0.5636\textcolor{customPurple}{\scriptsize$\pm$0.2554} & 50.62\textcolor{customPurple}{\scriptsize$\pm$62.30} & 0.4337\textcolor{customPurple}{\scriptsize$\pm$0.2361} & 0.5336\textcolor{customPurple}{\scriptsize$\pm$0.2998} & 0.7675\textcolor{customPurple}{\scriptsize$\pm$0.2463} \\
    \arrayrulecolor{black}\hline
    \textcolor{red}{$\varheartsuit$} TENT (2021, ICLR)~\cite{wang2021tent}         & 0.5546\textcolor{customPurple}{\scriptsize$\pm$0.2550} & 51.80\textcolor{customPurple}{\scriptsize$\pm$62.28} & 0.4244\textcolor{customPurple}{\scriptsize$\pm$0.2342} & 0.5254\textcolor{customPurple}{\scriptsize$\pm$0.2992} & 0.7613\textcolor{customPurple}{\scriptsize$\pm$0.2474} \\
    \textcolor{red}{$\varheartsuit$} UPL-SFDA (2023, TMI)~\cite{wu2023upl}      & 0.4950\textcolor{customPurple}{\scriptsize$\pm$0.2972} & 67.90\textcolor{customPurple}{\scriptsize$\pm$88.33} & 0.3806\textcolor{customPurple}{\scriptsize$\pm$0.2652} & 0.4933\textcolor{customPurple}{\scriptsize$\pm$0.3183} & \cellcolor{third}0.7784\textcolor{customPurple}{\scriptsize$\pm$0.2929} \\
        \rowcolor[HTML]{F0F0F0}
    \textcolor{red}{$\varheartsuit$} DG-TTA (2023, arXiv)~\cite{weihsbach2023dg}        & 0.4805\textcolor{customPurple}{\scriptsize$\pm$0.2712} & 61.96\textcolor{customPurple}{\scriptsize$\pm$69.93} & 0.3583\textcolor{customPurple}{\scriptsize$\pm$0.2393} & 0.4270\textcolor{customPurple}{\scriptsize$\pm$0.2953} & 0.7560\textcolor{customPurple}{\scriptsize$\pm$0.2818} \\
    \textcolor{red}{$\varheartsuit$} SaTTCA (2023, MICCAI)~\cite{li2023scale}    & 0.5640\textcolor{customPurple}{\scriptsize$\pm$0.0711} & 50.45\textcolor{customPurple}{\scriptsize$\pm$18.69} & 0.4319\textcolor{customPurple}{\scriptsize$\pm$0.0644} & 0.5326\textcolor{customPurple}{\scriptsize$\pm$0.0596} & 0.7630\textcolor{customPurple}{\scriptsize$\pm$0.0762} \\
    \textcolor{red}{$\varheartsuit$} SmaRT (2025, arXiv)~\cite{Wang2025SmaRTSR}        & \cellcolor{second}0.7957\textcolor{customPurple}{\scriptsize$\pm$0.1038} & \cellcolor{second}16.39\textcolor{customPurple}{\scriptsize$\pm$13.10} & \cellcolor{second}0.6722\textcolor{customPurple}{\scriptsize$\pm$0.1347} & \cellcolor{second}0.6796\textcolor{customPurple}{\scriptsize$\pm$0.0957} & 0.7583\textcolor{customPurple}{\scriptsize$\pm$0.1722} \\
    \arrayrulecolor{black}\hline
    \textcolor{blue}{$\vardiamondsuit$} AdaMI (2022, MICCAI)~\cite{bateson2022test}        & 0.5573\textcolor{customPurple}{\scriptsize$\pm$0.2505} & 53.01\textcolor{customPurple}{\scriptsize$\pm$65.82} & 0.4256\textcolor{customPurple}{\scriptsize$\pm$0.2297} & 0.5240\textcolor{customPurple}{\scriptsize$\pm$0.2903} & 0.7554\textcolor{customPurple}{\scriptsize$\pm$0.2507} \\
        \textcolor{blue}{$\vardiamondsuit$} PASS (2024, TMI)~\cite{zhang2024pass}         & \cellcolor{third}0.6533\textcolor{customPurple}{\scriptsize$\pm$0.1775} & \cellcolor{third}31.70\textcolor{customPurple}{\scriptsize$\pm$23.92} & \cellcolor{third}0.5092\textcolor{customPurple}{\scriptsize$\pm$0.1848} & \cellcolor{third}0.6590\textcolor{customPurple}{\scriptsize$\pm$0.2357} & 0.7563\textcolor{customPurple}{\scriptsize$\pm$0.2227} \\

    \textcolor{blue}{$\vardiamondsuit$} VPTTA (2024, CVPR)~\cite{chen2024each}       & 0.5681\textcolor{customPurple}{\scriptsize$\pm$0.2596} & 46.62\textcolor{customPurple}{\scriptsize$\pm$62.30} & 0.4380\textcolor{customPurple}{\scriptsize$\pm$0.2402} & 0.5379\textcolor{customPurple}{\scriptsize$\pm$0.3039} & 0.7718\textcolor{customPurple}{\scriptsize$\pm$0.2504} \\

    \textcolor{blue}{$\vardiamondsuit$} ExploringTTA (2025, ISBI)~\cite{omolegan2025exploring}& 0.5640\textcolor{customPurple}{\scriptsize$\pm$0.0711} & 50.45\textcolor{customPurple}{\scriptsize$\pm$18.68} & 0.4319\textcolor{customPurple}{\scriptsize$\pm$0.0644} & 0.5326\textcolor{customPurple}{\scriptsize$\pm$0.0596} & 0.7630\textcolor{customPurple}{\scriptsize$\pm$0.0762} \\
    \textcolor{blue}{$\vardiamondsuit$} ProSFDA (2026, PR)~\cite{hu2025source}      & 0.5451\textcolor{customPurple}{\scriptsize$\pm$0.2618} & 52.98\textcolor{customPurple}{\scriptsize$\pm$63.91} & 0.4173\textcolor{customPurple}{\scriptsize$\pm$0.2404} & 0.5065\textcolor{customPurple}{\scriptsize$\pm$0.3016} & 0.7717\textcolor{customPurple}{\scriptsize$\pm$0.2497} \\
    \arrayrulecolor{black}\hline
  \end{tabular}%
  }
\end{table}

\section{Challenges and Prospects}
\subsection{Stable Adaptation and Streaming Forgetting Prevention}

Clinical inference often proceeds in small batches or even single instances, where noise in early predictions can be easily amplified by entropy-driven online updates, leading to parameter drift and catastrophic forgetting. This issue is particularly pronounced in cases with gradually shifting distributions in sequential clinical settings~\cite{wang2021tent}. Therefore, real-world deployment should not only emphasize immediate gains but also address three critical questions. The update target refers to which parameters are permitted to be adjusted. The update magnitude concerns how the intensity of adaptation is controlled. The rollback mechanism determines when and how to trigger reversion. To address these challenges, future studies should explore more robust sample selection strategies and anti-forgetting mechanisms to achieve stable adaptation in streaming scenarios~\cite{niu2022eata}.

\subsection{Style-content Disentanglement}

During the TTA stage, features extracted by pretrained models often encode both style and content simultaneously. This makes it difficult to disentangle which dimensions correspond to appearance style and which to anatomical semantics, thereby limiting the effectiveness of style alignment and content modeling. Existing studies suggest that shallow features tend to capture texture and style, whereas deeper features encode shape and content~\cite{geirhos2019imagenet,gatys2016style}. However, the validity of this hierarchical assumption in medical segmentation encoders remains insufficiently verified. Future research should develop interpretable disentanglement frameworks that can quantitatively assess the extent of style alteration and the degree of content preservation, thereby enabling the separate modeling and independent optimization of style and content.

\subsection{Test-Time Hyperparameter Adaptation}

In the absence of annotations, adaptation performance is highly sensitive to hyperparameters such as the learning rate, update steps, and the number of updatable layers; however, the lack of a validation set renders grid search infeasible. Existing approaches typically rely on entropy or consistency as surrogate signals for hyperparameter adjustment~\cite{wang2021tent,sun2020ttt}. However, these signals fluctuate substantially under domain shifts and noise perturbations, leading to unstable performance across different centers. Future efforts should advance in two directions: first, hyperparameter robustness, namely designing objective functions and optimization schemes that remain stable across broad hyperparameter ranges. Second, hyperparameter self-adaptation, namely leveraging surrogate signals at the case level to automatically determine whether to update, the magnitude of update, and the set of updatable layers, thereby reducing manual tuning costs and alleviating cross-domain variability.

\subsection{Representation Alignment and Class Prototypes}

In unlabeled target domains, pseudo-label noise and class imbalance distort the alignment direction, making it particularly prone to boundary drift across institutions and scanners. The introduction of class prototypes provides a stable anchor to mitigate this problem. In 3D organ segmentation, prototype constraints have been employed to alleviate spatial deformations during cross-institutional transfer, demonstrating robustness~\cite{li2022prototypical}. In scenarios that more closely approximate domain shifts, prototypes can further map target-domain features into a source-style decodable space, thereby reducing decoders' sensitivity to distributional discrepancies~\cite{wang2024pfmnet}. Key challenges for future research include developing structure-aware prototype estimation under unlabeled streaming inputs, quantifying prototype uncertainty, and maintaining stable pixel-to-prototype associations in cases involving small objects or ambiguous boundaries.

\subsection{Prior Estimation and Weak Prompts}

When priors do not match the target domain, they can cause excessive smoothing or incorrect segmentation, while organ deformation and lesion heterogeneity further undermine their reliability across cases~\cite{zhang2024pass}. In practical TTA scenarios, the fundamental problem is how to extract case-level priors on intensity distribution, scale, and coarse shape with associated confidence scores, and how to dynamically attenuate their influence when they conflict with observations while retaining traceable records, thereby enabling systematic evaluation of heterogeneous effects across tissues and modalities~\cite{zhu2025improving}.

\subsection{Semantic Anchors and Retrieval-Enhanced Adaptive Closed Loop}

With the expansion of unlabeled testing scenarios, large-scale medical-domain VLMs such as BioViL and BiomedCLIP, owing to their stronger generalization across institutions and scanners, can provide robust semantic representations for segmentation TTA~\cite{boecking2022biovil,zhang2023biomedclip}. Meanwhile, general-purpose VLMs such as CLIP and BLIP-2 possess programmable prompting interfaces that, through appropriately designed prompts, can simultaneously capture both specific and standard features of the target domain, thereby facilitating source-target semantic alignment~\cite{radford2021clip}. A promising direction is to utilize VLMs as intelligent agents that dynamically retrieve and aggregate priors, such as shape, scale, and contextual information, for each incoming case. These priors are then injected into pretrained segmentation models in a controlled manner, creating a self-adaptive closed loop that delivers stronger performance at test time.

\section{Use of Large Language Models and Research Integrity}
This manuscript used ChatGPT GPT-5 Thinking solely for language polishing and clarity improvement. The model was not used to generate or edit scientific content, analysis code, results, references, or citations. All references and citations were compiled by the authors from primary sources and verified line by line. Numerical values, method descriptions, and conclusions were checked against the original data and code to ensure accuracy and traceability.

\section{Conclusion}
MedSeg-TTA provides a unified, comprehensive assessment of TTA methods for medical image segmentation across seven medical imaging modalities and multiple clinically relevant tasks. By rigorously controlling preprocessing procedures, backbone settings, and evaluation protocols, the benchmark reveals the heterogeneous behavior of four major adaptation paradigms and exposes their strengths, limitations, and stability boundaries under both mild and severe domain shifts. The findings demonstrate that adaptation performance is strongly modality-dependent and that several widely used methods may yield unreliable outcomes when confronted with significant distribution gaps. These insights underscore the importance of systematic evaluation and careful methodological consideration when implementing adaptive models in clinical settings. By releasing standardized datasets, reproducible implementations, and an open leaderboard, MedSeg-TTA establishes a reference point for future research, encouraging the development of more robust, interpretable, and clinically reliable adaptation strategies.

\section*{CRediT authorship contribution statement}
\textbf{Wenjing Yu}: Methodology, Visualization, Writing - original draft.  
\textbf{Shuo Jiang}: Methodology, Validation, Writing - original draft.  
\textbf{Yifei Chen}: Methodology, Conceptualization, Writing - original draft.  
\textbf{Shuo Chang}: Validation, Visualization.  
\textbf{Yuanhan Wang}: Validation, Visualization.  
\textbf{Beining Wu}: Visualization.  
\textbf{Jie Dong}: Visualization.  
\textbf{Mingxuan Liu}: Validation, Visualization.  
\textbf{Shenghao Zhu}: Validation, Visualization.  
\textbf{Feiwei Qin}: Supervision, Writing - review \& editing.  
\textbf{Changmiao Wang}: Writing - review \& editing.  
\textbf{Qiyuan Tian}: Project administration, Writing - review \& editing.

\section*{Declaration of competing interest}
The authors declare that they have no known competing financial interests or personal relationships that could have appeared to influence the work reported in this paper.

\section*{Acknowledgments}
This work was supported by the National Natural Science Foundation of China (No. 82302166), Tsinghua University Startup Fund, Fundamental Research Funds for the Provincial Universities of Zhejiang (No. GK259909299001-006), Anhui Provincial Joint Construction Key Laboratory of Intelligent Education Equipment and Technology (No. IEET202401), and the State Key Lab of CAD\&CG, Zhejiang University (A2510).

\bibliographystyle{elsarticle-num-names}

\end{document}